%% file: main.tex
\pdfoutput=1
\documentclass{article}
\usepackage[T1]{fontenc}
\usepackage{microtype}
\usepackage{amsmath,amssymb}
\usepackage{booktabs}
\usepackage{multirow}
\usepackage{makecell}
\usepackage{xcolor}
\usepackage{graphicx}
\usepackage{pifont}
\usepackage{algorithm}
\usepackage{algorithmic}
\usepackage{tikz}
\usepackage{pgfplots}
\usepgfplotslibrary{groupplots}
\usetikzlibrary{patterns,calc}
\usepackage{colortbl}
\pgfplotsset{compat=1.16}
\usepackage[hypertexnames=false]{hyperref}

\usepackage[accepted]{mlsys2025}
\makeatletter\renewcommand{\Notice@String}{}\makeatother
\hypersetup{pdfsubject={Preprint}, pdfauthor={Tianle Zhou, Yang Zhang, Tianyu Wu, Zhuohan Wang, William Nixon, Juncheng Yang}}

\newcommand{\sys}{LoopKV}

\newif\ifshowcomments\showcommentsfalse

\definecolor{negpink}{HTML}{E0218A}

\newif\iflucastodo\lucastodofalse

\newif\iflucasmark\lucasmarkfalse

\newcommand{\planR}{read-optimized plan}
\newcommand{\planM}{memory-optimized plan}
\newcommand{\planU}{KIVI-4 + loop-3 sharing plan}

\newcommand{\pp}{\,pp}

\newcommand{\x}{$\times$}
\newcommand{\Pz}{\ensuremath{\mathrm{KL}_{\mathrm{share}}}}
\definecolor{lkblue}{HTML}{2A78D6}
\definecolor{cBase}{HTML}{6B7585}\definecolor{cPtr}{HTML}{2C6AA8}\definecolor{cRes}{HTML}{E0893A}
\definecolor{cDm}{HTML}{B4232A}\definecolor{cPlan}{HTML}{7B4FA0}
\definecolor{fbLight}{HTML}{DEEBF7}\definecolor{fbMid}{HTML}{9ECAE1}\definecolor{fbDark}{HTML}{4292C6}\definecolor{fbNavy}{HTML}{1F3B73}
\definecolor{bfblue}{HTML}{DCEBF8}\definecolor{oursred}{HTML}{F6CFCF}
\definecolor{cellD}{HTML}{2C5F9E}\definecolor{celld}{HTML}{9DC3E6}\definecolor{cellP}{HTML}{F2B36B}\definecolor{cellA}{HTML}{3B4A5C}
\definecolor{syn}{HTML}{D9480F}\definecolor{hot}{HTML}{FFC233}

\tikzset{pD/.style={fill=cellD},pd/.style={fill=celld},pP/.style={fill=cellP},pA/.style={fill=cellA}}
\colorlet{pD}{cellD}\colorlet{pd}{celld}\colorlet{pP}{cellP}\colorlet{pA}{cellA}
\input{figures/v4/gen_cells.tex}
\input{figures/v4/gen_pareto.tex}

\graphicspath{{figures/}}
\newtheorem{proposition}{Proposition}
\newenvironment{tightitems}{\begin{itemize}\setlength{\itemsep}{1pt}\setlength{\parskip}{0pt}\setlength{\topsep}{2pt}}{\end{itemize}}

\mlsystitlerunning{LoopKV: Loop-Aware KV Cache Compression for Serving Looped Transformers}

\newsavebox{\sysbox}
\makeatletter\newcommand{\figcaptype}{\def\@captype{figure}}\makeatother
\begin{document}
\sbox{\sysbox}{\includegraphics[width=\textwidth]{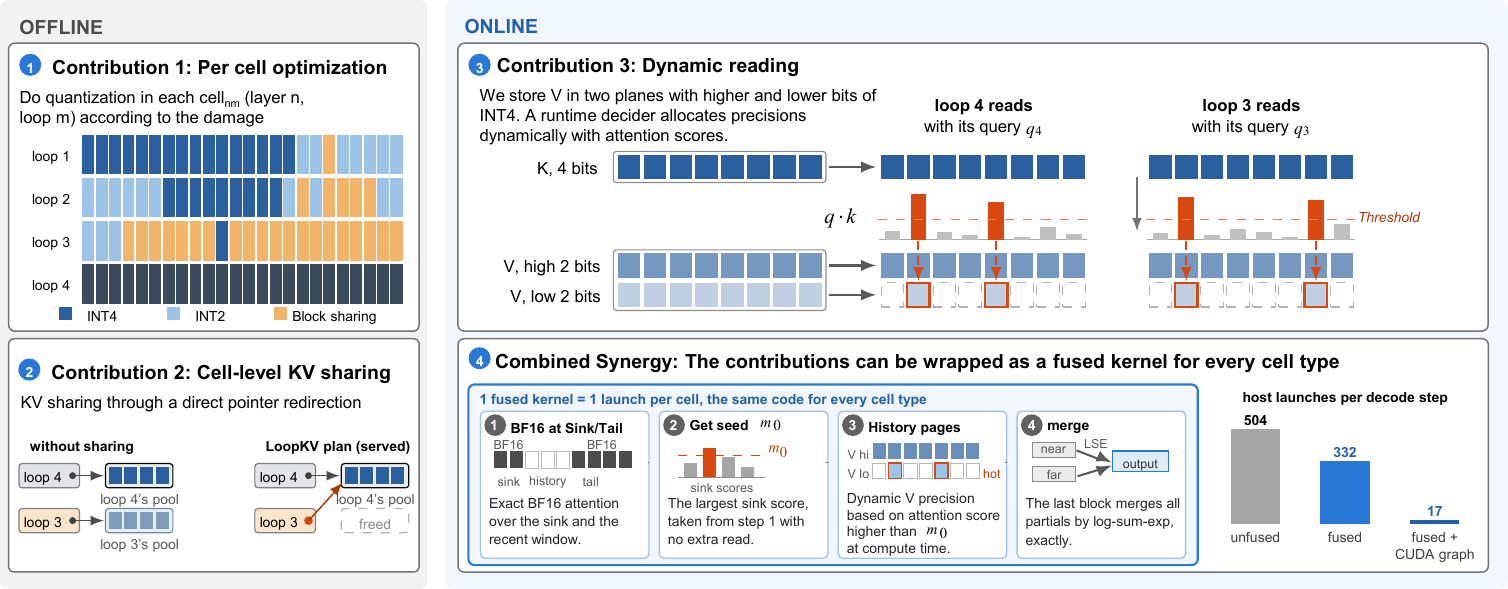}}
\twocolumn[
\mlsystitle{LoopKV: Loop-Aware KV Cache Compression\\ for Serving Looped Transformers}
\mlsyssetsymbol{equal}{*}
\begin{mlsysauthorlist}
\mlsysauthor{Tianle Zhou}{equal,harvard}
\mlsysauthor{Yang Zhang}{equal,ucsd}
\mlsysauthor{Tianyu Wu}{equal,harvard}
\mlsysauthor{Zhuohan Wang}{equal,harvard}
\mlsysauthor{William Nixon}{uchicago}
\mlsysauthor{Juncheng Yang}{harvard}
\end{mlsysauthorlist}
\mlsysaffiliation{harvard}{Harvard University}
\mlsysaffiliation{ucsd}{University of California San Diego}
\mlsysaffiliation{uchicago}{University of Chicago}
\mlsyscorrespondingauthor{Tianle Zhou}{tianle\_zhou@fas.harvard.edu}
\mlsyskeywords{Looped transformers, KV cache, quantization, inference serving}
\vskip 0.3in
\begin{abstract}
Looped transformers apply a shared block several times per token to improve accuracy without adding parameters.
For models such as Ouro that keep a separate key-value (KV) cache per loop, cache memory limits the batch size that amortizes repeated weight reads.
Uniform-bit quantization and fixed rules per loop do not allocate precision according to each layer's sensitivity in each loop.
Measurements on Ouro show that quantization sensitivity varies across (loop, layer) cells, and the sum of individual cell damages approximates the overall output damage.
Sharing damage also varies by cell, with later loops generally tolerating sharing better.
Skipping low-attention values to save reads produces errors that compound across loops.
\sys{} uses a one-time offline profile to choose 4-bit storage, 2-bit storage or reuse of the final loop's KV for each cell.
The planner minimizes reads under budgets for predicted damage and storage.
During decoding, each loop reads the high 2 bits of each 4-bit value and fetches the low bits only where its own attention weights require 4-bit precision.
A single fused kernel in vLLM executes the plan.
Serving Ouro-1.4B on one RTX PRO 6000, the memory-optimized \sys{}-M plan stores 35\% less KV than uniform 4-bit quantization (KIVI-4) on the same fused kernel and CUDA graph.
It sustains \textbf{1.7--2.2\x{}} KIVI-4's goodput under a latency service-level objective with a 16K-token maximum length.
None of its averages on six public benchmarks falls significantly below BF16's.
 
\end{abstract}
\vskip 0.15in
\begin{minipage}{\textwidth}
\centering
\usebox{\sysbox}
\vspace{-5mm}
\figcaptype\caption{\textbf{Overview of \sys{}.} The offline phase of \sys{}, run once per model, chooses a storage option for every (loop, layer) pair, which we call a cell, from the KL damage measured by profiling each cell independently (C1). A cell may reuse the final loop's KV through a pointer when its KL damage is low, which we show costs little accuracy (C2). In the online phase, V is stored as a high and a low 2-bit plane, so that the attention scores decide at decode time which tokens also read the low plane (C3). We then combine all three into one fused kernel that runs every cell type in a single launch. With vLLM's whole-step CUDA graph, this cuts \sys{}-M's host launches per decode step from 504 to 17.}
\label{fig:system}
\vspace{-2mm}
\end{minipage}
\vskip 0.12in
]
\printAffiliationsAndNotice{\textsuperscript{*}Equal contribution. }

\section{Introduction}\label{sec:intro}
\iffalse
\begin{figure*}[t]
\centering
\begin{minipage}[t]{0.322\textwidth}\centering
\includegraphics[width=\linewidth]{figures/v4/fig_ev_p1.pdf}\\[0.5mm]
\parbox{\linewidth}{\footnotesize\textbf{(a) P1: damage depends on the cell.} Top: INT2 damage of each (loop, layer) cell; 11 of the 72 cells cause half of it. Bottom: at the same bytes, planning per cell gives 1.4--1.5\x{} lower KL than planning per layer and 12\x{} lower than uniform INT2.}
\end{minipage}\hfill
\begin{minipage}[t]{0.322\textwidth}\centering
\includegraphics[width=\linewidth]{figures/v4/fig_ev_p2.pdf}\\[0.5mm]
\parbox{\linewidth}{\footnotesize\textbf{(b) P2: later loops can read loop 4's KV.} With one layer reading loop 4's KV, sharing costs less than storing that cell at INT2 in 21 of 24 layers of loop 3, 7 of loop 2 and 2 of loop 1.}
\end{minipage}\hfill
\begin{minipage}[t]{0.322\textwidth}\centering
\includegraphics[width=\linewidth]{figures/v4/fig_ev_p3.pdf}\\[0.5mm]
\parbox{\linewidth}{\footnotesize\textbf{(c) P3: loops sharing a page attend to different tokens.} Each value is stored as a high and a low 2-bit plane; a loop reads the low plane only for its own important tokens (weight $\ge 2^{-14}$ of the row maximum), at no extra bytes. Over 16 documents these tokens overlap loop 4's by 0.84, 0.71 and 0.51 for loops 3, 2 and 1.}
\end{minipage}
\vspace{-1mm}
\caption{Evidence for the three properties on Ouro-1.4B (fp32, 16 documents, KL relative to \Pz{}).}
\label{fig:obs}
\vspace{-2mm}
\end{figure*}
\fi

Looped transformers, which reuse one block of layers for several iterations per token, are increasingly used as an architectural component of large language models \cite{dehghani2019universal,saunshi2025reasoning}. Because every iteration reuses the same weights, a looped model reaches the effective depth of a much deeper model, i.e., the number of layers times the number of loops, while storing far fewer layers. It therefore needs less memory to hold its parameters and achieves higher accuracy per parameter \cite{saunshi2025reasoning}. For example, Ouro matches standard models with two to three times its parameter count by looping its layers four times \cite{zhu2025ouro}. Huginn, with 3.5B parameters, keeps improving its reasoning as it iterates its recurrent block more, up to a compute load equivalent to a 50B-parameter model \cite{geiping2025huginn}. Mixture-of-Recursions chooses for each token how many times the shared block runs and reaches lower perplexity than standard transformers at equal training compute with fewer unique parameters \cite{bae2025mor,bae2025relaxed}.

Serving these models, however, is costly, because for some of them the KV cache is large and each decode step reads many more KV blocks from memory than a standard model of the same size does. Ouro, for example, keeps a separate KV cache for every loop, so Ouro-1.4B stores 768\,KiB of KV per token in BF16, four times what a dense model with the same layers would store, and every decode step re-reads its 2.5\,GB of weights four times. A larger batch amortizes these weight re-reads, but then KV memory becomes the bottleneck, because the cache stored per request sets how many requests fit in a batch and the bytes read per step set the decode time (\S\ref{sec:bg-kv}). At BF16's peak-throughput batch for 8K-token contexts on Ouro-1.4B, the KV cache accounts for 84\% of the bytes each decode step reads. In this setting, compressing the cache raises both the batch size and the decoding speed, which is the key to higher throughput and goodput.\looseness=-1

General KV quantization methods, including FP8, KIVI \cite{liu2024kivi}, KVQuant \cite{hooper2024kvquant} and rotation-based methods \cite{ashkboos2024quarot,zandieh2025turboquant}, treat each loop of a looped transformer as one more layer with the same bits. Only a few works target the features specific to looped transformers, and we find two concurrent works that exploit the loop structure. FlashLoop updates only a subset of tokens in each loop, guides each loop's sparse attention with the previous loop and quantizes chained INT4 residuals \cite{yang2026flashloop}. ResidualQuant keeps an INT4 copy of the final loop and stores only 2-bit residuals for every earlier loop \cite{kim2026residualquant}. Both assign precision on a per-loop basis. A per-loop setting, however, treats every layer of a loop alike, which is harmful when the layers within a loop behave differently. For example, our experiments show that under aggressive quantization a few middle layers cause most of the damage, while some later layers are much less sensitive (\S\ref{sec:props}). Some layers can therefore use fewer bits and leave more of the budget to the layers that matter more. What remains missing is a decision at the granularity of each (loop, layer) pair, e.g., which precision each pair needs and whether a pair can reuse another loop's KV instead of storing its own when its damage is low enough.

\begin{figure}[t]%

\centering
\includegraphics[width=\columnwidth]{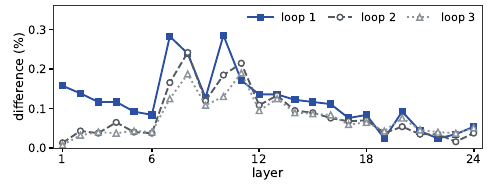}%
\vspace{-5mm}
\caption{\textbf{P1: damage depends on the cell.} Each point shows the damage when one cell (one layer in one loop) is stored at INT2 while every other cell stays exact. The damage on the y-axis is the share of next-token predictions that change (fp32, 16 documents). The top 19 of the 72 cells hold half of the total difference, and KL ranks the cells alike (Spearman 0.95).}
\label{fig:p1}
\vspace{-2mm}
\end{figure}

\begin{figure}[t]
\centering
\includegraphics[width=\columnwidth]{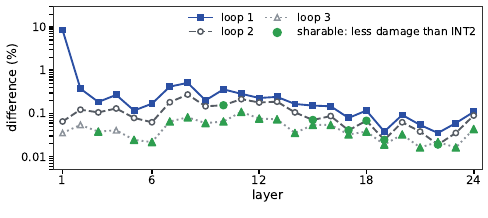}%
\vspace{-5mm}
\caption{\textbf{P2: some cells can share with little loss.} Each point shows the damage (the y-axis, as in Figure~\ref{fig:p1}) when one cell reads the final loop's KV instead of storing its own, while every other cell stays exact. Some cells (green) do less damage when sharing than when stored at INT2}
\label{fig:p2}
\vspace{-2mm}
\end{figure}

\begin{figure}[t]
\centering
\includegraphics[width=\columnwidth]{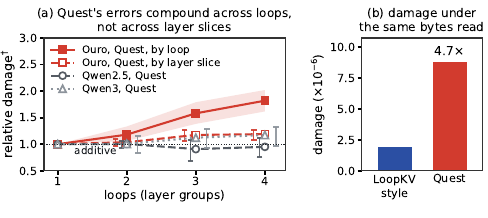}%
\vspace{-5mm}
\caption{\textbf{P3: the loops compound read errors.} Quest drops the low-ranked far pages that are not in the sink and window, while LoopKV-style reading keeps them and reads the values of low-attention far tokens at 2 bits. Both read 85\% of the far bytes. (a) Relative damage$^\dagger$ of Quest as $k$ grows, applied to the first $k$ loops of Ouro-1.4B, to $k$ of four layer slices of Ouro-1.4B that each span all loops, or to $k$ of four layer groups of Qwen2.5 and Qwen3. (b) Damage of LoopKV-style reading and of Quest in all four loops of Ouro-1.4B at the same 85\% of bytes read. Damage is the fp32 KL to the exact model. $^\dagger$Damage when the $k$ groups are treated together, divided by the sum of their damages when page skipping is applied to each group alone.}
\label{fig:p3}
\vspace{-2mm}
\end{figure}

We propose \sys{}, which compresses the KV cache of looped transformers at the granularity of a (loop, layer) pair, which we call a \emph{cell}. \sys{} first profiles the damage of each cell offline and then jointly chooses every cell's precision and whether it shares the final loop's KV, so that the bytes saved on robust cells can be spent on the more sensitive ones. This planner minimizes the bytes read per decode step under budgets for KL damage and storage. Looping also raises two questions that a transformer with one KV cache per layer does not, i.e., whether the KV of the late loops converges enough to be shared, and whether an error introduced in one loop compounds as later loops process it again. Our measurements of Ouro's KV cache answer both questions and show why the decision must be made per cell rather than per loop (\S\ref{sec:props}, Figures~\ref{fig:p1}--\ref{fig:p3}).

\textbf{(P1) Sensitivity depends on the cell:} a few cells cause most of the quantization damage, and which cells they are depends on the model. Cell damages also approximately add up to the model's final KL divergence, so a single profile per model predicts a plan's damage before it is served (Figure~\ref{fig:p1}).
\textbf{(P2) Sharing does less damage in later loops:} for some cells, reading the final loop's KV costs less than storing their own INT2 copy, e.g., in 21 of the 24 loop-3 layers and 6 loop-2 layers, but in no loop-1 layer (Figure~\ref{fig:p2}).
\textbf{(P3) The loops compound read errors:} each loop injects nearly the same error, so skipping the pages of unimportant tokens costs more on looped models than on plain ones. In Ouro-1.4B, page skipping in all four loops costs 1.8\x{} the sum of its one-loop damages in KL, over 50\% more than the at most 1.17\x{} of plain models such as Qwen2.5 and Qwen3. At equal bytes, LoopKV-style reads, which read the values of unimportant tokens at 2 bits instead of skipping them, cause 4.7\x{} less damage. Loops that share a block, moreover, attend to different tokens (Jaccard 0.83 for loops 3 and 4), so which values need full precision depends on the loop that reads them and must be decided at run time (Figure~\ref{fig:p3}).

\begin{figure*}[t]
\centering
\includegraphics[width=\textwidth]{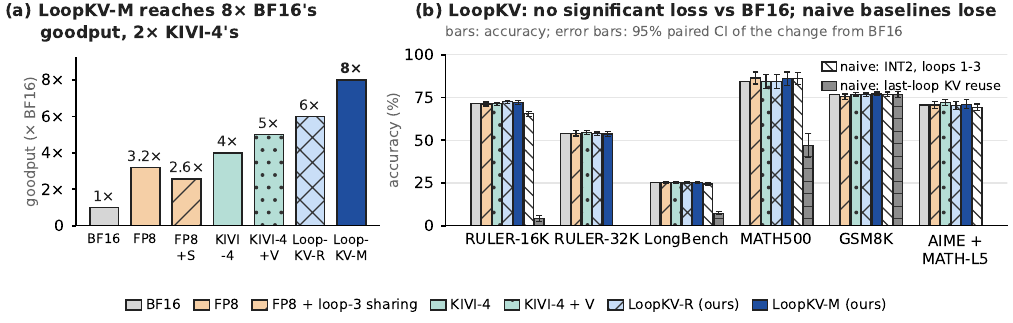}%
\vspace{-5mm}
\caption{\sys{} serves more requests with no significant accuracy loss on six public suites (tested on Ouro-1.4B). (a) Goodput relative to BF16 on one RTX PRO 6000 for the memory-optimized plan (\sys{}-M), which minimizes reads under a tight storage budget, the read-optimized plan (\sys{}-R), which minimizes reads with no storage budget, and comparable methods. Goodput is the highest request rate at which 95\% of requests meet the SLO\protect\footnotemark{}. In the figure, \sys{}-M sustains 8.0\x{} BF16's goodput and 2.0\x{} that of KIVI-4 (\S\ref{sec:eval-slo}). (b) Over all the public suites we tested (Table~\ref{tab:acc}), neither \sys{}-M nor \sys{}-R is significantly less accurate than BF16, whereas naive INT2 on loops 1--3 in the KIVI style and naive last-loop KV reuse lose accuracy significantly on several suites. Other comparable methods: KIVI-4 + V adds our V precision to KIVI-style INT4 on every loop, on the same fused kernel and graph. BF16, FP8 and FP8 + loop-3 sharing use vLLM's own kernels, and KIVI-4 is uniform INT4 without V precision.}%
\label{fig:perf}
\vspace{-2mm}
\end{figure*}

In the offline phase, \sys{} profiles every cell once and measures its damage under each storage option, i.e., 4-bit codes, 2-bit codes (P1) or a pointer to the final loop's KV (P2). It then solves a knapsack over these profiles that minimizes the KV bytes each decode step reads under a damage budget. Adding a tight storage budget gives the memory-optimized plan (\sys{}-M), and leaving storage unconstrained gives the read-optimized plan (\sys{}-R). In the online phase, \sys{} stores each 4-bit value as two 2-bit planes and reads them at a precision set by the attention each token receives, i.e., a loop reads the low plane only for the tokens that its own attention marks as important (P3). Loops that share a block can therefore read it at different precisions without extra bytes. In vLLM, one fused kernel executes every cell type in a single launch, and the whole decode step replays as one CUDA graph, which cuts \sys{}-M's host launches per decode step from 504 to 17 (Figure~\ref{fig:system}).
Our contributions are:
\begin{tightitems}
\item \textbf{Characterization.} We measure three properties of the per-loop KV cache on Ouro-1.4B (P1--P3) and show that its sensitivity profile is specific to each model, since it differs across Ouro-2.6B, Dory-4B and Huginn-3.5B. On each of these models, a plan computed from the model's own per-cell profile beats hand-made plans of equal bytes (\S\ref{sec:props}, \S\ref{sec:eval-general}).
\item \textbf{Design.} \sys{} turns each property into one mechanism, and one fused kernel executes all three in a single launch. They include per-cell mixed-precision allocation (C1), cross-loop KV sharing as a per-cell option, with the planner deciding which cells share (C2), and read-time, attention-aware V precision (C3) (\S\ref{sec:design}, \S\ref{sec:impl}). Restricting the plan to one option per layer costs 1.4--1.9\x{} the KL at equal bytes.
\item \textbf{Evaluation.} On Ouro-1.4B, Ouro-2.6B and Dory-4B, \sys{}-M reaches 1.13--4.44\x{} the decode throughput of vLLM's BF16 cache at 16K context on three kinds of GPUs, and none of its suite averages falls more than 0.5 percentage points below BF16's (up to six suites per model) (Table~\ref{tab:acc}, Table~\ref{tab:tputx}, \S\ref{sec:eval-general}). On Ouro-1.4B, against uniform INT4 (KIVI-4) on the same kernel and CUDA graph, \sys{}-M stores 35\% less KV and sustains 1.7--2.2\x{} KIVI-4's request rate under a latency SLO (Figure~\ref{fig:perf}).
\end{tightitems}

\edef\slofn{\arabic{footnote}}%
\section{Background and Motivation}\label{sec:bg}

\subsection{Looped and recurrent-depth transformers}\label{sec:bg-looped}
A looped transformer applies a block $f_\theta$ of $L$ layers $T$ times, $h^{(t)} = f_\theta(h^{(t-1)})$, and decodes from $h^{(T)}$ \cite{dehghani2019universal,saunshi2025reasoning}.
In loop $t$, layer $\ell$ writes keys and values $K^{(t)}_\ell, V^{(t)}_\ell$ and attends over the same loop and layer's KV of all earlier tokens.
Ouro-1.4B has $T{=}4$, $L{=}24$ and 16 KV heads of dimension 128 without grouped-query attention, so it stores 768\,KiB of BF16 KV per token \cite{zhu2025ouro}.
Huginn-3.5B iterates a 4-layer core $r{=}32$ times between a prelude and a coda \cite{geiping2025huginn}; Dory-4B \cite{nvidia2026dory} runs $T{=}4$ loops over 20 recurrent attention layers (5 full-attention, 15 sliding-window), plus 16 non-recurrent attention layers.
\sys{} keeps the model's weights, architecture and number of loops and compresses only its KV cache, exploiting how a looped model uses its KV blocks across loops and layers. We therefore compare \sys{} with general KV cache compression methods, e.g., KIVI \cite{liu2024kivi} and TurboQuant \cite{zandieh2025turboquant}, and with methods designed for looped transformers, e.g., FlashLoop \cite{yang2026flashloop} and ResidualQuant \cite{kim2026residualquant}.

\subsection{KV cache in decoding}\label{sec:bg-kv}
A decode step processes one token for each of $B$ requests in
\begin{equation}
t(B) = t_\text{fix} + \bigl(W_T + B\cdot D_\text{read}\bigr)/\text{BW},\label{eq:step}
\end{equation}
the memory-bound view of decoding \cite{williams2009roofline,pope2023scaling}. In the formula, $W_T$ counts the bytes of the $T$-fold weight read per step, $D_\text{read}$ the KV bytes one request reads per step, BW is the effective bandwidth of the whole step, and $t_\text{fix}$ is the step time that does not scale with bytes. In serving, the batch is bounded by $B\le(\text{KV pool})/D_\text{stored}$, where $D_\text{stored}$ is the KV bytes one request stores \cite{kwon2023vllm}.
Unlike a standard model, a looped transformer reads its weights $T$ times per step. Sharing, residuals and V precision also make the bytes it reads differ from the bytes it stores, so optimizing $D_\text{read}$ and optimizing $D_\text{stored}$ are separate goals.
In our experiments, Eq.~(\ref{eq:step}) predicts the peak throughput of the quantized formats we checked within 3\%\footnote{On our RTX PRO 6000, $W_T{=}10.07$\,GB, and a fit gives BW$\,{=}\,1.29$\,TB/s and $t_\text{fix}{=}6.84$\,ms. The formats were served by our plugin before kernel fusion and peak at 98--103\% of the prediction.}.
For a history of $N$ tokens, $D_\text{read}{=}N d_\text{read}+D_\text{exact}$, and likewise for storage, where $d$ is the bytes per history token over all layers and loops, which we report in KiB, and $D_\text{exact}$ is the cost of the sink and window tokens kept in full BF16 precision (Appendix~\ref{app:bytes}).

Our experiments show that once the KV cache is quantized, throughput is bounded by the bytes read, not the bytes stored. At the same 180\,KiB stored per token, KIVI-4 with loop 3 sharing loop 4's KV reads 240\,KiB per token and decodes 1.36\x{} as fast at 8K as a residual-coding baseline with the same KIVI-style quantization, which reads 360\,KiB. In contrast, uniform KIVI-4 reads the same 240\,KiB but stores 33\% more, and it is only 2\% slower (all before kernel fusion, Table~\ref{tab:tput} and Table~\ref{tab:o5} in Appendix~\ref{app:bytes}).

\subsection{Measuring KV damage}\label{sec:characterize}
\footnotetext[\slofn]{A request meets the SLO if its first token arrives within 10\,s and its mean time per output token stays below $3\tau_0{=}55.8$\,ms, for AIME prompts whose contexts grow to 16K tokens (\S\ref{sec:eval-setup}).}
We measure the damage of a KV policy with the KL divergence between the exact model and the model under that policy \cite{dutta2024accuracy}.
As in serving, the sink and the recent window stay BF16 (a 32-token sink and a 256-token window), and each 32-token page is processed by the policy when it leaves the window.
We report damage in units of \Pz{}, the KL of the cheapest whole-loop sharing\footnote{Naively letting loop 3 reuse loop 4's BF16 KV beyond the window, which gives a KL of $1.20{\times}10^{-4}$.}.
All these measurements run in fp32 activations\footnote{In BF16, rounding noise alone produces about half the KL of \Pz{} and would hide the differences between cells, whereas in fp32 every effect we report lies well above the noise floor (Appendix~\ref{app:extra}). Every KL in this paper is fp32 unless marked ``served''. \Pz{} is a unit, not an acceptance threshold, and the task scores of \S\ref{sec:eval} test whether a KL difference matters.}.
Under this measurement, the memory- and read-optimized plans of \sys{} keep the exact model's top token at 99.63\% and 99.72\% of positions.

\iffalse
\begin{figure}[t]
\centering
\resizebox{\columnwidth}{!}{\input{figures/v4/fig_sens2d.tex}}
\vspace{-2mm}
\caption{%
Damage depends on loop and layer, and sharing is cheap only in loop 3. Ouro-1.4B, fp32 KL of coding one cell (16 documents, log scale): (a) as own INT2, loop 1 costs most and layers 7, 8, 10 and 11 (shaded band) dominate; (b) reading loop 4's INT4 KV costs loop 3 at most 0.14\x{} \Pz{}, but loop 1 up to 2.6\x{} and 1074\x{} in its first layer. Loop sums and own INT4: Appendix~\ref{app:extra}.}
\label{fig:sens}
\vspace{-2mm}
\end{figure}
\fi

\subsection{Three loop-specific properties}\label{sec:props}
We observe three loop-specific properties in our experiments, and each of them leads to a corresponding optimization in \sys{} (\S\ref{sec:design}).

\paragraph{P1: Sensitivity depends on the cell, not just the layer.}
When we quantize one cell of the history at a time to KIVI-style INT2 (Figure~\ref{fig:p1}), a few layers within a loop turn out to be much more sensitive than the others. Loop 1 and four middle layers (7, 8, 10 and 11) cause most of the KL damage. These four layers are on average 4.9\x{} as sensitive as the other layers and hold half of the total damage, while the last layers add almost no damage. At INT4, three quarters of all cells cost less than 1\% of \Pz{}, so most cells can be quantized more aggressively. Sensitivity also varies as much with the loop as with the layer, and loop 1's cells are together 3 times as sensitive as loop 3's. Which loop dominates depends on the model, i.e., the first on Ouro but the last on Dory and Huginn in fp32 (\S\ref{sec:eval-general}), so no rule fixed by loop index transfers across models without a model-specific calibration. We therefore profile each cell once before planning. Fortunately, profiling each cell once is enough, because a plan's KL is approximately predictable from the sum of its cells' damages (Proposition~\ref{prop:add} in Appendix~\ref{app:additivity}). Over the 14 plans and residual designs we test, the summed per-cell damage and the measured KL have a correlation of 0.98.

\paragraph{P2: Some cells can be shared with less damage than INT2 quantization.}
A cell can read the final loop's KV instead of storing its own, which removes its storage entirely. For some cells, this sharing does less damage than quantizing the cell to INT2 in the KIVI style. On Ouro-1.4B, sharing changes fewer predictions than own INT2 in 21 of the 24 layers of loop 3 and in 6 layers of loop 2 (Figure~\ref{fig:p2}). Letting loop 3 read loop 4's KV stays non-inferior to the exact model under a 3-point margin on four of the five suites in Table~\ref{tab:quality}, all but the MQ641 stress test. After profiling, \sys{} can therefore share such cells with the final loop without a large impact on accuracy. Sharing in earlier loops costs far more, changing about 4 times as many predictions per loop-1 cell as per loop-3 cell.\footnote{The Ouro authors report that letting every loop read the last loop's KV costs at most 2 points with few-shot prompts on the base model \cite{zhu2025ouro}. When the Thinking model generates up to 8K tokens in our engine, the same reuse loses 44.4 points on MATH500, whereas loop 3 reading loop 4 loses at most 1.0 point (Appendix~\ref{app:negative}).} This loss cannot be recovered by aligning the loops, because their keys have nearly equal norms but different directions, so a MiniCache-style norm correction \cite{liu2024minicache} recovers only 2.7\% of the sharing KL (Appendix~\ref{app:negative}). Some models are trained so that their loops share KV, e.g., Dory's last-loop mode, whereas \sys{} serves the released model and treats sharing as one more per-cell option, deciding from each cell's measured KL where sharing pays off.

\paragraph{P3: The loops compound read errors.}
Every loop re-reads the whole history at each step, and these reads bound the decoding speed. At the 8K throughput peak, the history reader takes 75\% of GPU time (Table~\ref{tab:breakdown}). Yet attention is sparse. In every loop, about one history token in six receives an attention weight of at least $2^{-14}$ of the largest weight in its row (Table~\ref{tab:tau}), and the same holds for Dory and Huginn (Appendix~\ref{app:general}). Traditional methods skip or drop the low-attention values to save reads. Looped transformers, however, are more sensitive to such dropping. Each loop repeats the read with the same weights on converging states, so it injects nearly the same error, and these errors compound across loops into much larger end-to-end damage (Appendix~\ref{app:p3math}). Applying Quest's page dropping in all four loops costs 1.82\x{} the sum of its one-loop damages, whereas for plain models dropping in several layer groups together costs only 0.95--1.17\x{} the sum. At equal bytes, page dropping also does 4.7\x{} the damage of reading the same values at 2 bits (Figure~\ref{fig:p3}). Therefore, low-attention values should be kept and read at low precision rather than skipped. Because loops attend to different tokens, each reading loop should also choose for itself, at compute time, which values to read at full precision (C3).

\iffalse
\begin{figure}[t]
\centering
\includegraphics[width=\columnwidth]{figures/v4/fig_p3_example.pdf}
\vspace{-5mm}
\caption{Loops 3 and 4 read the same page of V at different depths. Twenty consecutive tokens of one page (layer 2, one head, one document); a reader fetches a token's low bits when its attention weight is at least $2^{-14}$ of the row maximum. The window is chosen so that the differences are easy to see; \S\ref{sec:props} gives the averages. %
}
\label{fig:depth}
\vspace{-3mm}
\end{figure}
\fi

\section{System Design}\label{sec:design}\label{sec:overview}
\sys{} works in two phases (Figure~\ref{fig:system}). Offline, once per model, the planner takes the model's profile (\S\ref{sec:characterize}) and budgets for damage and storage, and generates a plan that assigns each cell one storage option, i.e., 4-bit codes, 2-bit codes or the final loop's KV (\S\ref{sec:design-c1}, \S\ref{sec:design-c2}). Our vLLM plugin then turns the plan table into one descriptor per cell at load time. In the online phase, storage is fixed and one decision remains. \sys{} dynamically sets the precision at which each loop reads the values, based on the attention scores computed at run time (\S\ref{sec:design-c3}). One fused kernel then executes every descriptor (\S\ref{sec:design-kernel}).

\subsection{C1: Sensitivity-aware mixed-precision allocation}\label{sec:design-c1}
\paragraph{Profiling.}
For every cell $c{=}(t,\ell)$, profiling measures the single-cell damage $d_{c,o} = \mathrm{KL}(p_\text{exact}\,\|\,p_{c\leftarrow o})$, where we apply each of the three storage options $o$, to that cell alone, keep every other cell exact, and measure the KL on 16 of the model's own long generations in fp32 (\S\ref{sec:characterize}).
This profiling is cheap and predictive. In our experiments, summing the profiled damages of a plan's cells predicts the plan's measured KL with a correlation of 0.98 (\S\ref{sec:props})\footnote{On Ouro-1.4B, profiling covers 72 cells with three options each plus the 24 source cells in about 3 GPU-hours. The profile is also stable, i.e., two halves of the 16 documents rank the cells almost identically (rank correlation 0.985).}.

\paragraph{Allocation.}
Each cell takes one option $o$, which has stored bytes $s_o$, read bytes $r_o$ and damage $d_{c,o}$.
With KIVI-style codes \cite{liu2024kivi}, which quantize keys per channel and values per token in groups of 32 with an fp16 scale and zero point, own INT4 codes store and read 2.5\,KiB per token and layer and own INT2 codes 1.5\,KiB, metadata included (Appendix~\ref{app:bytes}).
With $x_{c,o}\in\{0,1\}$ and $\sum_o x_{c,o}=1$ for every cell, the allocation is
\begin{align}
\min_x\ & \textstyle\sum_{c,o} x_{c,o}\,r_o \nonumber\\
\text{s.t.}\ & \textstyle\sum_{c,o} x_{c,o}\,d_{c,o} \le \varepsilon,\quad \sum_{c,o} x_{c,o}\,s_o \le M,\label{eq:mckp}
\end{align}
a multiple-choice knapsack with two resource constraints \cite{kellerer2004knapsack}.
Because every option's byte cost is a multiple of 0.5\,KiB, a dynamic program over stored and read bytes solves this knapsack exactly over the 72 cells of loops 1--3. We keep loop 4, which every shared cell reads, at INT4, because coding it at INT2 alone would already cost 5.8\x{} \Pz{}. Its cost is therefore a constant in every plan, and $\varepsilon$ and $M$ bound the damage and the memory of the whole cache.

\begin{figure}[t]
\centering
\resizebox{\columnwidth}{!}{\input{figures/v4/fig_plans.tex}}
\vspace{-5mm}
\caption{The read-optimized plan beats the best per-loop allocation plan of \sys{}, i.e., INT4 in every loop with loop 3 sharing loop 4's KV, which has the lowest damage of all per-loop plans that store less than uniform INT4, on KL and throughput (42\% lower KL and 5\% higher throughput). The memory-optimized plan stores 14\% less than this plan and decodes 6\% faster. At equal bytes, the read- and memory-optimized plans have 0.36\x{} and 0.43\x{} the measured KL of the best per-loop plans, i.e., INT4 in loops 1--2 with INT2 in loop 3, and INT4 in loop 1, INT2 in loop 2 and sharing in loop 3.}
\label{fig:plans}
\vspace{-3mm}
\end{figure}
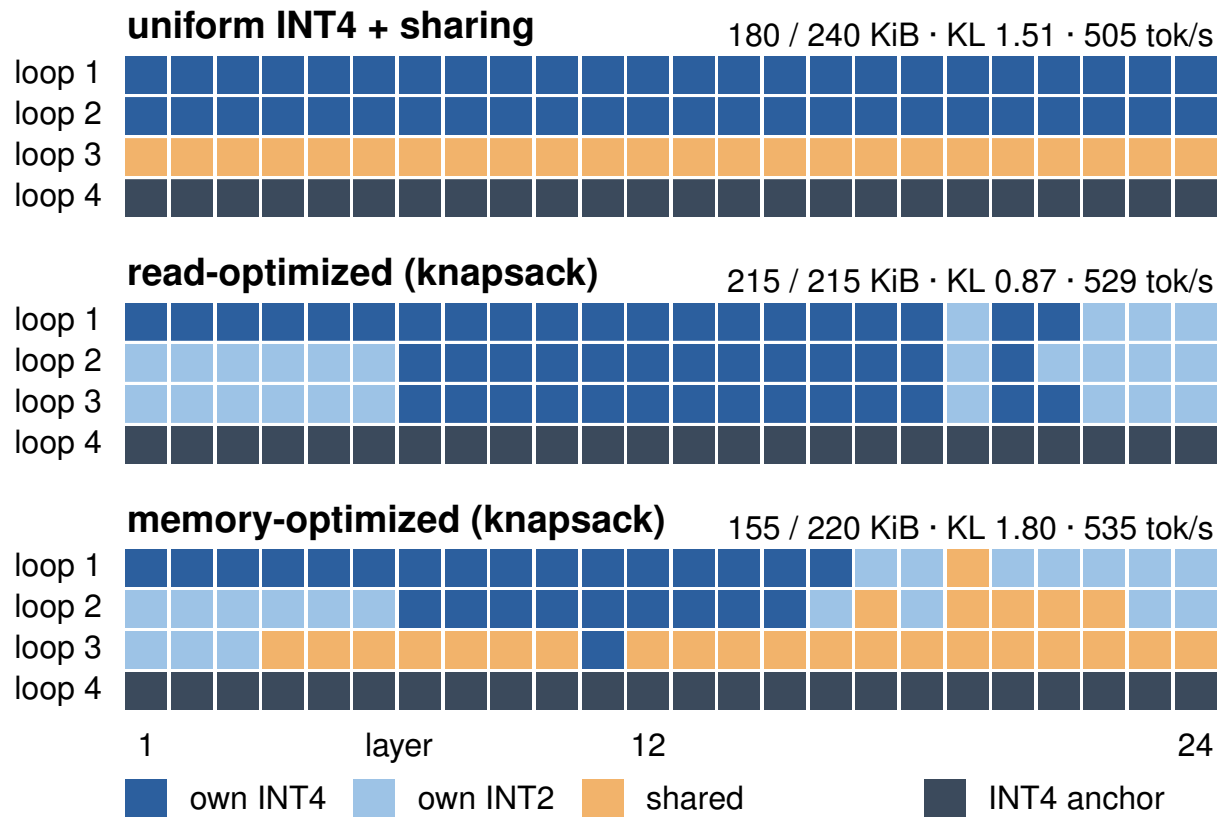

\paragraph{Two plans for the two ends of the frontier.}
The two budgets let us reach the two ends of the frontier with two plans. The read-optimized plan (\sys{}-R) minimizes reads under a damage budget of 1.19\x{} \Pz{}, with no storage budget. It keeps INT4 in the sensitive middle layers and uses INT2 in 25 of the 72 cells, all in the first or last six layers (Figure~\ref{fig:plans}). It shares no cell, because sharing saves storage but not reads. The memory-optimized plan (\sys{}-M) minimizes reads under a damage budget of 2.0\x{} \Pz{} and a storage budget of 156\,KiB per token, and it shares 26 cells, 20 of them in loop 3 as P2 predicts. Both budgets equal the damage of a residual-coding baseline\footnote{$\varepsilon$ and $M$ are parameters that a user can set as needed. Here we set them to the damage of residual-coding baselines (Table~\ref{tab:residual}) so that each plan can be compared with related work at equal damage. \sys{} itself stores no residuals.}. Both plans meet their targets, i.e., their measured KL stays below the prediction (Table~\ref{tab:additivity}), and they reach 6\x{} and 8\x{} BF16's goodput with no significant accuracy loss (Figure~\ref{fig:perf}).

\subsection{C2: Cross-loop KV sharing as a per-cell option}\label{sec:design-c2}
Sharing is a per-cell option because its damage, measured as KL, differs from cell to cell and from model to model (P2, Figure~\ref{fig:p2}).
A shared cell $(t,\ell)$ attends with its own query over the final loop's history at layer $\ell$. Its descriptor names the final loop's pool, so the cell does not store a copy of its own, while its sink and window stay its own and exact. When a 32-token page leaves the window, the cell releases its own copy, and its descriptor reads the final loop's page for those positions instead.
It stores nothing but reads the source's INT4 pages like an own INT4 cell, so the knapsack shares only when the storage budget binds.

\subsection{C3: Attention-aware V precision}\label{sec:design-c3}
P3 shows that only a few values receive high attention, about one history token in six, and that loops sharing a KV block attend to different tokens. V precision therefore lets every loop decide at read time which values to read at 4 bits and which at 2.

\paragraph{Bit-plane V layout.}
To support this design, we change the V layout. Keys, which are always read in full, stay one plane, while an INT4 cell stores its value codes $c\in\{0,\dots,15\}$ as two contiguous 2-bit planes, $\text{hi}{=}\lfloor c/4\rfloor$ and $\text{lo}{=}c \bmod 4$, followed by the scales and zero points. The cell therefore stores as many bytes as plain INT4, and skipping a token's lo plane saves 0.5\,KiB of reads per token and layer.

\paragraph{Read-time V precision at no extra read cost.} %
For each history page, the kernel reads all key codes first. When it reaches the values, it therefore already knows the score $s_j$ of every token for this query in this loop, and thus which values receive low attention before it reads them. The key read that attention performs anyway supplies these scores, with no predictor, extra pass or extra bytes. Methods that also skip keys must estimate importance before reading, e.g., from page bounds \cite{tang2024quest}, approximate scores \cite{ribar2024sparq,lee2024infinigen} or accumulated history \cite{zhang2023h2o}. \sys{} reads every key, so it decides from the scores attention computes and saves only V's lo plane.
Ideally, a token $j$ is \emph{cold} if its weight is below $2^{-\tau}$ times the largest in its row, i.e., $s_j < s_{\max} - \tau\ln 2$. Because $s_{\max}$ is costly to know before the row is complete, the kernel streams the row and tests $s_j < m - \tau\ln 2$ against a running maximum $m \le s_{\max}$ that starts at a seed $m_0$\footnote{The seed $m_0$ is the largest score over the row's 32 exact sink keys, which the kernel's near blocks compute anyway (\S\ref{sec:design-kernel}). As a lower bound on the row maximum, it only lowers the threshold, so the streaming rule never marks cold a token that the ideal rule keeps hot.}. A cold token reads only the hi plane and decodes to the midpoint of its bucket, $\hat c = 4\,\text{hi}+1.5$, while a hot token reads both planes and recovers its full INT4 code. We use $\tau{=}14$, which changes the fp32 KL of KIVI-4 on Ouro-1.4B by a factor of only 1.016, whereas lowering $\tau$ raises the cost quickly, e.g., $\tau{=}10$ already costs a factor of 1.35. Table~\ref{tab:tau} sweeps $\tau$ on the best per-loop allocation plan of \sys{}.

\begin{proposition}[Error bound]\label{prop:bound}
Let $a_j$ be the normalized attention weights of one row and head, $\mathcal C$ the set of tokens the seeded rule marks cold, and $\sigma_j$ the V scale of token $j$'s channel group. Then, relative to the INT4 reader that reads both V planes of every token, the attention output changes by at most
\[
\|\Delta o\|_\infty \;\le\; 1.5\sum_{j\in\mathcal C} a_j\,\sigma_j \;\le\; 1.5\,\sigma_{\max}\,|\mathcal C|\,2^{-\tau} a_{\max}.
\]
\end{proposition}
Appendix~\ref{app:extra} gives the proof.

\paragraph{A design opportunity specific to looped transformers.}
This rule applies to any INT4 cache, KIVI-4 included (\S\ref{sec:eval-tput}). In a looped transformer, however, each loop decides from its own attention scores when to read a value's lo plane. We therefore store every value at full INT4 once and decide at read time how much of it each loop reads. An alternative that fixes the precision in advance would have to keep every token that any reading loop marks hot at full precision. On Ouro-1.4B, for example, loops 3 and 4 mark different hot tokens, and only 83\% of the tokens are overlapped (Table~\ref{tab:overlap}). Across all four loops, such a fixed precision reads 1.6\x{} the lo-plane bytes of deciding at read time. In the two plans, V precision keeps 17--22\% of the history V reads per loop at full precision and cuts the reads per token by 13--14\%, from 215 to 186\,KiB for \sys{}-R and from 220 to 189\,KiB for \sys{}-M.

\subsection{The unified kernel}\label{sec:design-kernel}
\paragraph{Layout.}
A page holds 32 tokens of one (loop, layer) cell. An INT4 page stores the keys as interleaved 4-bit codes, and it stores the values as the two planes of different positions of the original V matrix \S\ref{sec:design-c3}. An INT2 page stores its own 2-bit codes. A shared cell stores no pages, and its descriptor points to the corresponding pages of the final loop instead. Sinks and windows stay in vLLM's BF16 blocks. There is one descriptor per cell, built when the plan is loaded and kept in device memory, and it carries all three of our designs, i.e., which pool to read (C2), how many bits are stored (C1) and whether to read the lo plane (C3).

\paragraph{One launch per cell.}
In our implementation, the CUDA launch grid has up to 8 \emph{near} thread blocks and up to 64 \emph{far} thread blocks per decode row (Algorithm~\ref{alg:kernel} in Appendix~\ref{app:extra}). The near blocks compute exact BF16 attention over the sink and window and publish their partial results and the sink maximum. Each far block reads its share of the history pages through the descriptor and uses the published maximum as the seed of V precision. To avoid a long wait, a far block uses the published maximum only if it arrives within 4\,$\mu$s and sets the seed to $-\infty$ otherwise, which would mark every token as a hot token, so the precision will not lose by a fake positive. The last block of a row merges all partial results by their log-sum-exp, which would give the exact softmax, so one launch replaces up to four per layer call (\S\ref{sec:impl}). Because the descriptors and every buffer the kernel reads have fixed addresses, the whole decode-only step, including the 96 attention launches of its four loops, also replays as one CUDA graph.
Window-exit quantization, whose page lists depend on the data, runs just before the replay.

\section{Implementation}\label{sec:impl}

\paragraph{vLLM integration.}
\sys{} is a plugin to vLLM 0.26 \cite{kwon2023vllm} and leaves the engine unmodified. It gives each loop its own KV-cache group, carves the history pools out of vLLM's KV budget, and turns the plan table into the 96 descriptors at load time. When a page leaves the window, a Triton KIVI codec \cite{tillet2019triton} quantizes it into its cell's pool and releases the BF16 block, so that more requests can be served at the same time.

\paragraph{History reader.}\label{sec:impl-kernel}
\sys{}'s CUDA history reader streams KV codes at 1.59--1.63\,TB/s, as fast as FlashAttention-2 streams BF16 on the same GPU\footnote{An RTX PRO 6000. The reader's techniques are described in Appendix~\ref{app:extra}.}. With V precision, V is stored as two planes, and the reader fetches a value's lo plane only for the tokens whose attention score is high enough. On real attention scores, this makes the reader 1.12--1.17\x{} faster than the plain INT4 reader\footnote{In a microbenchmark on an RTX PRO 4500.}.\looseness=-1

\begin{table}[t]
\centering\footnotesize
\caption{On every model and suite, no \sys{} average accuracy falls significantly below BF16 (best in bold). The other methods in the table include KIVI-4 + V, i.e., uniform 4-bit quantization plus \sys{}'s V precision, naive uniform INT2 quantization of loops 1--3, and naive last-loop KV reuse. R16K and R32K denote RULER-16K and RULER-32K \cite{hsieh2024ruler}, LB denotes LongBench v1 \cite{bai2024longbench}, and L5 denotes AIME + MATH level 5 \cite{hendrycks2021math}.}
\label{tab:acc}
\setlength{\tabcolsep}{2.4pt}\renewcommand{\arraystretch}{0.9}%
\newcommand{\bfc}{\cellcolor{bfblue}}\newcommand{\lkc}{\cellcolor{oursred}}
\resizebox{\columnwidth}{!}{%
\begin{tabular}{@{}l|cccccc@{}}
\toprule
 & R16K & R32K & LB & MATH500 & GSM8K & L5 \\
\midrule
\midrule
\multicolumn{7}{@{}l}{\textit{Ouro-1.4B}} \\
\bfc\textbf{BF16} & \bfc 71.44 & \bfc 54.04 & \bfc 25.29 & \bfc 84.5 & \bfc 76.6 & \bfc 70.6 \\
\lkc\textbf{LoopKV-M} & \lkc 72.12 & \lkc 53.63 & \lkc 25.19 & \lkc \textbf{86.0} & \lkc \textbf{77.1} & \lkc 71.1 \\
\lkc\textbf{LoopKV-R} & \lkc \textbf{72.49} & \lkc 53.86 & \lkc \textbf{25.32} & \lkc 84.5 & \lkc 76.6 & \lkc 70.3 \\
KIVI-4 + V & 71.37 & \textbf{54.57} & 25.25 & 84.5 & 76.9 & \textbf{72.0} \\
INT2, loops 1--3 & 65.50 & --- & 24.43 & \textbf{86.0} & 77.0 & 69.3 \\
Last-loop KV reuse & 4.14 & --- & 7.29 & 47.0 & 76.8 & --- \\
\midrule
\multicolumn{7}{@{}l}{\textit{Ouro-2.6B}} \\
\bfc\textbf{BF16} & \bfc 80.73 & \bfc --- & \bfc 33.70 & \bfc 89.5 & \bfc \textbf{79.7} & \bfc --- \\
\lkc\textbf{LoopKV-M} & \lkc 80.98 & \lkc --- & \lkc \textbf{34.74} & \lkc 89.5 & \lkc 79.5 & \lkc --- \\
\lkc\textbf{LoopKV-R} & \lkc \textbf{81.00} & \lkc --- & \lkc 34.17 & \lkc \textbf{90.5} & \lkc 79.2 & \lkc --- \\
KIVI-4 + V & 80.61 & --- & 34.07 & \textbf{90.5} & 79.5 & --- \\
\midrule
\multicolumn{7}{@{}l}{\textit{Dory-4B}} \\
\bfc\textbf{BF16 (last loop)} & \bfc 93.54 & \bfc --- & \bfc 32.43 & \bfc 91.5 & \bfc \textbf{88.6} & \bfc --- \\
\lkc\textbf{LoopKV-M} & \lkc \textbf{94.23} & \lkc --- & \lkc 32.37 & \lkc \textbf{91.7} & \lkc 88.2 & \lkc --- \\
\lkc\textbf{LoopKV-R} & \lkc 94.09 & \lkc --- & \lkc \textbf{32.62} & \lkc 91.5 & \lkc 87.9 & \lkc --- \\
\bottomrule
\end{tabular}}
\vspace{-3mm}
\end{table}
\paragraph{Fused kernel and whole-step graph.}
Without fusion, each cell runs up to four kernels, i.e., FlashAttention-2 \cite{dao2024flashattention2} over the sink and window, a seed kernel on INT4 cells, the history reader and a log-sum-exp merge, so a decode step issues 504 host launches in all. \sys{}'s fused kernel combines these four kernels into a single launch, which takes their GPU time within 2\% and cuts the launches to 337 (\sys{}-R) or 332 (\sys{}-M). It also makes it possible to replay each decode-only step as one CUDA graph, which cuts the launches further to 26 and 17. In our experiments, fusion has little effect on KL, i.e., served KL is 0.99\x{} that of the unfused path, and graph replay keeps it within 4\% of eager execution.\looseness=-1

\section{Evaluation}\label{sec:eval}

\subsection{Setup}\label{sec:eval-setup}
\paragraph{Models and hardware.}
We evaluate \sys{} on three looped models, Ouro-1.4B and Ouro-2.6B \cite{zhu2025ouro} and Dory-4B \cite{nvidia2026dory}, and add Huginn-3.5B \cite{geiping2025huginn} for the planning results in fp32 (\S\ref{sec:eval-general}). For Ouro, we use the base checkpoint for retrieval and GSM8K and the Thinking checkpoint, on which the plans are profiled, for everything else. We serve with vLLM 0.26.0 (Dory with its own vLLM fork) on NVIDIA RTX PRO 6000, RTX PRO 4500 and A100-80 GPUs, and each table and figure names the GPU it uses (configuration in Appendix~\ref{app:tputslo}). Throughput is measured on the full system, with our fused kernel and the whole-step graph. We use RULER-16K, LongBench, MATH500 and GSM8K to measure serving quality, and we also use RULER-32K and long reasoning (AIME + MATH level 5) to evaluate long-context and reasoning abilities.\looseness=-1
\paragraph{Quality.}
Served tasks are paired against the served BF16 model with 95\% bootstrap CIs.
The public suites are official RULER \cite{hsieh2024ruler} (13 tasks of 100 items, at 16K and 32K), the 16 English and code subsets of LongBench v1 \cite{bai2024longbench}, MATH500 \cite{hendrycks2021math,lightman2024verify} (first 200 items, greedy, 8K output cap), GSM8K \cite{cobbe2021gsm8k} (1,319 items, 3-shot) and sampled long reasoning on AIME and MATH level-5 problems.
Appendix~\ref{app:quality} adds two stress tests at 16K and details long reasoning.
No plan was selected on any task, since the plans come from the fp32 KL knapsack (Appendix~\ref{app:quality} checks its documents' overlap with MATH500).
We use fp32 KL (\S\ref{sec:characterize}) as the sensitivity measure because KL and answer flips expose changes that accuracy averages hide \cite{dutta2024accuracy}, and non-inferiority means that a suite average's lower CI bound lies above $-3$\pp{}.
\paragraph{Throughput and SLO.}
We measure peak decode throughput by sending $B$ requests at context length $N\in\{8\text{K},16\text{K},32\text{K}\}$ to the vLLM server, which decode 128 tokens together, and report the best throughput over the values of $B$ at which no request is queued.
We evaluate the SLO workload by sending 60 prompts from AIME 2024 (the American Invitational Mathematics Examination)~\cite{maa2024aime} to the Thinking checkpoint at Poisson arrivals, the standard model of independent user requests in LLM serving~\cite{kwon2023vllm,zhong2024distserve}.
A request meets the SLO if its time to first token (TTFT) is at most 10\,s and its mean time per output token (TPOT) is at most $3\tau_0$.\footnote{$\tau_0$ is the decode step time of stock vLLM on the RTX PRO 6000 predicted by a linear fit of step time to context length (17.31\,ms plus 0.495\,ms per 1K tokens). The fit gives $\tau_0 = 18.60$\,ms or 20.69\,ms under two rules for the context length, i.e.\ $3\tau_0 = 55.8$\,ms or 62.1\,ms, and we report both.}
The goodput~\cite{zhong2024distserve} is therefore the highest tested arrival rate at which 95\% of requests meet the SLO.
\paragraph{Baselines and ablations.}
We compare with BF16, vLLM's FP8 cache (FlashInfer \cite{ye2025flashinfer}), KIVI-4 and KIVI-2 (uniform INT4 and INT2 on all loops, groups of 32), FlashLoop (official code and our reproduction) and ResidualQuant (reimplemented).
To the best of our knowledge, FlashLoop and ResidualQuant have no vLLM implementation, so we compare them on quality and bytes stored for one token.
Cross-loop KV sharing itself is adopted by some of the models: Dory is trained to read the last loop's KV in every loop \cite{nvidia2026dory}, and Mixture-of-Recursions shares KV across recursions \cite{bae2025mor}.
We therefore also compare with BF16 + loop-3 sharing, FP8 + loop-3 sharing and the hand-made \planU{}, in which loop 3 reads loop 4's KV, the placement that P2 finds cheap. Configurations that use our allocation (C1) or V precision (C3) are ablations.
KIVI-4 runs on the same fused kernel and whole-step graph as \sys{} and is our main baseline: it is the strongest prior method we serve, and the gains we report are over it. KIVI-4 + V adds our V precision (C3) to it and is the ablation step before allocation (C1) and sharing (C2).
On our bit-plane INT4 pages (\S\ref{sec:design-c3}), KIVI-4 reads both V planes of every token, as KIVI does.

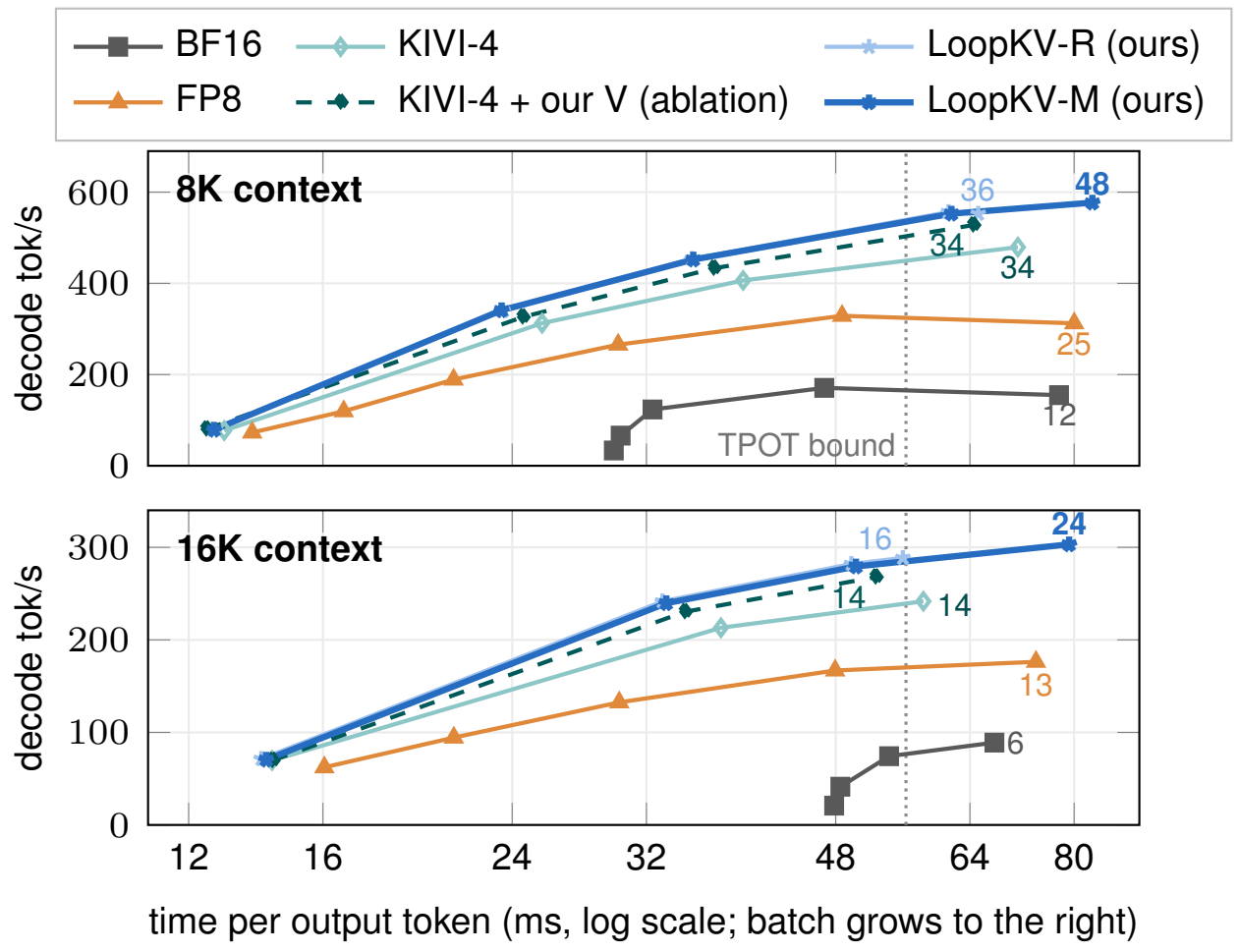
\begin{figure}[t]
\centering
\input{figures/v4/fig_tlat_col.tex}
\vspace{-2mm}
\caption{\sys{} keeps more requests resident: in this sweep it matches or beats the other configurations at each time per token and keeps scaling after they run out of KV memory. Ouro-1.4B on one RTX PRO 6000: decode tok/s against time per output token as the batch grows, up to each configuration's largest resident batch (numbers); INT4 configurations on the fused kernel and whole-step graph, BF16 and FP8 as vLLM serves them; time per token on a log scale; the dotted line marks the SLO's time-per-token bound ($3\tau_0$, \S\ref{sec:eval-setup}). KIVI-4 + our V is an ablation.}
\label{fig:tlat}
\vspace{-2mm}
\end{figure}

\subsection{Quality}\label{sec:eval-quality}
\textbf{No \sys{} plan loses accuracy significantly on any public suite average} (Table~\ref{tab:acc}). %
On Ouro-1.4B, \sys{}-M is non-inferior to BF16 on all six public suites under a 3-point margin and \sys{}-R on five, its MATH500 interval reaching $-4.0$ points on 200 items (Appendix~\ref{app:quality} lists every interval).
\sys{}-R even gains on official RULER-16K by about one point, a change carried by a single task.

\sys{}-M stores 35\% less than KIVI-4 at about twice the fp32 KL of \sys{}-R, yet it sustained the accuracy over the final scores of the 6 public suites.
\sys{}-R, which shares no cell, keeps variable tracking, with no RULER-32K task changing significantly, and still sustains 1.3--1.6\x{} KIVI-4's goodput (\S\ref{sec:eval-slo}); it is the plan to serve when long-context tracking matters.
KIVI-4 + V and FP8 + loop-3 sharing behave the same way, staying within noise on the RULER, LongBench, MATH500 and GSM8K averages, although each loses variable tracking on RULER-16K.
Allocation alone leaves answers as stable as uniform quantization does: \sys{}-R changes about as many RULER-16K answers as KIVI-4 + V (Table~\ref{tab:flips}).
FlashLoop's official code, in contrast, loses 12.75 points against its dense path on the same RULER-16K items in a run on an H100, mostly on variable tracking and multi-key needles. %
Its residual codec alone loses nothing in our earlier reproduction (Appendix~\ref{app:quality}), which points to its sparse, lazily updated attention as the cause.
\textbf{Allocation buys quality where bytes are scarce.}
Uniform INT2 in loops 1--3 stores 8\% more than \sys{}-M (168 vs 155\,KiB) yet loses 5.9 points on RULER-16K, where \sys{}-M stays within noise (Table~\ref{tab:acc}); a hand-made plan of \sys{}-M's storage has 2.3\x{} its fp32 KL (4.21 vs 1.80; Table~\ref{tab:additivity}).
At KIVI-4's own bytes, the knapsack's minimum-damage plan only matches KIVI-4's KL (0.47 vs 0.46; Table~\ref{tab:additivity}).\looseness=-1

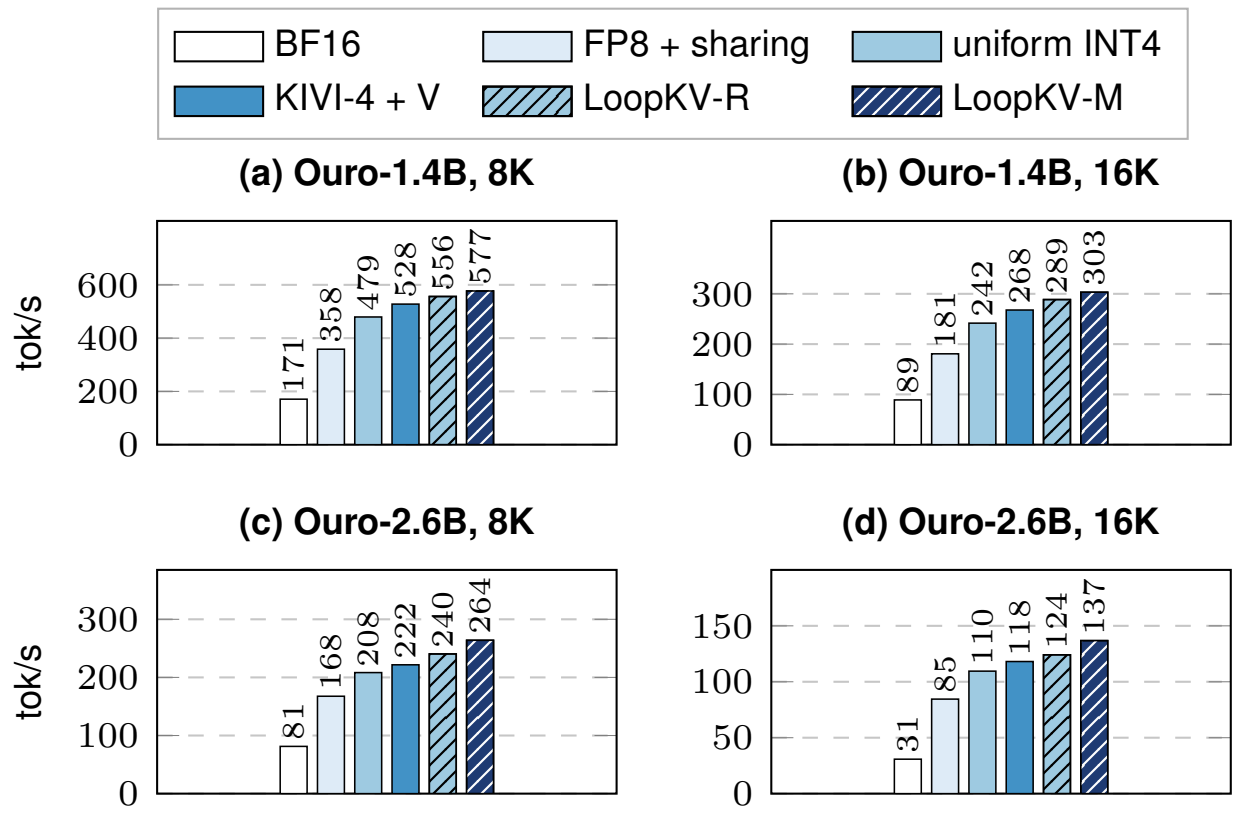
\begin{figure}[t]
\centering
\resizebox{\columnwidth}{!}{\input{figures/v4/fig_tput.tex}}
\vspace{-6mm}
\caption{\sys{}'s storage saving becomes throughput where memory binds. Peak decode tok/s at the resident peak for Ouro-1.4B (a, b) and Ouro-2.6B (c, d) on an RTX PRO 6000, each GPU's arms in one session, fused kernel and graph for INT4 arms; Ouro-1.4B within 1.1\% of the separate runs quoted in the text (repeat means: \S\ref{sec:eval-tput}); 32K and a 32\,GB GPU: Tables~\ref{tab:tput}, \ref{tab:tputx} and~\ref{tab:smallcard}; Dory-4B: Table~\ref{tab:doryplan}.}
\label{fig:tput}
\vspace{-2mm}
\end{figure}

\subsection{Throughput}\label{sec:eval-tput}
\sys{}-M decodes 575.9 and 301.9\,tok/s at 8K and 16K, 3.37\x{} and 3.39\x{} BF16, while storing 46\% less KV than FP8 with sharing.
\sys{}-R reaches 550.2\,tok/s at 8K, 3.22\x{} BF16 (all peaks: Figure~\ref{fig:tput} and Table~\ref{tab:tput} in Appendix~\ref{app:tputslo}).
\textbf{\sys{}-M decodes 1.20\x{} as fast as KIVI-4 at 8K: V precision gives about half of the gain, and allocation and sharing the rest.}
On the fused path on one RTX PRO 6000, V precision, which needs no plan, lifts uniform INT4 by 10--12\% at 1.06\x{} its served KL (Figure~\ref{fig:tput}a, b).

At the resident peak, allocation alone adds 5.6--7.6\% to KIVI-4 + V (\sys{}-R) and allocation with sharing 8.4\% at 8K, 12.5\% at 16K and 15.6\% at 32K (\sys{}-M; means of three repeats at 8K and 16K, one run at 32K, Table~\ref{tab:tput}).
The gains keep their shape on a second GPU type: on an A100-80, \sys{}-M decodes 2.2\x{}, 2.2\x{} and 4.7\x{} faster than BF16 at 8K, 16K and 32K and 1.21--1.30\x{} faster than uniform INT4 (Table~\ref{tab:tputx}).
The mechanism is residency (Figure~\ref{fig:tlat}), since at KIVI-4 + V's peak batch the plans gain 4.3--6.1\% from fewer reads and the rest from the requests that the saved storage keeps resident; at 16K \sys{}-M peaks at batch 24 against 14, and at a matched time per token both plans remain faster than KIVI-4 + V (Appendix~\ref{app:tputslo}).

\textbf{Where memory binds, the gain grows further.}
On a 32\,GB RTX PRO 4500, \sys{}-M keeps 4 requests resident at 32K against uniform INT4's 2 and decodes 29\% faster than KIVI-4 + V, against 3--4\% at equal batch (Table~\ref{tab:smallcard} in Appendix~\ref{app:smallcard}).\looseness=-1

\paragraph{On an H100.} A port of the fused kernel to an H100 SXM (80\,GB; tensor-core key path; stored codes byte-identical and served KL within 2\% of ours) keeps the ordering: \sys{}-M decodes 2.69\x{} as fast as BF16 at 8K and at 16K, against 2.27\x{} and 1.96\x{} for KIVI-4 and 2.08\x{} and 2.05\x{} for FP8, and served MATH500 and GSM8K stay within noise of BF16 for every INT4 arm (Table~\ref{tab:h100}). On this GPU, V precision adds no speed: the H100 reader is bound by latency and instructions rather than bandwidth, and a variant that fetches the low plane only for attended tokens reads 17\% fewer bytes per page but runs no faster.

\subsection{SLO capacity (goodput)}\label{sec:eval-slo}
\textbf{\sys{}-M sustains 1.7--2.2\x{} the SLO capacity of KIVI-4 on the same GPU.}
On the second RTX PRO 6000, which ran every SLO arm, \sys{}-M sustains 2.00\x{}, 1.72\x{} and 2.20\x{} KIVI-4's rate in the three runs at $\tau_0{=}18.60$\,ms and 1.80--2.00\x{} at 20.69\,ms (Figure~\ref{fig:slocap}, Table~\ref{tab:sloladder}); over the ablation KIVI-4 + V, which adds our V precision, the ratios are 1.60\x{}, 1.72\x{} and 1.83\x{}.
Again the mechanism is residency. Wherever both arms ran the same trace at the same rate, only KIVI-4 + V failed; at 0.0714\,req/s it preempts 100 times and misses the SLO for 11.6\% of requests on time to first token, while \sys{}-M preempts nothing.
The ratio to KIVI-4 + V holds under resampling and grows as the time-per-token bound loosens: at $\tau_0{=}18.60$\,ms it stays at 1.5\x{} or more in at least 86\% of 10,000 bootstrap draws in every run, and with time per token relaxed to $5\tau_0$ it rises to at least 1.80\x{} in every run (Appendix~\ref{app:tputslo}).
\sys{}-R sustains 1.3--1.6\x{} KIVI-4's rate at 18.60\,ms, and on an A100-80 the plans sustain 1.20--1.56\x{} uniform INT4's (Appendix~\ref{app:tputslo}). FP8 and BF16, as vLLM serves them with piecewise graphs, are limited by time per token, and \sys{}-M sustains 2.5\x{} and 8.0\x{} their rates (2.8\x{} and 7.5\x{} at 20.69\,ms); FP8 + loop-3 sharing sustains less than plain FP8 at 18.60\,ms (3.1\x{} below \sys{}-M), because its pointer path costs time per token.
V precision alone moves uniform INT4's capacity by one rate step or none.\looseness=-1

\subsection{The quality--throughput frontier}\label{sec:eval-pareto}
KIVI-4, KIVI-4 + V, \sys{}-R and \sys{}-M form the measured frontier from 0.46\x{} to 1.8\x{} \Pz{}, and the per-layer optimized baseline \planU{} and every residual design lie behind it (Figure~\ref{fig:paretofull}; Figure~\ref{fig:pareto} shows one point per comparable method and omits KIVI-4 + V). %
\looseness=-1

\subsection{Ablation}\label{sec:eval-ablation}
Each contribution buys a different resource at a stated KL (Table~\ref{tab:ablation}): allocation (C1) cuts reads by 10\% while KL rises from 0.46\x{} to 0.87\x{} \Pz{}, sharing (C2), with a larger damage budget, cuts storage by 28\% and raises the resident batch from 36 to 48, and V precision (C3) cuts reads by 14\% at unchanged KL.
Fusion and the graph buy latency more than peak throughput, cutting host launches from 504 to 17 for \sys{}-M and speeding up batch-1 decoding of the two plans by 1.07--1.19\x{} at 8K and 16K, against 0.5--2.5\% at the peak.\looseness=-1\relax{}
Sharing alone buys only part of the capacity: with V precision on the same fused path, the hand-made \planU{} (180\,KiB stored per token, against \sys{}-M's 155) sustains 0.75\x{} \sys{}-M's 16K SLO capacity, the same as \sys{}-R.

\paragraph{Comparison with codec baselines.}\label{sec:eval-codecs}
No other codec reading less than 216\,KiB, TurboQuant-3.5 \cite{zandieh2025turboquant} and KVQuant-3b \cite{hooper2024kvquant} included, reaches the \planR{}'s KL in our implementations (Figure~\ref{fig:codecs} in Appendix~\ref{app:codecs}).
The loop axis matters as well: restricting the knapsack to one option per layer, as layer-wise allocators do \cite{li2025kvtuner}, raises the KL at equal bytes by 1.52\x{} at the \planR{}'s bytes and by 1.43\x{} at the \planM{}'s bytes with loop 3 shared (Appendix~\ref{app:additivity}).

\subsection{Generality: Ouro-2.6B, Dory and Huginn}\label{sec:eval-general}
On Ouro-2.6B, loop 1 again dominates, damages add up, and the \planM{} computed from its own profile has 0.48\x{} the KL of the hand-made plan of equal bytes (Appendix~\ref{app:general}).
Served by us on the RTX PRO 6000, Ouro-2.6B's \sys{}-M decodes 3.25\x{}, 4.44\x{} and 11.0\x{} faster than BF16 at 8K, 16K and 32K, because BF16 holds only 6, 3 and 1 requests (Table~\ref{tab:tputx} adds an A100-80).
No Ouro-2.6B arm loses accuracy significantly on any suite average (Table~\ref{tab:acc}).
On Dory-4B and Huginn-3.5B, where the last loop carries 79--93\% of the INT2 damage (P1), the knapsack at 10\% fewer bytes than uniform INT4 has 1.33\x{} uniform INT4's KL on Dory, against 2.00\x{} for INT2 in the first loops and 29.3\x{} in the last; Huginn shows the same order against the first loops (Table~\ref{tab:general}).
Dory's trained last-loop mode, in which all loops read loop 4's KV, confirms the cross-loop redundancy that \sys{} exploits post hoc.
\sys{}'s descriptor path and fused reader serve this cache with per-plane plans from the same knapsack, which pass the in-serving byte checks and show no significant loss on any suite average (Table~\ref{tab:acc}); on a 32\,GB GPU, \sys{}-M decodes 3\% and 13\% faster than Dory's BF16 cache at 8K and 16K, and the plans hold their KL budget, whereas INT4 on all nine full-attention planes buys further speed with a 35--52\% higher served KL (Appendix~\ref{app:dory}).\looseness=-1

\section{Related Work}\label{sec:related}

\paragraph{Looped models and cross-loop compression.}
In looped models, later loops refine earlier ones \cite{blayney2026mechanistic}.
Several designs train the model so that loops share their KV: MELT keeps one gated cache per layer \cite{melt2026}, the Parallel Loop Transformer reuses the first loop's KV in every loop \cite{wu2025plt}, and pretraining in which only the first recursion writes KV cuts context memory by 76--79\% \cite{monea2026shared}; these change the model, whereas \sys{} serves the released model.
Looped latent attention keeps the model frozen and learns a low-rank cross-loop codec by distillation \cite{oneill2026lla}; \sys{} needs no training, only a profile.
Without retraining, Huginn reads and writes a fixed number of cache slots across recurrence steps \cite{geiping2025huginn}, and a parallel sampler for it matches GSM8K accuracy with a single slot \cite{geiping2026parallel}; Huginn was trained so that keys and values from different depths are compatible, whereas on Ouro such reuse loses accuracy over long outputs (P2), and WaveFront Decoding, a self-speculative decoder for looped models, keeps Ouro's per-recurrence cache for the same reason \cite{ha2026wavefront}.
FlashLoop introduced closed-loop residual coding of cross-loop KV with lazily updated sparse attention in a batch-1 research harness \cite{yang2026flashloop}, and ResidualQuant stores 2-bit residuals against an INT4 final loop, a fixed loop-wise precision rule, and reports up to 4.15\x{} higher peak throughput from larger batches on an RTX 5090 with its own kernels \cite{kim2026residualquant}; we reimplement its codec in our harness.
Building on their observation that adjacent loops are similar, \sys{} measures where this pays (P1, P2) and what a looped step reads (P3), serving the released model unchanged.\looseness=-1

\paragraph{KV quantization, sharing and eviction.}
Uniform-bit codecs \cite{liu2024kivi,hooper2024kvquant,ashkboos2024quarot,zandieh2025turboquant,zhou2026oscar,lin2025qserve} and rate--distortion allocation \cite{zhang2026rdkv} follow sensitivity-based mixed precision \cite{dong2019hawq}, and KVTuner \cite{li2025kvtuner} and AutoQuant-KV, which picks one K/V format per layer by KL sensitivity under an average-bit budget \cite{nvidia2026autoquantkv}, allocate over layers; \sys{} adds the loop axis and a sharing option, so its knapsack separates stored from read bytes.
MiKV \cite{yang2024mikv} and ZipCache \cite{he2024zipcache} set token precision at store time by importance heuristics, and DiffKV \cite{zhang2025diffkv} treats keys, values, tokens and heads differently.
SpAtten fetches low-order bits and recomputes only when attention from the high-order bits is flat \cite{wang2021spatten}, and Any-Precision LLM and Matryoshka Quantization nest lower precisions in a weight code's high-order bits \cite{park2024anyprecision,nair2025matryoshka}; \sys{} stores INT4 values as two such planes, so a cold token reads only the high plane, and decides per loop and token from the exact scores its fused kernel already computes.
AlignedKV also chooses how many bits of each value to read from the scores of the same attention computation, to cut memory access to an FP16 cache \cite{tan2024alignedkv}, and QuantSpec stores INT8 KV as two INT4 parts so that a draft model reads the upper part and the target reads both \cite{tiwari2025quantspec}; in \sys{} the readers of one INT4 value are the loops of one model.
Concurrent ReadKV also reads key and value prefixes of a progressive code according to the query and its attention \cite{bhandari2026readkv}; \sys{} applies the value side per loop, inside the fused kernel that serves the knapsack's plan.
Cross-layer sharing, trained \cite{brandon2024cla,sun2024yoco,wu2024lckv} or post hoc \cite{liu2024minicache,chang2025xkv,shutova2025aquakv,yang2024kvsharer,qiao2024swiftkv}, aliases layers; \sys{} aliases loops, cell by cell.
Eviction and page selection \cite{xiao2024streamingllm,zhang2023h2o,li2024snapkv,ge2024fastgen,tang2024quest,li2025scbench} drop tokens, while \sys{} keeps StreamingLLM's sink and window, evicts nothing and is evaluated in the engine \cite{gao2025rethinking}.\looseness=-1

\section{Conclusion}\label{sec:conclusion}
\sys{} turns three properties of Ouro's looped KV into allocation, sharing and V precision, executed by one fused kernel.
On Ouro-1.4B, on the same kernel and graph as uniform INT4 (KIVI-4), \sys{}-M stores 35\% less KV and sustains 1.7--2.2\x{} KIVI-4's goodput under a 16K latency SLO across three runs, and \sys{}-R sustains 1.3--1.6\x{}.
No public suite average of either plan falls significantly below BF16's.
The lesson is that a looped model's KV is a grid, not a stack, in which a cheap profile and an exact knapsack decide once per model where each cell spends its bits and whose KV it reads.\looseness=-1

We will release the plugin, the kernels, the plans and the profiles.

\paragraph{Limitations.}
\sys{} targets looped models that keep a KV cache per loop: one such family, Ouro, is served in full; as transfer evidence, Dory-4B is served in its trained last-loop mode and Huginn-3.5B is evaluated in fp32. %

We measure speed on RTX PRO 6000, RTX PRO 4500 and A100-80 GPUs, and SLO capacity on one decode-heavy workload (60 AIME prompts at 16K, up to three runs). %
On RULER-32K, \sys{}-M loses variable tracking ($-21.2$ points, 95\% CI $[-29.6,\,-12.8]$) while its average stays within noise; \sys{}-R keeps it. %

\bibliographystyle{mlsys2025}
\bibliography{references}

\clearpage
\appendix
\onecolumn

\section{Full quality results}\label{app:quality}
Table~\ref{tab:final} gives the $\Delta$s of the suites run on the fused path, Table~\ref{tab:quality} those of the runs before fusion, including the in-house stress tests MQ641 and RULER-style (\S\ref{sec:eval-setup}), and Table~\ref{tab:ouro26served} those of Ouro-2.6B served; the remaining cells of Table~\ref{tab:acc} follow below.
On the 300-item RULER-style set, the intervals are too wide for any quantized configuration to establish the 3-point margin.
MQ641 has 641 multi-key items with seven-digit values at 16K in public PG19 books; its pre-registration was written before any run, its generator, seeds and hashes are released, its CIs bootstrap over about 131 book clusters, and an independent run reproduced it on another GPU (BF16 85.96\%, KIVI-4 87.05, the two plans 86.27--86.43, the hand-made sharing plans 81.75--84.09).
No plan was selected on it.
In the independent runs, the \planM{} scores $+4.7$ points above the hand-made INT4/INT2 + sharing plan of equal storage.

\paragraph{Protocols.}
Long reasoning samples 194 AIME and MATH level-5 problems at temperature 1.0 and top-p 0.7, 6 times per AIME and 8 times per MATH problem, with a 16,384-token cap; its accuracy is the mean of the items' sample accuracies, with CIs paired by item.
In LongBench, narrativeqa, musique and passage\_count are at the BF16 floor.
The 16 documents of the fp32 KL profile are generations for 16 of the 200 MATH500 prompts, and removing them moves no pre-fusion MATH500 delta by more than 0.4 points.
On the fused path, MATH500 caps each INT4 arm's concurrency at its pool's capacity, because without the cap the pages that overflow the pool stay in BF16 (a pool audit counted 15,008 for \sys{}-R); uncapped, each of these arms scored $+2.5$ %
, and the MATH500 runs of Table~\ref{tab:quality} predate the cap.

\paragraph{Remaining cells of Table~\ref{tab:acc}.}
On the fused path, long reasoning (AIME + MATH-L5) changes by $-0.26$ for \sys{}-R and $+0.54$ for \sys{}-M; before fusion it changed by $+1.14$ and $-0.43$ (Table~\ref{tab:quality}), and an independent run of that version on H100 and H200 GPUs with the same protocol gives $+1.25$ and $-0.62$.
On the A100-80, against BF16 on the same GPU, long reasoning changes by $+1.10$ for KIVI-4 + V and by $-0.13$ for FP8 + loop-3 sharing.
On Ouro-2.6B, RULER-16K changes by $-0.12$ for KIVI-4 + V, $+0.27$ for \sys{}-R and $+0.25$ for \sys{}-M, and LongBench by $+0.37$, $+0.47$ and $+1.05$.
On Dory-4B, against its last-loop BF16 cache, \sys{}-M and \sys{}-R change RULER-16K by $+0.69$ and $+0.55$, LongBench by $-0.06$ and $+0.19$, and GSM8K by $-0.5$ and $-0.7$; their MATH500 results are in Table~\ref{tab:doryplan}.

\paragraph{Replication on another RTX PRO 6000.}
An independent run on an RTX PRO 6000 Server Edition repeats the final-path quality suites against its own BF16. It passes the same gate first: decode KL within 2\% of ours and no differing byte. Every comparison stays within noise. On MATH500-200 and GSM8K, the differences for KIVI-4 + V, \sys{}-R, \sys{}-M and FP8 + loop-3 sharing lie between $-1.5$ and $+2.5$ points, and LongBench for \sys{}-R is $+0.00$.
On RULER-32K, \sys{}-M scores $-0.78$ and again loses variable tracking ($-18.0$) while gaining frequent-word extraction ($+15.7$), whereas KIVI-4 + V scores $+0.62$ and keeps variable tracking ($-0.4$).
The variable-tracking loss therefore follows loop-3 sharing rather than V precision or the codec. FP8 + loop-3 sharing loses it too on an A100-80 (Table~\ref{tab:final}), whereas KIVI-4 + V, which shares nothing, keeps variable tracking at 32K, and its frequent-word gain ($+5.7$) is about a third as large.

\begin{table}[h]
\centering\small
\caption{Quality before kernel fusion, including the in-house stress tests MQ641 and RULER-style: neither knapsack plan loses significantly against BF16; read-optimized + V meets the 3-point margin on MQ641, MATH500, GSM8K and long reasoning but not on RULER-style. fp32 KL / \Pz{} (\Pz{} $=1.20{\times}10^{-4}$; 16 documents); tasks: $\Delta$ (pp) vs served BF16, $^+$\,/\,$^-$ where the paired 95\% CI lies above / below zero (single runs; greedy except long reasoning). ``---'' = not run. $^\dagger$\,Earlier RTX PRO 4500 harness, vs its own BF16. $^\ddagger$\,Codec only, no sparse path. FP8 / FP8 + loop-3 sharing: fp32 KL 0.19\x{} / 1.20\x{}, no runs here.}
\label{tab:quality}
\setlength{\tabcolsep}{4pt}
\begin{tabular}{@{}llrrrrrr@{}}
\toprule
 & plan & \makecell{fp32\\KL} & \makecell{MQ641\\(stress)} & \makecell{RULER-style\\16K} & \makecell{MATH500\\(200)} & \makecell{GSM8K\\(1,319)} & \makecell{long reasoning\\(194)} \\
\midrule
\multirow{4}{*}{\sys{}} & read-optimized & 0.87 & $+0.8$ & $-1.6$ & --- & --- & --- \\
 & \quad + V precision & 0.87 & $+0.0$ & $-1.1$ & $+3.0$ & $-0.5$ & $+1.14$ \\
 & memory-optimized & 1.80 & $-0.8$ & $-0.2$ & $+2.0$ & $+0.8$ & --- \\
 & \quad + V precision & 1.80 & $+0.3$ & --- & $+1.5$ & --- & $-0.43$ \\
\midrule
\multirow{6}{*}{baselines} & KIVI-4 & 0.46 & $+0.9$ & $-1.7$ & --- & --- & --- \\
 & KIVI-2 & 21.5 & $-13$ to $-18^\dagger$ & --- & --- & --- & --- \\
 & residual, INT4 final + INT4 & 1.42 & $-0.8$ & $-2.0$ & --- & --- & --- \\
 & residual, INT4 final + INT2 & 4.38 & $-3.0^{-}$ & --- & --- & --- & --- \\
 & FlashLoop (our reproduction) & 0.30$^\ddagger$ & $-6.4^\dagger$ & $-8.2^{-}$$^\dagger$ & --- & --- & --- \\
 & ResidualQuant (reimpl.) & 2.01 & $-0.8^\dagger$ & niah\_mq $-9.0^{-}$$^\dagger$ & --- & --- & --- \\
\midrule
\multirow{3}{*}{+ sharing} & KIVI-4 + loop-3 sharing (hand-made) & 1.51 & $-1.6$ & $-0.2$ & $+2.5$ & $+0.2$ & --- \\
 & \quad + our V precision (ablation) & 1.49 & $-1.2$ & --- & --- & --- & --- \\
 & BF16 + loop-3 sharing (\Pz{}) & 1.00 & $-1.6$ & $+0.8$ & $+0.41$$^\dagger$ & $+0.3$$^\dagger$ & $-0.45$ \\
\bottomrule
\end{tabular}
\end{table}

In our earlier reproduction, FlashLoop's residual codec alone costs 0.0 points on MQ641 and its sparse, lazily updated attention alone $-7.8$ (full method $-6.4$).

\begin{table}[h]
\centering\small
\caption{Quality on the fused path (fused kernel, whole-step graph, V-precision seed $m_0$ as served): $\Delta$ (pp) vs served BF16, $^+$\,/\,$^-$ where the paired 95\% CI lies above / below zero; tasks listed if significant (unadjusted); $^*$\,BF16-floor task. FP8 + loop-3 sharing runs as vLLM serves it (no \sys{} kernel); its RULER-32K run is on an A100-80 against BF16 on the same GPU. RULER tasks: vt variable tracking, fwe frequent-words extraction, cwe common-words extraction, qa\_1 single-hop QA; LongBench subsets by name.}
\label{tab:final}
\setlength{\tabcolsep}{4pt}
\begin{tabular}{@{}>{\raggedright\arraybackslash}p{2.3cm}>{\raggedright\arraybackslash}p{2.6cm}>{\raggedright\arraybackslash}p{2.8cm}>{\raggedright\arraybackslash}p{3.4cm}>{\raggedright\arraybackslash}p{1.3cm}>{\raggedright\arraybackslash}p{1.3cm}>{\raggedright\arraybackslash}p{1.3cm}@{}}
\toprule
arm & official RULER-16K & official RULER-32K & LongBench v1 (16) & MATH500 (200) & GSM8K (1,319) & MQ641 (stress) \\
\midrule
KIVI-4 + V & $-0.07$; vt $-3.0$ & $+0.53$ & $-0.04$; qasper $-0.87$, gov\_report $-0.64$ & $+0.0$ & $+0.3$ & $+0.5$ \\
\sys{}-R & $+1.05^{+}$; qa\_1 $+7.0$ & $-0.18$ & $+0.03$; multifieldqa\_en $+0.93$ & $+0.0$ & $+0.0$ & $+0.6$ \\
\sys{}-M & $+0.68$; fwe $+5.0$, qa\_1 $+8.0$ & $-0.41$; vt $-21.2$, fwe $+18.3$, cwe $+0.6^*$ & $-0.10$ & $+1.5$ & $+0.5$ & $+0.5$ \\
FP8 + loop-3 sharing (as served) & $-0.22$; vt $-6.8$, cwe $+1.5$ & $+0.29$ (A100-80); vt $-14.2$, fwe $+19.7$ & $+0.07$; narrativeqa $-0.44$, musique $+1.64$$^*$ & $+2.0$ & $-1.1$ & --- \\
\bottomrule
\end{tabular}
\end{table}

\paragraph{Answer flips.}
Following \citet{dutta2024accuracy}, Table~\ref{tab:flips} counts items whose score changed against BF16 on the same items.
On RULER-16K \sys{}-R changes about as many items as KIVI-4 + V (70 vs 67 of 1,300), \sys{}-M 125 and FP8 + loop-3 sharing 146.
Before fusion, \sys{}-M also lost two RULER-16K needle tasks (niah\_multikey\_2 $-7.0$, niah\_multivalue $-2.75$); on the fused path neither loss is significant.

\begin{table}[h]
\centering\small
\caption{Answer flips against BF16: items whose score went down / up (graded suites: items whose official score changed, down / up). $^\dagger$\,Before fusion. Long reasoning: items whose sample accuracy went down / up.}
\label{tab:flips}
\setlength{\tabcolsep}{4pt}
\begin{tabular}{@{}lrrrrrr@{}}
\toprule
suite (items) & KIVI-4 + V & \sys{}-R & \sys{}-M & KIVI-4$^\dagger$ & KIVI-4 + loop-3 sharing$^\dagger$ & FP8 + loop-3 sharing \\
\midrule
RULER-16K (1,300) & 30 / 37 & 26 / 44 & 54 / 71 & 29 / 31 & 45 / 87 & 66 / 80 \\
RULER-32K (1,300) & 37 / 47 & 44 / 45$^\dagger$ & 84 / 89 & 40 / 47 & --- & --- \\
LongBench v1 (3,750) & 499 / 480 & 513 / 540 & 556 / 596 & 483 / 495 & 579 / 570 & 681 / 719 \\
MATH500 (200) & 8 / 8 & 9 / 9 & 7 / 10 & --- & 9 / 14 & 5 / 9 \\
GSM8K (1,319) & 21 / 25 & 22 / 22 & 27 / 34 & --- & 27 / 30 & 59 / 45 \\
long reasoning (194) & --- & 34 / 39$^\dagger$ & 33 / 38$^\dagger$ & --- & --- & --- \\
MQ641 (641) & 13 / 16 & 12 / 16 & 18 / 21 & 9 / 15 & 29 / 19 & --- \\
\bottomrule
\end{tabular}
\end{table}

\paragraph{Ouro-2.6B, served.}
Ouro-2.6B (48 layers) is served on the fused path (fused kernel, whole-step graph, $\tau{=}14$, production seed) with the two plans that the same allocator computes from its profile (Table~\ref{tab:ouro26}).
GSM8K (all 1,319 items, base checkpoint, 3-shot, strict match) ran on A40 GPUs, with every arm of an item slice on the same GPU; we ran MATH500 (the same 200 items as Ouro-1.4B, Thinking checkpoint, greedy, 8K output cap) on one RTX PRO 6000, with each arm's concurrency capped at its pool's capacity for 8K requests, and the engine counted no full pool (Table~\ref{tab:ouro26served}).
Before the A40 runs, served decode KL matched the H200 graph-path values of the independent run within 5\% (1.00--1.03\x{}; 1.008\x{} / 1.000\x{} for \sys{}-R, which ran later on other A40s against a BF16 re-run there), and their in-serving byte checks of 12 pages differ in 1, 3, 4 and 2 bytes for KIVI-4, KIVI-4 + V, \sys{}-M and \sys{}-R, all on the shared source pages, the rounding class also seen on the RTX PRO 6000 (Appendix~\ref{app:checks}).

\begin{table}[h]
\centering\small
\caption{Ouro-2.6B served (fused path): no quantized arm differs significantly from BF16; \sys{}-M misses the 3-point margin on MATH500 by half a point. MATH500 on one RTX PRO 6000, every arm at our 8,192-token output cap; GSM8K on A40 GPUs, same-GPU pairs. Accuracy (\%) and $\Delta$ vs BF16 on the same GPU (pp, from unrounded scores); history KV stored per token. $^\S$\,Second set of A40s.}
\label{tab:ouro26served}
\setlength{\tabcolsep}{4pt}
\begin{tabular}{@{}lrll@{}}
\toprule
arm & KiB / token & MATH500 (200) & GSM8K (1,319) \\
\midrule
BF16 & 1,536 & 89.5 & 79.7 \\
KIVI-4 + V & 480 (31\%) & 90.5, $+1.0$ & 79.5, $-0.2$ \\
\sys{}-M & 310 (20\%) & 89.5, $+0.0$ & 79.5, $-0.2$ \\
\sys{}-R & 430 (28\%) & 90.5, $+1.0$ & --- \\
\midrule
BF16$^\S$ & 1,536 & --- & 79.8 \\
\sys{}-R$^\S$ & 430 (28\%) & --- & 79.2, $-0.5$ \\
\bottomrule
\end{tabular}
\end{table}

\section{Throughput and SLO details}\label{app:tputslo}
Part of V precision's lift recovers the cost of the two-plane V layout: uniform INT4 decodes 479.4\,tok/s with two planes on the fused path, against 494.4 with interleaved V on the reader before fusion.

\begin{table}[h]
\centering\small
\caption{Ablation of \sys{}'s components on one RTX PRO 6000: each step improves throughput, KV footprint, or batch-1 latency at a stated KL. Columns report stored / read history KV (KiB/token), fp32 KL / \Pz{}, peak decode throughput at 8K and 16K (tok/s), relative speedup over the preceding row measured back-to-back in the same job (``same block $\times$'', at 8K peak / $B{=}1$), and host launches per step. The read- and memory-optimized plans are separate knapsack solutions, and the indented rows incrementally add C3, kernel fusion, and whole-step CUDA graphs to the memory-optimized plan.}
\label{tab:ablation}
\setlength{\tabcolsep}{4pt}
\begin{tabular}{@{}lrrrrrr@{}}
\toprule
configuration & bytes & KL & 8K & 16K & \makecell{same\\ block $\times$} & \makecell{host\\ launches} \\
\midrule
KIVI-4 & 240/240 & 0.46 & 494.4 & 249.1 & --- & --- \\
read-optimized plan (C1) & 215/215 & 0.87 & 528.9 & 271.5 & --- & 433 \\
memory-optimized plan (C1 + C2) & 155/220 & 1.80 & 535.1 & 280.7 & --- & --- \\
\quad + V precision (C3) & 155/189 & 1.80 & 561.6 & 295.5 & 1.044 / 0.98 & 504 \\
\quad + fused kernel & 155/189 & 1.80 & 571.2 & 300.0 & 1.019 / 1.033 & 332 \\
\quad + whole-step CUDA graph (\sys{}-M) & 155/189 & 1.80 & 575.9 & 301.9 & 1.006 / 1.067 & 17 \\
\bottomrule
\end{tabular}
\end{table}

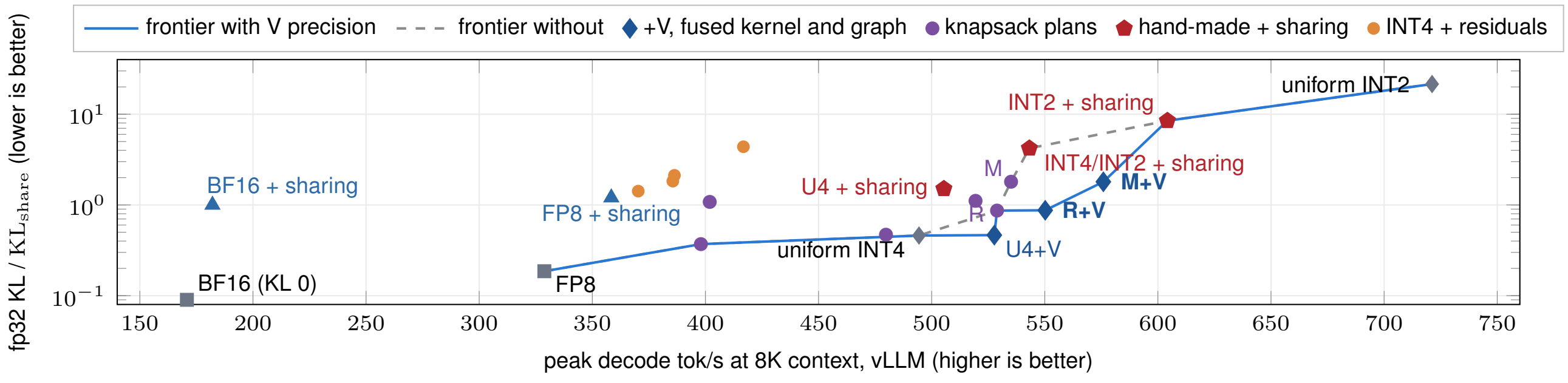
\begin{figure*}[t]
\centering
\resizebox{\textwidth}{!}{\input{figures/v4/fig_pareto_full.tex}}
\vspace{-3mm}
\caption{All served configurations at 8K context, including \sys{}'s own variants (Figure~\ref{fig:pareto} shows one point per comparable method): knapsack plans without V precision, KIVI-4 + V, hand-made placements with sharing, and INT4 designs with residuals. KIVI-4 + V adds our V precision (C3) to KIVI-4 and is an ablation; the sharing points combine loop-3 sharing with BF16, FP8, KIVI-4 or hand-made INT4 / INT2 placements. The frontier with V precision runs through uniform INT4, KIVI-4 + V and \sys{}-R and \sys{}-M; every hand-made and residual design with at most \sys{}-M's KL lies behind it. Points without +V use our unfused reader, except BF16 and FP8 (vLLM) and uniform INT2 (a separate run).}
\label{fig:paretofull}
\vspace{-2mm}
\end{figure*}

\begin{figure*}[h]
\centering
\resizebox{\textwidth}{!}{\input{figures/v4/fig_pareto.tex}}
\vspace{-3mm}
\caption{At 8K context, \sys{}-R and \sys{}-M (fused kernel and graph) lie on the measured frontier of served configurations between KIVI-4 (0.46\x{} \Pz{}) and KIVI-2 (21.5\x{} \Pz{}), the only faster method. Comparable methods, one point each: BF16 and FP8 as vLLM serves them, KIVI-4 and KIVI-2, and INT2 residuals for loops 1--2 against an INT4 loop 4, with loop 3 shared, in the style of ResidualQuant~\cite{kim2026residualquant} (4.38\x{} \Pz{}; our reimplementation of ResidualQuant itself has 2.01\x{}, Table~\ref{tab:quality}); the loop-sharing points combine loop-3 sharing with BF16 (\Pz{}), FP8 and KIVI-4. The KIVI, sharing and residual points use our unfused reader. Figure~\ref{fig:paretofull} adds KIVI-4 + V, also on the frontier, and our other variants. Single runs.}
\label{fig:pareto}
\vspace{-2mm}
\end{figure*}

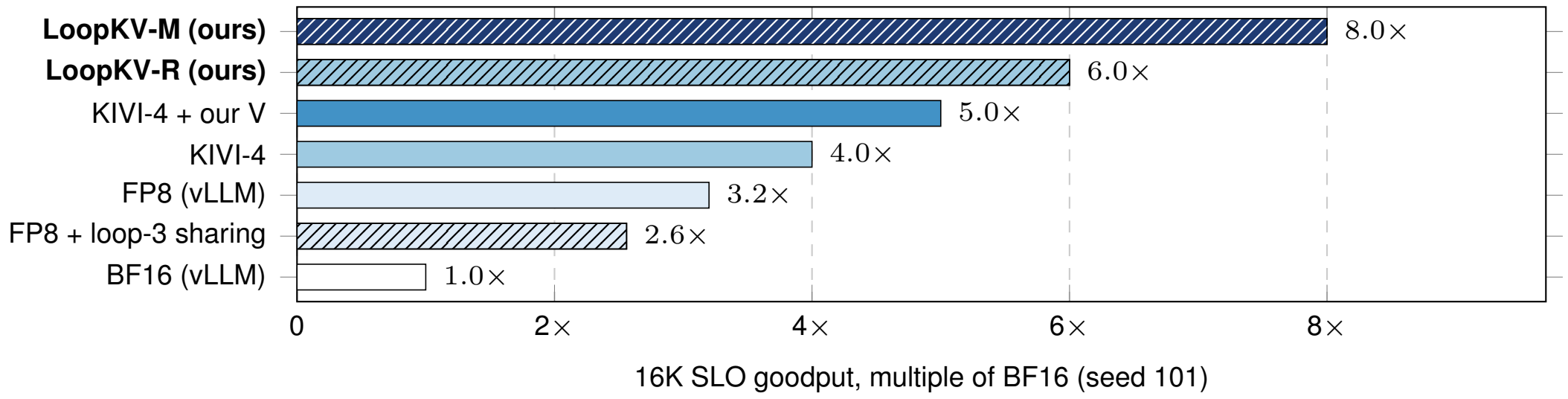
\begin{figure}[h]
\centering
\input{figures/v4/fig_slocap.tex}
\vspace{-2mm}
\caption{\sys{}-M sustains 8.0\x{} BF16's 16K SLO goodput on one RTX PRO 6000, and 2.0\x{} that of KIVI-4 on the same kernel. Highest request rate with 95\% of requests in SLO ($\tau_0{=}18.60$\,ms, \S\ref{sec:eval-setup}), arrival seed 101, which every arm ran, relative to BF16's 0.0119\,req/s; on seeds 102 and 103, \sys{}-M sustains 1.72\x{} and 2.20\x{} KIVI-4's rate and \sys{}-R 1.29\x{} and 1.60\x{}.}
\label{fig:slocap}
\end{figure}

On the RTX PRO 6000, vLLM 0.26.0 runs with PyTorch 2.11 (CUDA 13), memory utilization 0.90.
Table~\ref{tab:tput} lists the peak decode throughput of every configuration.
The SLO prompts have a mean output of 10,999 tokens, and attainment is scored over the requests that arrive in the middle 80\% of the arrival window, 43, 49 and 47 of 60 for seeds 101--103.
The SLO's $\tau_0$ comes from a stock-vLLM step-time fit of $17.31$\,ms plus $0.495$\,ms per 1K tokens of context, evaluated at a shorter workload's mean context for the 18.60\,ms rule and at this workload's own for the 20.69\,ms rule (at a maximum model length of 17,408), which therefore allow $3\tau_0{=}55.8$\,ms and 62.1\,ms per output token.
Capacity uses the 95\% rule, fixed before the ladders ran.

Table~\ref{tab:tputx} repeats the own-peak protocol for Ouro-2.6B and on a second GPU type, an A100-80. On the A100-80 and for Ouro-2.6B, every arm of a context runs on one physical GPU, so each ratio is a same-card ratio. Ouro-2.6B's plugin arms need a smaller code pool than the plugin's default share to stay resident up to their formula batch (0.80, and 0.75 for the memory-optimized plan at 8K). At 32K, Ouro-1.4B's plugin arms use a share of 0.88, as on the RTX PRO 6000.
Under the SLO protocol of \S\ref{sec:eval-setup} on one A100-80, with this GPU's $\tau_0$ rules of 16.60 and 26.34\,ms and every arm on this GPU, the \planR{} sustains 0.0244 / 0.0381\,req/s, the \planM{} 0.0244 / 0.0476, uniform INT4 0.0204 / 0.0305 and BF16 0.0083 / 0.0133, so the two plans sustain 1.20--1.56\x{} uniform INT4's request rate and 2.9--3.6\x{} BF16's.
\begin{table}[h]
\centering\small
\caption{Peak decode tok/s at the resident peak (batch in parentheses) and the ratio to BF16 on the same GPU, fused kernel and whole-step graph, one run per cell. Uniform INT4 is KIVI-4 without V precision; LoopKV-R and -M are the read- and memory-optimized plans with V precision; Ouro-2.6B uses its own allocator plans.}
\label{tab:tputx}
\setlength{\tabcolsep}{4pt}
\begin{tabular}{@{}llrrrr@{}}
\toprule
model, GPU & context & BF16 & uniform INT4 & LoopKV-R & LoopKV-M \\
\midrule
Ouro-1.4B, A100-80 & 8K & 188.4 (10) & 336.6 (27) 1.79\x{} & 377.6 (30) 2.00\x{} & 408.5 (42) 2.17\x{} \\
 & 16K & 95.0 (5) & 161.3 (11) 1.70\x{} & 189.9 (13) 2.00\x{} & 209.2 (20) 2.20\x{} \\
 & 32K & 22.7 (2) & 84.6 (6) 3.73\x{} & 105.4 (8) 4.65\x{} & 107.0 (11) 4.72\x{} \\
\midrule
Ouro-2.6B, RTX PRO 6000 & 8K & 81.3 (6) & 208.3 (16) 2.56\x{} & 240.4 (18) 2.96\x{} & 263.9 (23) 3.25\x{} \\
 & 16K & 30.8 (3) & 109.5 (8) 3.56\x{} & 124.1 (8) 4.03\x{} & 136.8 (11) 4.44\x{} \\
 & 32K & 6.2 (1) & 52.7 (4) 8.51\x{} & 58.6 (4) 9.46\x{} & 68.1 (5) 11.0\x{} \\
\midrule
Ouro-2.6B, A100-80 & 8K & 78.5 (5) & 133.2 (13) 1.70\x{} & 148.5 (14) 1.89\x{} & 174.1 (19) 2.22\x{} \\
 & 16K & 20.2 (2) & 68.6 (6) 3.40\x{} & 76.6 (7) 3.80\x{} & 87.3 (10) 4.33\x{} \\
 & 32K & 5.7 (1) & 36.2 (3) 6.35\x{} & 39.4 (3) 6.91\x{} & 46.4 (5) 8.13\x{} \\
\bottomrule
\end{tabular}
\end{table}

\begin{table}[h]
\centering\small
\caption{With the fused kernel and whole-step graph, \sys{}-R / -M decode 5.4 / 9.3\% (8K), 7.8 / 13.2\% (16K) and 7.4 / 15.6\% (32K) more than KIVI-4 + V given the same kernel and graph (single runs; repeat means at 8K and 16K: \S\ref{sec:eval-tput}). Peak decode tok/s (batch); KiB per token, stored / read. Top two blocks: before fusion, uniform INT4 with interleaved V; bottom: fused + graph, two-plane V, one block at 8K and 16K (read-optimized + V at batch 36: 552.7) and a second block on the same GPU at 32K. BF16 peaks at batch 8 at 8K: at batch 12 stock FlashAttention-2 falls to one split per head (154.9\,tok/s). At batch 1 (8K, bottom block) the plans match KIVI-4 + V: 79.1 and 78.9 vs 80.3\,tok/s.}
\label{tab:tput}
\setlength{\tabcolsep}{4pt}
{%
\begin{tabular}{@{}lrrrrr@{}}
\toprule
plan & bytes & 8K & 16K & 32K & \makecell{$\times$ BF16 /\\ FP8+S, 8K} \\
\midrule
BF16 & 768/768 & 170.8 (8) & 89.0 (6) & --- & 1.00 / 0.48 \\
FP8 & 384/384 & 328.8 (16) & 176.3 (13) & --- & 1.93 / 0.92 \\
FP8 + sharing (FP8+S) & 288/384 & 358.4 (33) & 180.7 (16) & --- & 2.10 / 1.00 \\
uniform INT4 & 240/240 & 494.4 (34) & 249.1 (14) & --- & 2.89 / 1.38 \\
\midrule
uniform-INT4 + sharing & 180/240 & 505.4 (40) & 264.2 (20) & --- & 2.96 / 1.41 \\
\quad + V precision & 180/200 & 521.1 (40) & 277.4 (20) & --- & 3.05 / 1.45 \\
read-optimized & 215/215 & 528.9 (36) & 271.5 (16) & --- & 3.10 / 1.48 \\
\quad + V precision & 215/186 & 544.5 (36) & 282.9 (16) & --- & 3.19 / 1.52 \\
memory-optimized & 155/220 & 535.1 (48) & 280.7 (24) & --- & 3.13 / 1.49 \\
\quad + V precision & 155/189 & 561.6 (48) & 295.5 (24) & --- & 3.29 / 1.57 \\
\midrule
uniform INT4, fused + graph & 240/240 & 479.4 (34) & 241.7 (14) & 127.2 (8) & 2.81 / 1.34 \\
\quad + V precision & 240/200 & 527.7 (34) & 267.8 (14) & 142.9 (8) & 3.09 / 1.47 \\
read-opt.\ + V, fused + graph & 215/186 & 556.3 (34) & 288.6 (16) & 153.5 (10) & 3.26 / 1.55 \\
memory-opt.\ + V, fused + graph & 155/189 & 577.0 (48) & 303.2 (24) & 165.2 (14) & 3.38 / 1.61 \\
\bottomrule
\end{tabular}}
\end{table}

\paragraph{Throughput at a matched time per token.}
Every decode step gives each of the $B$ resident requests one token, so the time per output token at a peak is $B$ divided by its tok/s.
In the fused block of Table~\ref{tab:tput} it is 64.4, 52.3 and 56.0\,ms for KIVI-4 + V at 8K, 16K and 32K, 61.1, 55.4 and 65.2\,ms for \sys{}-R and 83.2, 79.2 and 84.8\,ms for \sys{}-M; BF16 and FP8 + sharing take 46.8 and 92.1\,ms at their 8K peaks.
To compare the arms at the SLO's bound of $3\tau_0{=}55.8$\,ms, we interpolate the step time, which is linear in $B$ as Eq.~\ref{eq:step} predicts, between the two measured batches that bracket the bound (batches 1, 8, 16, 34 and each arm's peak at 8K; 1, 8, 14 and the peak at 16K; 1, 4, 6, 8 and the peak at 32K), and keep the peak of an arm whose peak already meets the bound.
At 55.8\,ms, KIVI-4 + V decodes 508, 268 and 142\,tok/s at 8K, 16K and 32K, \sys{}-R 543, 289 and 152 (1.069\x{}, 1.078\x{} and 1.069\x{}) and \sys{}-M 539, 286 and 152 (1.061\x{}, 1.067\x{} and 1.067\x{}).
Eq.~\ref{eq:step}, with the constants of \S\ref{sec:bg-kv} and the read bytes of Table~\ref{tab:tput} plus the BF16 sink and window, gives 1.05--1.09\x{} for the same comparisons, and \sys{}-M measured at 16K and batch 16 in an earlier block on the same GPU (56.05\,ms, 285.5\,tok/s) agrees with the interpolation.
At a matched time per token both plans are therefore 6--8\% faster than KIVI-4 + V at every context, and \sys{}-M's larger resident batches lift its peak gains to 8.4--15.6\%.
Under the same bound BF16 and FP8 + sharing decode 171 and 331\,tok/s at 8K and 77 and 167\,tok/s at 16K, so \sys{}-M's multiples are 3.2\x{} and 1.6\x{} at 8K and 3.7\x{} and 1.7\x{} at 16K; at 32K, BF16 takes 81.5\,ms per token already at batch 1.
At the 62.1\,ms bound of the 20.69\,ms rule the plans gain 6--9\%.

\begin{figure*}[t]
\centering
\begin{tikzpicture}
\begin{groupplot}[group style={group size=3 by 1, horizontal sep=0.9cm},
  width=0.345\textwidth, height=4.0cm, xmin=0.03, xmax=0.185, ymin=0.68, ymax=1.03,
  xlabel={arrival rate (req/s)}, tick label style={font=\scriptsize}, label style={font=\footnotesize},
  title style={font=\footnotesize, yshift=-1ex}, xtick={0.04,0.08,0.12,0.16},
  x tick label style={/pgf/number format/fixed, /pgf/number format/precision=2},
  ytick={0.7,0.8,0.9,0.95,1.0}, yticklabels={0.70,0.80,0.90,0.95,1.00}, grid=major, grid style={gray!20},
  legend style={font=\scriptsize, draw=none, fill=white, fill opacity=0.85, text opacity=1, at={(0.02,0.02)}, anchor=south west, row sep=-2pt}]
\nextgroupplot[title={seed 101}, ylabel={attainment ($3\tau_0$, 18.60\,ms)}]
\addplot[red!70!black, thin, forget plot] coordinates {(0.03,0.95) (0.185,0.95)};
\addplot[gray!80, densely dotted, mark=x, mark size=1.6pt] coordinates {(0.0476,1.000) (0.0595,0.930) (0.0714,0.698)};
\addplot[cBase, dashed, mark=square*, mark size=1.3pt] coordinates {(0.0381,1.000) (0.0476,1.000) (0.0595,1.000) (0.0714,0.884)};
\addplot[cRes, mark=triangle*, mark size=1.6pt] coordinates {(0.0714,0.977) (0.0833,0.791)};
\addplot[lkblue, thick, mark=*, mark size=1.3pt] coordinates {(0.0595,1.000) (0.0714,1.000) (0.0833,1.000) (0.0952,0.977) (0.1071,0.930) (0.1190,0.744)};
\nextgroupplot[title={seed 102}, legend style={at={(0.5,-0.42)}, anchor=north, legend columns=4, /tikz/every even column/.append style={column sep=8pt}}]
\addplot[red!70!black, thin, forget plot] coordinates {(0.03,0.95) (0.185,0.95)};
\addplot[gray!80, densely dotted, mark=x, mark size=1.6pt] coordinates {(0.0476,1.000) (0.0595,1.000) (0.0714,1.000) (0.0833,0.980) (0.0952,0.857)};
\addplot[cBase, dashed, mark=square*, mark size=1.3pt] coordinates {(0.0595,1.000) (0.0714,1.000) (0.0833,1.000) (0.0952,0.939) (0.1071,0.776)};
\addplot[cRes, mark=triangle*, mark size=1.6pt] coordinates {(0.0714,1.000) (0.0833,1.000) (0.0952,1.000) (0.1071,0.980) (0.1190,0.857)};
\addplot[lkblue, thick, mark=*, mark size=1.3pt] coordinates {(0.0952,1.000) (0.1071,1.000) (0.1190,1.000) (0.1310,1.000) (0.1429,0.980) (0.1548,0.918) (0.1667,0.918) (0.1786,0.837)};
\legend{uniform INT4, KIVI-4 + V, \sys{}-R, \sys{}-M}
\nextgroupplot[title={seed 103}]
\addplot[red!70!black, thin, forget plot] coordinates {(0.03,0.95) (0.185,0.95)};
\addplot[gray!80, densely dotted, mark=x, mark size=1.6pt] coordinates {(0.0476,1.000) (0.0595,1.000) (0.0714,0.936) (0.0833,0.872)};
\addplot[cBase, dashed, mark=square*, mark size=1.3pt] coordinates {(0.0595,1.000) (0.0714,1.000) (0.0833,0.894)};
\addplot[cRes, mark=triangle*, mark size=1.6pt] coordinates {(0.0714,1.000) (0.0833,1.000) (0.0952,0.957) (0.1071,0.872)};
\addplot[lkblue, thick, mark=*, mark size=1.3pt] coordinates {(0.0952,1.000) (0.1071,1.000) (0.1190,1.000) (0.1310,0.957) (0.1429,0.894) (0.1548,0.894)};
\end{groupplot}
\end{tikzpicture}
\vspace{-2mm}
\caption{SLO attainment against arrival rate for the INT4 arms on the fused path (one RTX PRO 6000, 16K maximum length; TTFT $\le$ 10\,s and TPOT $\le 3\tau_0$ at $\tau_0{=}18.60$\,ms; red line: the 95\% rule). Each point is one cell of 60 requests, of which 43--49 are scored; Table~\ref{tab:sloladder} gives the scored counts and the other SLO bounds.}
\label{fig:sloatt}
\vspace{-2mm}
\end{figure*}
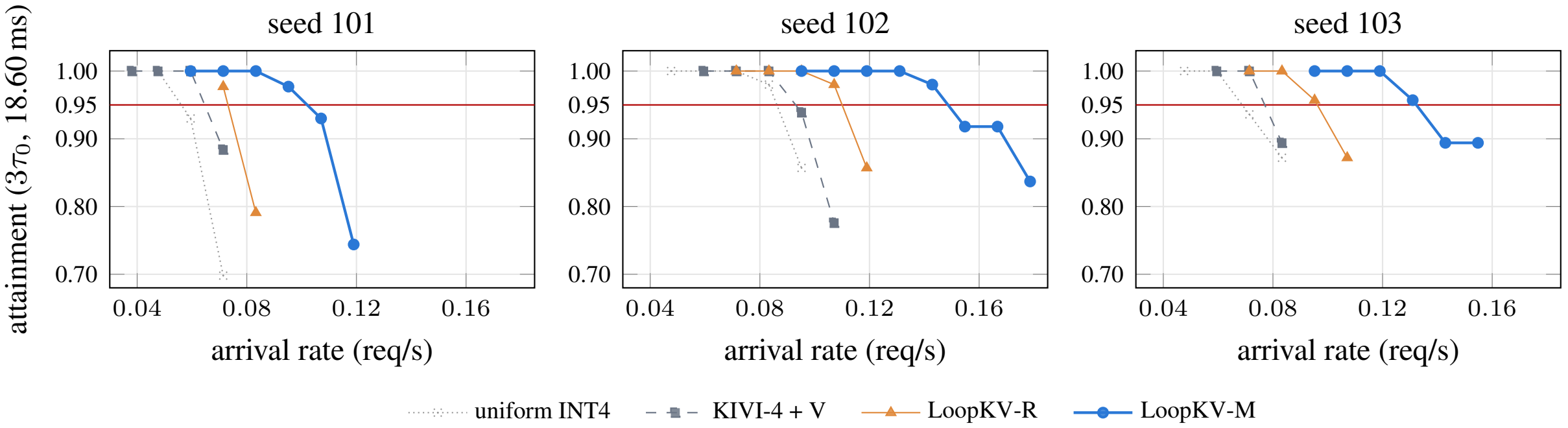

\paragraph{SLO ladders.}
Table~\ref{tab:sloladder} and Figure~\ref{fig:sloatt} give every ladder of Figure~\ref{fig:slocap}, rescored from the stored per-request records with the scorer's own code, which reproduces every reported attainment exactly.
The INT4 ladders step by $1/84{\approx}0.0119$\,req/s.
Resampling the scored requests of every cell (10,000 draws) keeps the \sys{}-M / KIVI-4 + V ratio at 1.5\x{} or more in 86\%, 97\% and 96\% of the draws for seeds 101--103 at 18.60\,ms; in a draw in which KIVI-4 + V passes its highest tested rung, its capacity is censored at that rung.
At their first failing rungs, KIVI-4 + V misses mostly on time to first token after preemptions (11 of 13 misses at 18.60\,ms, 20 of 21 at 20.69\,ms), and so does \sys{}-R (22 of 22 at 18.60\,ms), whereas \sys{}-M at 18.60\,ms misses mostly on time per token (8 of 12), with its 95th-percentile TPOT within 0.5\,ms of the 55.8\,ms bound; BF16 and the FP8 arms miss on time per token only.
With TPOT relaxed to $5\tau_0$, KIVI-4 + V keeps its capacity on two seeds and \sys{}-M rises, to 1.80\x{}, at least 1.88\x{} and 2.00\x{}; moving the TTFT bound to 5 or 20\,s keeps the ratio within 1.50--1.83\x{}.
V precision alone adds one rung to uniform INT4 on seeds 101 and 103 at 18.60\,ms and on seed 102 at 20.69\,ms, and none in the other three cases.
Plain FP8 without sharing, run on seed 101 on the same GPU after the other ladders, sustains 0.0381\,req/s under both rules and fails 0.0476 on time per token (0.535 and 0.791), one rung more than FP8 + sharing at 18.60\,ms, as the decode sweeps suggest (at 55.8\,ms it matches FP8 + sharing at 8K, 329 against 331\,tok/s, and leads at 16K, 171 against 167); against plain FP8, \sys{}-M's multiple is 2.5\x{} at 18.60\,ms and 2.8\x{} at 20.69\,ms.

\begin{table}[h]
\centering\small
\caption{SLO ladders on one RTX PRO 6000 (Ouro-1.4B, 16K maximum length, fused path for the INT4 arms; 60 requests per cell, of which 43, 49 and 47 are scored for seeds 101--103). Rungs: lowest and highest rate tested; INT4 rungs step by 0.0119\,req/s. Capacity: highest tested rate before the first rung below 95\% attainment, with TTFT $\le$ 10\,s and TPOT $\le k\tau_0$; $<$: the lowest rung fails; $\ge$: no tested rung fails. First failing rung (the rung above the $3\tau_0$ capacity at $\tau_0{=}18.60$\,ms): attainment, misses by component (TTFT only / TPOT only / both) and preemptions. Ratio rows: capacity ratios on the same trace, and the share of 10,000 request-bootstrap draws in which the $3\tau_0$ ratio stays at 1.5\x{} or more.}
\label{tab:sloladder}
\setlength{\tabcolsep}{4pt}
\begin{tabular}{@{}llcrrrrccr@{}}
\toprule
 & & & \multicolumn{4}{c}{capacity (req/s), $\tau_0{=}18.60$\,ms or as noted} & \multicolumn{3}{c}{first failing rung, $3\tau_0$, 18.60\,ms} \\
\cmidrule(lr){4-7}\cmidrule(l){8-10}
arm & seed & rungs (req/s) & $2\tau_0$ & $3\tau_0$ & $5\tau_0$ & $3\tau_0$, 20.69 & attainment & T / P / B & preempt. \\
\midrule
BF16 & 101 & 0.0119--0.0170 & $<$0.0119 & 0.0119 & $\ge$0.0170 & 0.0143 & 0.605 & 0 / 17 / 0 & 3 \\
FP8 & 101 & 0.0305--0.0476 & $<$0.0305 & 0.0381 & $\ge$0.0476 & 0.0381 & 0.535 & 0 / 20 / 0 & 6 \\
FP8 + sharing & 101 & 0.0305--0.0476 & $<$0.0305 & 0.0305 & $\ge$0.0476 & 0.0381 & 0.930 & 0 / 3 / 0 & 0 \\
uniform INT4 & 101 & 0.0476--0.0714 & $<$0.0476 & 0.0476 & 0.0595 & 0.0595 & 0.930 & 1 / 2 / 0 & 92 \\
 & 102 & 0.0476--0.0952 & 0.0476 & 0.0833 & 0.0833 & 0.0833 & 0.857 & 5 / 0 / 2 & 93 \\
 & 103 & 0.0476--0.0833 & $<$0.0476 & 0.0595 & 0.0714 & 0.0714 & 0.936 & 2 / 1 / 0 & 112 \\
KIVI-4 + V & 101 & 0.0381--0.0714 & 0.0381 & 0.0595 & 0.0595 & 0.0595 & 0.884 & 3 / 0 / 2 & 100 \\
 & 102 & 0.0595--0.1071 & $<$0.0595 & 0.0833 & 0.0952 & 0.0952 & 0.939 & 2 / 1 / 0 & 44 \\
 & 103 & 0.0595--0.0833 & $<$0.0595 & 0.0714 & 0.0714 & 0.0714 & 0.894 & 4 / 1 / 0 & 80 \\
\makecell[l]{uniform INT4\\\quad+ sharing + V} & 101 & 0.0714--0.0952 & $<$0.0714 & 0.0714 & 0.0833 & 0.0714 & 0.907 & 1 / 2 / 1 & 189 \\
\sys{}-R & 101 & 0.0714--0.0833 & $<$0.0714 & 0.0714 & 0.0714 & 0.0714 & 0.791 & 8 / 0 / 1 & 209 \\
 & 102 & 0.0714--0.1190 & $<$0.0714 & 0.1071 & 0.1071 & 0.1071 & 0.857 & 7 / 0 / 0 & 220 \\
 & 103 & 0.0714--0.1071 & $<$0.0714 & 0.0952 & 0.0952 & 0.0952 & 0.872 & 6 / 0 / 0 & 209 \\
\sys{}-M & 101 & 0.0595--0.1190 & $<$0.0595 & 0.0952 & 0.1071 & 0.1071 & 0.930 & 1 / 1 / 1 & 128 \\
 & 102 & 0.0952--0.1786 & $<$0.0952 & 0.1429 & $\ge$0.1786 & 0.1667 & 0.918 & 1 / 3 / 0 & 117 \\
 & 103 & 0.0952--0.1548 & $<$0.0952 & 0.1310 & 0.1429 & 0.1429 & 0.894 & 2 / 3 / 0 & 182 \\
\midrule
\makecell[l]{\sys{}-M /\\\quad KIVI-4 + V} & 101 & & $<$1.56 & 1.60 & 1.80 & 1.80 & \multicolumn{3}{l}{$\ge$1.5\x{} in 86\% of draws} \\
 & 102 & & -- & 1.72 & $\ge$1.88 & 1.75 & \multicolumn{3}{l}{$\ge$1.5\x{} in 97\% of draws} \\
 & 103 & & -- & 1.83 & 2.00 & 2.00 & \multicolumn{3}{l}{$\ge$1.5\x{} in 96\% of draws} \\
\makecell[l]{\sys{}-R /\\\quad KIVI-4 + V} & 101 & & -- & 1.20 & 1.20 & 1.20 & & & \\
 & 102 & & -- & 1.29 & 1.13 & 1.13 & & & \\
 & 103 & & -- & 1.33 & 1.33 & 1.33 & & & \\
\bottomrule
\end{tabular}
\end{table}

\begin{table}[h]
\centering\small
\caption{On an H100 SXM (80\,GB), \sys{}-M decodes 2.69\x{} as fast as BF16 at 8K and 16K, against 2.27\x{} and 1.96\x{} for KIVI-4, and no INT4 arm differs significantly from BF16 on MATH500 or GSM8K. Own-peak decode throughput relative to BF16 (each arm on its own H100; a rerun of every arm on one GPU agrees within 2\%), step time at $B{=}1$ and 8K, and served accuracy on MATH500 (500 items) and GSM8K (1,319 items). The H100 kernel reads both V planes.}
\label{tab:h100}
\setlength{\tabcolsep}{4pt}
\begin{tabular}{@{}lrrrrr@{}}
\toprule
arm & 8K peak & 16K peak & $B{=}1$, 8K (ms) & MATH500 (\%) & GSM8K (\%) \\
\midrule
BF16 & 1.00\x{} & 1.00\x{} & 10.68 & 83.2 & 76.6 \\
FP8 & 2.08\x{} & 2.05\x{} & --- & --- & --- \\
KIVI-4 & 2.27\x{} & 1.96\x{} & 10.63 & 83.2 & 77.0 \\
KIVI-4 + V & 2.26\x{} & 1.96\x{} & 10.55 & 85.4 & 76.5 \\
\sys{}-R & 2.54\x{} & 2.41\x{} & 10.35 & 84.6 & 76.8 \\
\sys{}-M & 2.69\x{} & 2.69\x{} & 10.33 & 84.4 & 77.0 \\
\bottomrule
\end{tabular}
\end{table}

\section{Byte model}\label{app:bytes}
Ouro-1.4B stores, per token and layer, $2\times16\times128 = 4096$ K and V elements per loop.
KIVI-style codes with group size 32 carry one fp16 scale and one fp16 zero point per 32 elements, 1 bit per element of metadata: INT4 is 5 bits per element, 2.5\,KiB per token and layer, and INT2 3 bits, 1.5\,KiB.
Over 24 layers and 4 loops, INT4 on every loop is 240\,KiB per token and BF16 768\,KiB.
A shared cell stores nothing and reads the source's 2.5\,KiB; a residual cell stores its residual and reads the residual plus the source (INT4 residual: 2.5 / 5.0\,KiB; INT2: 1.5 / 4.0).
V precision skips the 2-bit lo plane of V, 0.5\,KiB per INT4 cell, for the cold fraction of tokens.
The sink and window (288 tokens, BF16, 4 loops) add 216\,MiB per request and step to $D_\text{read}$ in Eq.~\ref{eq:step}, 10\% of the 8K read of the \planU{} and 5\% at 16K.

\begin{table}[h]
\centering\small
\caption{Once quantized, throughput follows bytes read: at equal storage, own INT4 codes decode 1.36\x{} faster than residuals at 8K. KiB per token of history (4 loops); measured peaks in vLLM on one RTX PRO 6000, except the 16K peak of INT4 residuals, run on the second GPU (\S\ref{sec:eval-setup}). The cells of the two quantized rows that ran on this GPU, served by our plugin before fusion, lie at 98--103\% of Eq.~\ref{eq:step}; vLLM's own BF16 and FP8 rows run faster than Eq.~\ref{eq:step} predicts (BF16 at 8K: 170.8 against about 146\,tok/s).}
\label{tab:o5}
\setlength{\tabcolsep}{4pt}
{%
\begin{tabular}{@{}lrrr@{}}
\toprule
KV format & stored & read & tok/s 8K / 16K \\
\midrule
BF16 & 768 & 768 & 170.8 / \phantom{0}89.0 \\
BF16 + loop-3 sharing (\Pz{}) & 576 & 768 & 182.1 / \phantom{0}95.8 \\
FP8 & 384 & 384 & 328.8 / 176.3 \\
FP8 + loop-3 sharing & 288 & 384 & 358.4 / 180.7 \\
INT4 final + INT4 residuals & 180 & 360 & 370.3 / 192.0 \\
KIVI-4 + loop-3 sharing & 180 & 240 & 505.4 / 264.2 \\
\bottomrule
\end{tabular}}
\end{table}

\section{Additivity of cell damages and further planner plans}\label{app:additivity}
The knapsack scores a plan by the sum of its cells' single-cell damages. Proposition~\ref{prop:add} gives the reason this sum predicts a plan's KL, and the measurements below test it.

\begin{proposition}[Approximate additivity]\label{prop:add}
Let $x_c$ be the exact model's history KV of cell $c$. A plan stores $x_c+\delta_c$, with $\delta_c=Q(x_c)-x_c$ for the cell's own codes and $\delta_c=Q(x_s)-x_c$ for a cell that reads its source $s$, coded as when the cell was profiled; each $\delta_c$ is supported on cell $c$ and does not depend on the other cells' options. Let the KL $D$ be twice differentiable in the stored KV, $D^{(2)}(\delta)=\tfrac12\delta^\top H\delta$ its second-order part, so that $D(\delta)=D^{(2)}(\delta)+O(\|\delta\|^3)$, and $D_c=D^{(2)}(\delta_c)$, which the profile measures up to $O(\|\delta_c\|^3)$. Then
(a) $D^{(2)}\bigl(\sum_c\delta_c\bigr)=\sum_c D_c+\sum_{c<c'}\delta_c^\top H\,\delta_{c'}$, with $H\succeq 0$;
(b) if the errors of a set $\mathcal I$ of cells are random with zero mean and uncorrelated with each other and with every other cell's error, every interaction term that involves a cell of $\mathcal I$ has zero mean, so $\mathbb E\,D^{(2)}(\delta)=\sum_{c\in\mathcal I}\mathbb E\,D_c+\mathbb E\,D^{(2)}\bigl(\sum_{c\notin\mathcal I}\delta_c\bigr)$.
\end{proposition}
\emph{Proof.} The KL is zero and minimal at $\delta=0$, so $D(0)=0$, $\nabla D(0)=0$, and its Hessian there is $H=J^\top FJ\succeq0$: the Fisher information $F$ of the output distribution pulled back through the Jacobian $J$ of the logits with respect to the stored KV, averaged over positions and documents. The term with second derivatives of the logits vanishes, because the gradient of the KL with respect to the logits is zero at $\delta=0$. Expanding the quadratic form of $\sum_c\delta_c$ gives (a). Under (b), a term $\delta_c^\top H\delta_{c'}$ with $c\in\mathcal I$ has expectation $\mathrm{tr}(H\,\mathbb E[\delta_{c'}\delta_c^\top])=0$, since the two errors are uncorrelated and at least one has zero mean. \hfill$\square$

Here rounding errors are modeled as random, the usual additive-noise model of quantization, and $\mathbb E$ is the expectation under this model. Each interaction term sums products of errors over many tokens and channels, so it concentrates near its zero mean, and the sum of single-cell damages becomes a stable predictor of the plan's KL. That separately coded cells (own codes or residuals) satisfy (b) is an assumption, which the measurements test. A shared cell does not satisfy it: its error $(x_s-x_c)+(Q(x_s)-x_s)$ contains the systematic difference between the two loops, which has nonzero mean, and the rounding error of the source, which the source and every cell that reads it share, so shared cells add interactions that the sum omits. Neither do higher-order terms enter the sum.

The measurements show that the sum predicts the KL nonetheless. Over the ten plans of Table~\ref{tab:additivity} and the four residual designs of Table~\ref{tab:residual}, the measured KL is linear in the sum on a log scale (slope 1.05, correlation 0.98, rank correlation 0.98) and lies within 0.67--1.20\x{} of it. The deviation is systematic rather than random: the five plans without shared cells measure 0.67--0.79\x{} the sum, so the sum is conservative for them, and the nine with 24--28 shared cells measure 0.73--1.20\x{}.
Forty random plans, drawn and with their predictions fixed before any was measured (20 with own codes and residuals only, 20 with shared cells in loops 2--3; predicted 0.73--9.47\x{} \Pz{}), show the same on plans the planner did not choose: their measured KL is linear in the sum (log--log slope 1.11, correlation 0.98, rank correlation 0.94) and lies within 0.65--1.37\x{} of it. All twenty without shared cells measure 0.65--0.94\x{} the sum, and the two largest ratios (1.37\x{} and 1.03\x{}) belong to plans with 27 and 26 shared cells. Two plans re-measured in the same run on another GPU reproduce their earlier KL within 0.4\%.
A two-family correction fitted on the 14 plans of Table~\ref{tab:additivity} and frozen before the random plans were scored, $0.73\,S$ for plans without shared cells and $0.97\,S$ for plans with them ($S$ the sum of single-cell damages), predicts the 40 random plans with a mean absolute error of 15\%, against 28\% for $S$ itself; its largest miss is a factor of 1.41.

Table~\ref{tab:additivity} compares the measured fp32 KL of every plan the planner was asked to build, and of the hand-made plans, with the sum of single-cell damages; every planner plan meets its target, and for the residual designs the sums rank the designs in the same order as their served KL.
Plans a, b, c and d use a menu that also contains residual options; c and d are alternative budgets to the memory- and read-optimized plans.

\begin{table}[h]
\centering\small
\caption{Stored / read: KiB per token (4 loops). Predicted (sum of single-cell damages plus source) and measured fp32 KL / \Pz{}; served KL relative to the served BF16 sharing reference; peak tok/s in vLLM; no V precision, before fusion.}
\label{tab:additivity}
\setlength{\tabcolsep}{4pt}
\begin{tabular}{@{}l>{\raggedright\arraybackslash}p{4.3cm}rrrrr@{}}
\toprule
plan & rule & stored / read & predicted & measured & served KL & tok/s 8K / 16K \\
\midrule
read-optimized & min reads, damage $\le$ pred.\ of INT4 residuals & 215 / 215 & 1.15 & 0.87 & 1.28 & 528.9 / 271.5 \\
memory-optimized & min reads, stored $\le$ 156, damage $\le$ pred.\ of layer-split residuals & 155 / 220 & 1.97 & 1.80 & 1.77 & 535.1 / 280.7 \\
d & min reads, damage $\le$ 1.42 & 210 / 210 & 1.41 & 1.11 & 1.42 & 519.4 / 276.2 \\
c & min damage, stored $\le$ 156 & 156 / 321 & 1.09 & 1.08 & 1.41 & 401.9 / 209.1 \\
a & min damage, read $\le$ 240 & 237 / 240 & 0.67 & 0.47 & 1.01 & 479.8 / 213.0 \\
b & min damage, read $\le$ 312 & 239 / 312 & 0.49 & 0.37 & 0.98 & 398.0 / 179.3 \\
KIVI-4 & --- & 240 / 240 & 0.68 & 0.46 & 1.02 & 494.4 / 249.1 \\
KIVI-4 + loop-3 sharing & --- & 180 / 240 & 1.32 & 1.51 & 1.54 & 505.4 / 264.2 \\
INT4/INT2 + sharing & --- & 156 / 216 & 4.21 & 4.21 & 3.02 & 543.2 / 285.4 \\
INT2 + sharing & --- & 132 / 192 & 9.43 & 8.48 & 5.11 & 604.2 / 319.9 \\
\bottomrule
\end{tabular}
\end{table}

\paragraph{Layer-only allocation.} To test whether the loop axis matters, we solve the same knapsack with one option per layer, shared by loops 1--3, as layer-wise allocators do, on the same profile and codes (an exact dynamic program over the 24 layers), and measure the plans in the fp32 harness in one run with the two shipped plans (16 documents, paired bootstrap).
At the \planR{}'s bytes, the layer-only plan (INT4 or INT2 per layer, 213\,KiB) has 1.30\x{} \Pz{} against 0.86\x{} for the \planR{} in the same run (0.87\x{} in Table~\ref{tab:additivity}), a paired ratio of 1.52\x{}.
At the \planM{}'s bytes, it has 2.58\x{} \Pz{} with loop 3 shared in every layer (154 / 214\,KiB) and 3.35\x{} with sharing as a per-layer option (151.5 / 219\,KiB), against 1.80\x{}: ratios of 1.43\x{} and 1.86\x{}.

\paragraph{Residual designs.}
Table~\ref{tab:residual} lists the four closed-loop residual designs of this analysis (INT4 final loop, residuals for loops 1--2, loop 3 shared), the design family of FlashLoop and ResidualQuant, served with our reader.

\begin{table}[h]
\centering\small
\caption{Residual designs served in vLLM before fusion (stored / read: KiB per token; served KL relative to the served BF16 sharing reference; fp32 KL / \Pz{}; MQ641: $\Delta$ in pp vs BF16). The 16K peak of INT4 residuals ran on the second GPU (\S\ref{sec:eval-setup}).}
\label{tab:residual}
\setlength{\tabcolsep}{4pt}
\begin{tabular}{@{}lrrrrr@{}}
\toprule
design & stored / read & tok/s 8K / 16K & served KL & fp32 KL & MQ641 \\
\midrule
INT4 residuals & 180 / 360 & 370.3 / 192.0 & 1.53 & 1.42 & $-0.8$ \\
loop 1 INT4, loop 2 INT2 & 156 / 336 & 386.3 / 199.7 & 1.88 & 2.12 & $-0.8$ \\
layer split: layers 1--12 INT4, 13--24 INT2 & 156 / 336 & 385.6 / 200.1 & 1.79 & 1.83 & $-2.3$ \\
INT2 residuals & 132 / 312 & 416.8 / 215.9 & 2.97 & 4.38 & $-3.0$ \\
\bottomrule
\end{tabular}
\end{table}

\paragraph{Why the loops compound read errors (P3).}\label{app:p3math}
Proposition~\ref{prop:add}(a) also explains P3. Split a perturbation into $k$ groups of sites, with damages $D_i$ alone and logit errors $e_i$. To second order the joint damage is $\sum_i D_i+\sum_{i<j}\langle e_i,e_j\rangle_F$, so the relative damage of Figure~\ref{fig:p3}(a) is
\[
\frac{D\bigl(\sum_i\delta_i\bigr)}{\sum_i D_i}\;=\;1+\frac{\sum_{i\neq j}\rho_{ij}\sqrt{D_iD_j}}{\sum_i D_i},
\]
where $\rho_{ij}$ is the cosine of $e_i$ and $e_j$ in the Fisher metric: 1 for orthogonal errors and $k$ for $k$ identical ones.
We record every arm's log-probability change on 64 documents and form the Fisher Gram of the single-group errors. The Gram predicts each measured ratio without a joint run: for Quest on Ouro-1.4B grouped by loop, 1.81 predicted against 1.82 measured (95\% CI 1.62--2.02), with a mean pairwise cosine of 0.31 (CI 0.25--0.37); for the same approximated layers grouped into four layer slices that each span all four loops, 1.21 against 1.19 (CI 1.08--1.27), with a mean cosine of 0.07 (CI 0.03--0.09); and for four layer groups of Qwen2.5-1.5B and Qwen3-1.7B, 0.97 and 1.17 against 0.95 and 1.17, with mean cosines of $-0.00$ and 0.06.
The alignment is already present where the errors are injected. The residual-stream error that dropping pages injects at one layer has a cosine of 0.45, 0.71 and 0.82 between loops 1 and 2, 2 and 3, and 3 and 4, but 0.06 between two adjacent layers of one loop (0.04--0.05 on the plain models): the loops apply the same weights to states that converge, so each loop injects nearly the same error, and aligned errors add in amplitude rather than in energy.
Coarse reads compound as well: by loop, LoopKV-style 2-bit reads cost 1.51\x{} the sum of their one-loop damages (CI 1.13--2.10; 1.93\x{} without one outlier document) and 1.39\x{} by layer slice, because decoding each code to its bucket midpoint shrinks the extreme channels of a group by the same amount in every loop. What keeps their damage low is the size of each read's error, not its direction: at equal bytes they inject 4.7\x{} less error energy than dropping on Ouro-1.4B. The advantage of coarse reads over dropping is therefore not specific to looped models, and the plain models show an even larger one; once the outlier document is excluded, it does not grow with the number of loops. What is specific to looped models is the compounding itself.

\section{Additional results}\label{app:extra}

\paragraph{Full sensitivity profile.}
Summed over layers, own INT2 costs 5.6\x{}, 3.0\x{} and 1.9\x{} \Pz{} in loops 1--3 and own INT4 0.36\x{}, 0.11\x{} and 0.05\x{}, and at own INT4, 54 of the 72 history cells cost less than 0.01\x{} \Pz{}; sharing is prohibitive in loop 1 (1074\x{} \Pz{} at layer 1) and cheap in loop 3 (0.69\x{}, summed over its cells).

\paragraph{Proof of Proposition~\ref{prop:bound}.}
Scores use the full INT4 keys, so the weights equal those of the INT4 reader. A decoded value is $z+\sigma c$, and $|c-(4\lfloor c/4\rfloor+1.5)|\le 1.5$ for $c\in\{0,\dots,15\}$. Every cold token has $s_j<m-\tau\ln 2\le s_{\max}-\tau\ln 2$ because $m$ never exceeds the row's true maximum, hence $a_j<2^{-\tau}a_{\max}$. \hfill$\square$

\paragraph{V precision threshold.}
Table~\ref{tab:tau} sweeps $\tau$ on loops 1--2 of the \planU{} and on all loops.
From $\tau{=}10$ upward on loops 1--2, and at $\tau{=}14$ on all loops, the KL change is not significant; smaller $\tau$ costs 0.1--0.76\x{} \Pz{}.

\begin{table}[h]
\centering\small
\caption{V precision threshold sweep (base: KIVI-4 + loop-3 sharing, 1.51\x{} \Pz{}, 1.50\x{} with explicit attention; fp32, 16 documents; $\Delta$ against the same base with explicit attention).}
\label{tab:tau}
\setlength{\tabcolsep}{4pt}
\begin{tabular}{@{}llrrrr@{}}
\toprule
loops & $\tau$ & KL / \Pz{} & $\Delta$ & full-precision V fraction & read KiB \\
\midrule
1--2 & 3 & 2.261 & $+0.760$ & .001 / .000 & 216.0 \\
1--2 & 6 & 1.768 & $+0.267$ & .008 / .002 & 216.1 \\
1--2 & 8 & 1.606 & $+0.105$ & .020 / .007 & 216.3 \\
1--2 & 10 & 1.502 & $+0.001$ & .042 / .024 & 216.8 \\
1--2 & 12 & 1.492 & $-0.009$ & .090 / .067 & 217.9 \\
1--2 & 16 & 1.501 & $-0.000$ & .321 / .303 & 223.5 \\
1--4 & 10 & 1.580 & $+0.080$ & .042 / .024 / .024 / .025 & 193.4 \\
1--4 & 14 & 1.488 & $-0.012$ & .181 / .158 / .167 / .173 & 200.1 \\
\bottomrule
\end{tabular}
\end{table}

\paragraph{Overlap of the loops' hot sets.}
Table~\ref{tab:overlap} compares the tokens each loop attends to above the threshold, on Ouro-1.4B in fp32 over the last 256 query positions of 16 documents (a different sample from Table~\ref{tab:tau}).
Per query row and head, each loop's set is a quarter of the far history at these positions (Table~\ref{tab:tau}'s sample gives about one in six) and the union over four loops 37\%; the hot sets of loop 1 and each later loop share 39--44\% of their union, those of loops 3 and 4 share 76\%.
A precision fixed per token at write time must serve every loop that reads the page: on a page that loops 3 and 4 share it keeps both loops' sets at 4 bits, and a single write-time precision for all four loops would keep their union and read 1.6\x{} the V lo-plane bytes of the per-loop rule at $\tau{=}14$.

\begin{table}[h]
\centering\small
\caption{Loops attend to overlapping but different history tokens, so a per-loop read-time V precision reads fewer lo-plane bytes than a write-time per-token precision that all loops share and that keeps every token any loop marks hot. Ouro-1.4B, fp32, 16 documents, last 256 query positions; hot = attention weight above the $\tau$ threshold; 3s = loop 3's query on loop 4's keys (a shared cell).}
\label{tab:overlap}
\setlength{\tabcolsep}{4pt}
\begin{tabular}{@{}lr@{\qquad}lr@{}}
\toprule
\multicolumn{2}{@{}l}{hot fraction of the far history, $\tau{=}14$} & \multicolumn{2}{l@{}}{Jaccard of hot sets, $\tau{=}14$} \\
\midrule
loop 1 / 2 / 3 / 4 & .27 / .24 / .25 / .25 & loop 1 vs 2 / 3 / 4 & .44 / .40 / .39 \\
union of 4 loops & .37 & loops 2--3 / 2--4 / 3--4 & .63 / .57 / .76 \\
shared cell 3s & .25 & 3s vs 4 & .83 \\
\midrule
\multicolumn{4}{@{}l@{}}{V lo-plane bytes, write-time per-token precision / per-loop read-time rule: 2.2\x{} ($\tau{=}10$), 1.6\x{} ($\tau{=}14$), 1.3\x{} ($\tau{=}18$)} \\
\bottomrule
\end{tabular}
\end{table}

\paragraph{History reader.}
\sys{}'s CUDA history reader streams KV codes at 1.59--1.63\,TB/s, against 1.57--1.62\,TB/s for FlashAttention-2 on BF16, both on an RTX PRO 6000 at 16K (batch 8 and 32).
It uses three techniques.
First, one block of 512 threads covers all 16 heads of a request row and split, so it loads whole rows with coalesced 16-byte vector loads, and it double-buffers the values in registers.
Second, KIVI's per-channel key scale and zero point are constant over a page, so they are multiplied into the query once per page, and the per-token value scale and zero point fold into the attention weight $p_j$ and a running sum.
Third, each code decodes in two instructions, one LOP3 bit operation and one fused multiply-add.
With V precision on real attention scores, the reader's 1.12--1.17\x{} speedup over the plain INT4 reader (\S\ref{sec:impl}) compares with 1.19--1.23\x{} predicted from its bytes.
The fused kernel (Algorithm~\ref{alg:kernel}) keeps this thread mapping with 128 registers and no spills and runs INT2 cells within 1--3\% of the dedicated reader; in a microbenchmark on the RTX PRO 4500, it is 1.19--1.30\x{} faster than the four launches it replaces at batch 1 and 1.00--1.04\x{} at batch 8--48.

\begin{algorithm}[t]
\caption{Fused cell kernel for loop $t$, layer $\ell$ (one launch).}
\label{alg:kernel}
\small
\begin{algorithmic}
\STATE $d \leftarrow$ desc$[t][\ell]$; row $\leftarrow$ blockIdx.x
\IF{blockIdx.y $<$ n\_near}
  \STATE $(o_0, m, l) \leftarrow$ exact BF16 attention over sink + window
  \STATE publish $(o_0, m, l)$; $m_\text{row}\leftarrow m$; set ready flag
\ELSE
  \STATE $m_0 \leftarrow m_\text{row}$ if ready within 4\,$\mu$s, else $-\infty$
  \FOR{page $p$ of this block's split, in $d$.pool}
    \STATE fold K scales of $p$ into $q$; read K codes $\rightarrow$ scores $s_j$
    \STATE update running max $m$, online softmax
    \STATE hot$_j \leftarrow d$.v\_planes\_read${=}2 \wedge s_j {\ge} \max(m, m_0) {-} \tau\ln 2$
    \STATE read V hi plane; read V lo plane only if hot$_j$
    \STATE accumulate $p_j\,\sigma_j\,(4\,\text{hi} + (\text{hot}\ ?\ \text{lo} : 1.5))$
  \ENDFOR
  \STATE publish partial $(o_y, m, l)$
\ENDIF
\STATE \textbf{if} last block of the row \textbf{then} $o \leftarrow$ LSE-merge of all partials
\end{algorithmic}
\end{algorithm}

\paragraph{Where the decode step goes.}
Table~\ref{tab:breakdown} breaks one decode step of the \planU{} at 8K and batch 40 into GPU time per component, with and without V precision.
The history reader is 75\% of the step; V precision makes it 1.06\x{} faster, and the sink seed costs 0.51\,ms.

\begin{table}[h]
\centering\small
\caption{GPU time per decode step (ms), KIVI-4 + loop-3 sharing, 8K, batch 40 (torch.profiler), before fusion: seed, history reader and sink/window attention run as separate kernels; the two V-precision columns take $m_0$ from the window or from the 32 sink keys.}
\label{tab:breakdown}
\setlength{\tabcolsep}{4pt}
\begin{tabular}{@{}lrrr@{}}
\toprule
component & plain & + V precision (window seed) & + V precision (sink seed) \\
\midrule
history reader & 53.69 & 50.38 & 50.71 \\
seed & --- & 2.02 & 0.51 \\
sink + window FlashAttention-2 & 7.17 & 7.21 & 7.21 \\
GEMMs & 8.23 & 8.29 & 8.28 \\
GPU busy & 71.38 & 70.42 & 69.07 \\
\bottomrule
\end{tabular}
\end{table}

\paragraph{BF16 measurement floor.}
In BF16 activations, coding a single cell as an INT2 or INT8 residual against the INT4 source, or perturbing it by $\pm$1 ulp, gives the same KL, 0.9--1.6$\times10^{-4}$; in fp32, INT2 rises 27--561\x{} above the $\pm$1-ulp control while INT8 stays at or below it (Table~\ref{tab:floor}).
Over all cells at once, a random $\pm$1-ulp change gives $1.37{\times}10^{-4}$ in BF16 activations, 0.56\x{} the $2.44{\times}10^{-4}$ that the BF16 sharing configuration has with BF16 activations, but $2.2{\times}10^{-6}$ in fp32 activations, 0.02\x{} \Pz{} ($1.20{\times}10^{-4}$).
This is why all damage in this paper is measured with fp32 activations.

\begin{table}[h]
\centering\small
\caption{KL of single-cell perturbations in BF16 and fp32 activations (16 documents). INT2 and INT8 code the cell as a residual against loop 4's INT4 source, unlike the own codes of Figure~\ref{fig:p1}; $\pm$1 ulp changes the cell's KV by one unit in the last place.}
\label{tab:floor}
\setlength{\tabcolsep}{4pt}
\begin{tabular}{@{}lrrrrrrr@{}}
\toprule
cell & BF16 INT2 & BF16 INT8 & BF16 $\pm$1 ulp & fp32 INT2 & fp32 INT8 & fp32 $\pm$1 ulp & fp32 INT2 / ulp \\
\midrule
loop 1, layer 1 & 1.63e-4 & 1.37e-4 & 1.37e-4 & 2.89e-5 & 7.5e-8 & 8.3e-8 & 348 \\
loop 1, layer 12 & 1.44e-4 & 1.23e-4 & 1.25e-4 & 1.78e-5 & 3.0e-8 & 3.2e-8 & 561 \\
loop 2, layer 12 & 1.18e-4 & 1.10e-4 & 1.08e-4 & 7.53e-6 & 4.2e-9 & 1.7e-8 & 433 \\
loop 3, layer 12 & 0.98e-4 & 0.98e-4 & 0.98e-4 & 8.30e-7 & 1.0e-9 & 7.2e-9 & 115 \\
loop 3, layer 24 & 0.92e-4 & 0.90e-4 & 0.90e-4 & 9.26e-8 & 1.2e-10 & 3.4e-9 & 27 \\
\bottomrule
\end{tabular}
\end{table}

\section{Generality: full results}\label{app:general}
Table~\ref{tab:general} lists the Dory and Huginn plans.
Additivity holds on random cell sets (measured / summed 0.57--1.09); Dory's plans measure 0.49--0.58\x{} their predicted KL, and Huginn's, predicted from interpolated steps, 1.5--2.3\x{}.
V precision at $\tau{=}10$, which keeps 17\% (Dory) and 16\% (Huginn) of far tokens hot, cuts far reads by 16--17\% and changes KL over INT4 by 0.999\x{} on Dory and 0.978\x{} on Huginn; $\tau$ is set per model from an fp32 sweep over 10, 14 and 16.

\begin{table}[h]
\centering\small
\caption{The same knapsack has lower mean KL than uniform plans of equal bytes on Dory and Huginn; against INT2 in the first loops, the CIs separate only for Dory at $-10\%$ (Huginn: 6 short documents). fp32 KL relative to uniform INT4 on all loops; Dory 16 documents, Huginn 6. At Dory $-10\%$ the knapsack has 1.33 [1.25, 1.40] against 2.00 [1.59, 2.35] for INT2 in the first loops (95\% CIs); the other rows' CIs overlap and are not shown. Uniform plans of the same bytes put INT2 in the first or the last loops.}
\label{tab:general}
\setlength{\tabcolsep}{4pt}
{%
\begin{tabular}{@{}llrrr@{}}
\toprule
model & bytes & knapsack & uniform, first & uniform, last \\
\midrule
Dory-4B & $-10\%$ & 1.33 & 2.00 & 29.3 \\
Dory-4B & $-25\%$ & 2.44 & 2.92 & 33.4 \\
Huginn-3.5B & $-10\%$ & 1.04 & 1.20 & 85.0 \\
Huginn-3.5B & $-25\%$ & 2.78 & 3.62 & 98.3 \\
\bottomrule
\end{tabular}}
\end{table}

Table~\ref{tab:ouro26} applies the same profile and knapsack to Ouro-2.6B-Thinking ($T{=}4$, 48 layers, 144 history cells; 8 documents, \Pz{} of this model $5.75{\times}10^{-5}$).
Loop 1 carries the most damage again (own-INT2 loop sums 3.7\,:\,2.5\,:\,1; 1.4B: 3.0\,:\,1.6\,:\,1), and two random 8-cell sets measure 0.87\x{} and 1.00\x{} their summed single-cell damage.
The knapsack's plans measure 0.85\x{} and 0.98\x{} their prediction; with 8 documents the per-plan intervals are wide and overlap.

\begin{table}[h]
\centering\small
\caption{Ouro-2.6B: at equal bytes the memory-optimized plan has 0.48\x{} the KL of the hand-made INT4/INT2 + sharing plan (95\% document-bootstrap CIs overlap; not shown), and the read-optimized plan 1.39\x{} against KIVI-4's 0.82\x{} at 10\% fewer reads. fp32 KL / \Pz{} of Ouro-2.6B, 8 documents; KiB per token (4 loops); V precision at $\tau{=}14$ on all loops, $\Delta$ paired.}
\label{tab:ouro26}
\setlength{\tabcolsep}{4pt}
\begin{tabular}{@{}lrrrr@{}}
\toprule
plan & stored / read & KL & + V precision (read) & $\Delta$ V precision \\
\midrule
KIVI-4 & 480 / 480 & 0.82 & --- & --- \\
KIVI-4 + loop-3 sharing & 360 / 480 & 1.86 & --- & --- \\
INT4/INT2 + sharing & 312 / 432 & 6.30 & --- & --- \\
INT2 + sharing & 264 / 384 & 12.70 & --- & --- \\
read-optimized (knapsack) & 430 / 430 & 1.39 & 1.41 (374.1) & $+0.021$ \\
memory-optimized (knapsack) & 310 / 430 & 3.04 & 3.04 (373.9) & $+0.001$ \\
\bottomrule
\end{tabular}
\end{table}

\section{Dory-4B served}\label{app:dory}
Dory-4B \cite{nvidia2026dory} ships a trained last-loop mode in which all four loops read loop 4's KV, which cuts its KV per token by 2.67\x{} by design.
\sys{} serves on top of this mode, as a plugin to Dory's vLLM fork: when a page of the shared loop-4 KV of a coded full-attention layer leaves the window, it is coded as KIVI-4 (groups of 32), and all four loops read it as shared cells through the same descriptor model as on Ouro; sliding-window layers are unchanged.
The plan step is the same as Ouro's: we measure in fp32 the single-layer damage of each of the nine full-attention planes at INT4 and INT2, and an exact knapsack chooses BF16, INT4 or INT2 per plane, with $\varepsilon$ set to the KL of all nine planes at INT4; the memory-optimized plan minimizes stored bytes under $\varepsilon$, and the read-optimized plan minimizes read bytes under $\varepsilon/2$.
INT2 costs 4--20\x{} INT4's damage on every plane, so neither plan selects it; the memory-optimized plan codes six planes at INT4 and keeps three at BF16 (0.54\x{} BF16's bytes), and the read-optimized plan codes two (0.85\x{}).
Two fixed configurations, which use no profile or knapsack, code the five recurrent full-attention layers (rec5) or all nine full-attention layers (full9).
All four pass the in-serving byte check with no differing byte, and with the fused reader on the RTX PRO 4500 their served KL keeps the fp32 order: read-optimized $4.4{\times}10^{-4}$, memory-optimized $4.9{\times}10^{-4}$, rec5 $6.2{\times}10^{-4}$, full9 $6.6{\times}10^{-4}$ (4 documents, 256-token exits).
Table~\ref{tab:doryplan} gives MATH500 accuracy and the 32\,GB GPU's own peaks under vLLM's graphs, with the fused kernel templated to Dory's four KV heads of dimension 256.
\sys{}-M's largest resident batch is 1.33\x{} and 2.0\x{} last-loop BF16's (64 vs 48 requests at 8K, 48 vs 24 at 16K), and it decodes 3\% and 13\% faster at 8K and 16K; \sys{}-R, which keeps seven of the nine planes in BF16, decodes as fast as the BF16 cache.
On MATH500, no plan loses significantly against last-loop BF16, and with 484 items both planner plans meet a 3-point margin (lower bounds $-2.1$ and $-1.9$).
\begin{table}[h]
\centering\small
\caption{Dory-4B with allocator and fixed plans on one RTX PRO 4500 (32\,GB), graph mode, fused GQA reader. MATH500: 484 items, greedy, 8,192-token cap, paired against last-loop BF16. Bytes: KV stored per token relative to last-loop BF16. Peaks: tok/s at the throughput peak (batch at the peak, which can lie below the largest resident batch).}
\label{tab:doryplan}
\setlength{\tabcolsep}{4pt}
\begin{tabular}{@{}lrrrr@{}}
\toprule
arm & bytes & MATH500 ($\Delta$) & 8K peak & 16K peak \\
\midrule
last-loop BF16 & 1.00 & 91.5 & 352.0 (40) & 219.7 (24) \\
\sys{}-R & 0.85 & 91.5 ($+0.0$) & 352.9 (40) & 221.6 (24) \\
\sys{}-M & 0.54 & 91.7 ($+0.2$) & 362.5 (56) & 247.9 (48) \\
rec5 (fixed) & 0.62 & 91.7 ($+0.2$) & 367.7 (56) & 254.8 (40) \\
full9 (fixed) & 0.31 & --- & 370.3 (48) & 259.6 (56) \\
\bottomrule
\end{tabular}
\end{table}

\section{Serving on a 32\,GB GPU}\label{app:smallcard}
Serving Ouro-1.4B on an RTX PRO 4500 (32\,GB; 24.5\,GiB of KV memory at utilization 0.90), the same final path (fused kernel, whole-step graph, $\tau{=}14$ on loops 1--4) and protocol, with a 1024-token prefill chunk, give the peaks of Table~\ref{tab:smallcard}.
In-serving byte checks (12 pages per arm) find no differing byte on own pages; on the shared INT4 source pages, 0 bytes differ for \sys{}-M, 1 for \sys{}-R and 3--5 for KIVI-4 and KIVI-4 + V.

\begin{table}[h]
\centering\small
\caption{Ouro-1.4B on a 32\,GB RTX PRO 4500, 1024-token prefill chunk: peak decode tok/s (resident batch). The plans' gain over KIVI-4 + V grows with context because they keep more requests resident. 8K: one repeat; 16K and 32K: mean of three fresh processes at the peaks.}
\label{tab:smallcard}
\setlength{\tabcolsep}{4pt}
\begin{tabular}{@{}lrrrrrr@{}}
\toprule
context & KIVI-4 & KIVI-4 + V & \sys{}-R & \sys{}-M & \makecell{gain, own peak\\ (\sys{}-R / -M)} & \makecell{gain, same batch\\ (\sys{}-R / -M)} \\
\midrule
8K & 190.4 (10) & 202.9 (10) & 218.9 (11) & 242.2 (15) & 1.078 / 1.194 & 1.042 / 1.036 \\
16K & 100.7 (5) & 109.1 (5) & 122.3 (6) & 132.6 (8) & 1.121 / 1.215 & 1.037 / 1.031 \\
32K & 47.2 (2) & 50.6 (2) & 60.2 (3) & 65.5 (4) & 1.190 / 1.293 & 1.042 / 1.033 \\
\bottomrule
\end{tabular}
\end{table}

\section{Correctness checks}\label{app:checks}
Served KL agrees with the fake-quantized reference within $\pm$15\%; on the RTX PRO 6000, in-serving byte checks of 12 pages per run never differ on a \sys{} cell's own pages, unfused, fused or graph-replayed; on the shared loop-4 source pages, 0--5 bytes differ per run, on the same pages and in the same counts under every seed mode and plugin copy (one code step at a round-half boundary of the codec), and KIVI-4 differs in at most 2 bytes, own or source; unit tests cover the codec, the reader, mixed prefill and decode batches, and V precision with $\tau{=}\infty$, which is bit-identical to the plain reader.
Plain BF16 vLLM already differs between two identical processes from the first generated token, so served comparisons are paired against one BF16 run.
All fused-kernel numbers use the production seed of V precision, which waits at most 4\,$\mu$s for the sink maximum (\S\ref{sec:design-kernel}); a deterministic seed that keeps the sink maximum makes graph replay match eager execution bit for bit and costs 0.1--0.7\% at the served resident peaks, while one that starts at $-\infty$ costs 5--9\%.

\section{Codec baselines}\label{app:codecs}
\begin{figure}[h]
\centering
\input{figures/v4/fig_codecs.tex}
\caption{No codec reading less than 216\,KiB matches the read-optimized plan's KL (0.87\x{} \Pz{} at 215\,KiB without and 186\,KiB with V precision). Ouro-1.4B, 16 documents, fp32 KL / \Pz{} against history bytes read per token; published recipes in our implementation: TurboQuant without QJL, KVQuant without Fisher weighting. Grey: codecs on all four loops; orange: cross-loop residual codecs (FlashLoop's chain reads all earlier planes); purple and blue: \sys{} plans without and with V precision.}
\label{fig:codecs}
\end{figure}
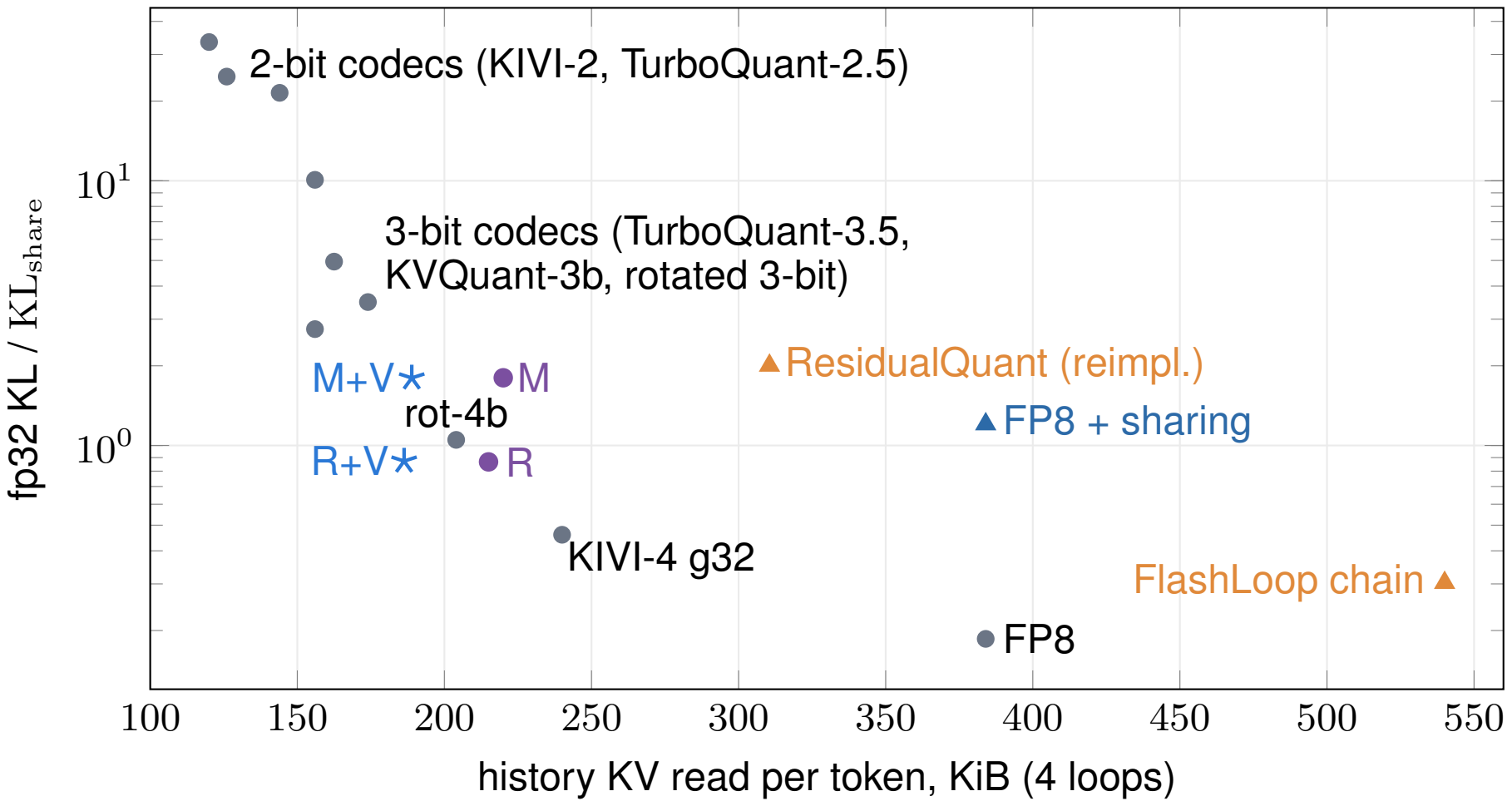

Figure~\ref{fig:codecs} places the plans among published KV codecs in the fp32 harness, and Table~\ref{tab:isocodec} compares them with uniform KIVI-4 designs on the same codes (32-token groups, as served).
The table predicts each design's 16K resident batch from its stored bytes and its peak decode throughput from Eq.~\ref{eq:step} with its reads.
These predictions reproduce the measured 16K throughput ratios over KIVI-4 within 4\%, before and after fusion (2.2\% on the fused path; up to 6.4\% at 8K), although at 16K the served designs keep 2--11\% more requests than the byte ratio predicts, because near the peak a 10\% larger batch raises Eq.~\ref{eq:step}'s throughput by about 2\%.

\begin{table}[h]
\centering\small
\caption{At the 32-token groups we serve, the read-optimized plan has lower KL and fewer reads than KIVI-4 + loop-3 sharing, the memory-optimized plan stores less at a higher KL (1.80\x{} against 1.51\x{} \Pz{}), and stored bytes set the resident batch. Bytes: KiB per token of history (4 loops), stored / read without V precision; KL: fp32 KL / \Pz{} of the stored codes, which V precision changes by at most 1.6\% (\S\ref{sec:design-c3}). Resident batch and peak decode tok/s at 16K, relative to KIVI-4 + V (14 requests, 267.8\,tok/s on the fused path), are predicted for every row with V precision: the batch scales with stored bytes from KIVI-4 + V's 14, and the step time follows Eq.~\ref{eq:step} with the measured reads. Parentheses: measured, from the fused-path block of Table~\ref{tab:tput} and, for the hand-made plan's batch, from its run before fusion.}
\label{tab:isocodec}
\setlength{\tabcolsep}{4pt}
\begin{tabular}{@{}lrrrr@{}}
\toprule
design & bytes & KL & \makecell{resident\\ batch $\times$} & \makecell{tok/s\\ $\times$} \\
\midrule
\multicolumn{5}{@{}l}{\emph{32-token groups with V precision: served}} \\
KIVI-4 + V & 240/240 & 0.46 & 1 & 1 \\
\quad + loop-3 sharing (hand-made) & 180/240 & 1.51 & 1.33 (1.43) & 1.07 \\
\sys{}-R (knapsack) & 215/215 & 0.87 & 1.12 (1.14) & 1.08 (1.08) \\
\sys{}-M (knapsack) & 155/220 & 1.80 & 1.55 (1.71) & 1.16 (1.13) \\
\bottomrule
\end{tabular}
\end{table}

\section{Why integer codes}\label{app:int}
\sys{} uses KIVI-style integer codes.
Decode attention is memory-bound, so FP4 tensor-core throughput does not shorten the step, and FP4 block formats, although they carry less metadata (NVFP4 0.5 bit per element, MXFP4 \cite{rouhani2023mx} 0.25 bit, KIVI 1 bit), are less accurate on this KV (Table~\ref{tab:fp4}): NVFP4 has 2.08\x{} \Pz{} at 216\,KiB and MXFP4 6.40\x{} at 204\,KiB, against 0.46\x{} for KIVI-4 at 240\,KiB and 0.87\x{} for the \planR{}, whose KIVI-style INT4 and INT2 codes read 215\,KiB.
Rotation helps FP4 most but leaves it behind INT4 (NVFP4 0.97\x{}, MXFP4 2.46\x{}, KIVI-4 0.31\x{}).
V precision needs a 2-bit read of a 4-bit code; at four hot cells, reading V's top two bits does 5--10\x{} the KL damage of the full INT4 cell with integer codes, and with NVFP4 codes 2.7--3.6\x{} the damage of INT4's top-two-bit read.

\begin{table}[h]
\centering\small
\caption{4-bit formats on all four loops (fp32 KL / \Pz{}, 16 documents). Bits per element include scales and zero points; KL with Hadamard: the same format after a Hadamard rotation.}
\label{tab:fp4}
\setlength{\tabcolsep}{4pt}
\begin{tabular}{@{}lrrrr@{}}
\toprule
format & bits / element & KiB per token & KL & KL with Hadamard \\
\midrule
KIVI-4, group 32 (asymmetric INT4) & 5.0 & 240 & 0.46 & 0.31 \\
INT4 symmetric, group 16 & 5.0 & 240 & 1.25 & 0.67 \\
NVFP4 (blocks of 16, E4M3 scale) & 4.5 & 216 & 2.08 & 0.97 \\
MXFP4 (blocks of 32, E8M0 scale) & 4.25 & 204 & 6.40 & 2.46 \\
FP8 E4M3 (reference) & 8.0 & 384 & 0.19 & --- \\
\bottomrule
\end{tabular}
\end{table}

\section{Alternatives that did not pay}\label{app:negative}
\textbf{Previous-loop guidance (for C3).} Reading only the previous loop's top quarter of pages at INT4 and the rest at 2 bits costs $+0.69$\x{} \Pz{} over its KIVI-4 + loop-3 sharing base (paired, 16 documents), because the previous loop's top quarter holds 62--67\% of the next loop's attention per head (49--50\% when one page set serves all heads).
\textbf{Aligning unconverged loops (for C2).} Keys of loops 2 and 3 differ mainly by a rotation (norm ratio 0.996, cosine 0.976); a MiniCache-style norm correction recovers 2.7\% of the sharing KL, and fitted rotations raise it from 4.17\x{} to 4.75\x{} \Pz{} (BF16 harness).
\textbf{Decode-time last-step reuse.} Ouro's authors let every loop read the last loop's KV for generated tokens and report 78.85 vs 78.92 on GSM8K and 80.40 vs 82.40 on MATH-500 \cite{zhu2025ouro}. Under one length-controlled protocol in our engine (rotated INT4 reference cache, greedy decoding, the same 484 MATH500 items at every output cap), this reuse loses 44.4 points at an 8K cap, and at 16K it loses 56.0 points on MATH level 5 and 31.7 on AIME. Letting only loop 3 read loop 4's KV, outside an exact recent region, loses 1.0 at 8K and stays within noise at 16K ($+0.7$ and $+1.7$). Long outputs therefore expose what short benchmarks hide, and the early loops cannot share.

\textbf{Fewer loops or early exit.} The exit head after loops 1, 2 and 3 agrees with loop 4 on only 90\%, 96\% and 99\% of tokens, so stopping after fewer loops changes the model's output. %

\end{document}

%% file: figures/v4/gen_cells.tex
\newcommand{\SensCells}{%
\fill[fill={rgb,255:red,205;green,93;blue,102}] (0,2) rectangle ++(0.92,0.92);%
\fill[fill={rgb,255:red,208;green,103;blue,110}] (1,2) rectangle ++(0.92,0.92);%
\fill[fill={rgb,255:red,204;green,88;blue,97}] (2,2) rectangle ++(0.92,0.92);%
\fill[fill={rgb,255:red,217;green,142;blue,146}] (3,2) rectangle ++(0.92,0.92);%
\fill[fill={rgb,255:red,215;green,134;blue,138}] (4,2) rectangle ++(0.92,0.92);%
\fill[fill={rgb,255:red,220;green,153;blue,156}] (5,2) rectangle ++(0.92,0.92);%
\fill[fill={rgb,255:red,195;green,52;blue,65}] (6,2) rectangle ++(0.92,0.92);%
\fill[fill={rgb,255:red,192;green,39;blue,53}] (7,2) rectangle ++(0.92,0.92);%
\fill[fill={rgb,255:red,216;green,136;blue,141}] (8,2) rectangle ++(0.92,0.92);%
\fill[fill={rgb,255:red,194;green,47;blue,61}] (9,2) rectangle ++(0.92,0.92);%
\fill[fill={rgb,255:red,204;green,86;blue,96}] (10,2) rectangle ++(0.92,0.92);%
\fill[fill={rgb,255:red,214;green,128;blue,133}] (11,2) rectangle ++(0.92,0.92);%
\fill[fill={rgb,255:red,210;green,111;blue,118}] (12,2) rectangle ++(0.92,0.92);%
\fill[fill={rgb,255:red,218;green,145;blue,148}] (13,2) rectangle ++(0.92,0.92);%
\fill[fill={rgb,255:red,216;green,138;blue,142}] (14,2) rectangle ++(0.92,0.92);%
\fill[fill={rgb,255:red,220;green,155;blue,157}] (15,2) rectangle ++(0.92,0.92);%
\fill[fill={rgb,255:red,227;green,183;blue,182}] (16,2) rectangle ++(0.92,0.92);%
\fill[fill={rgb,255:red,226;green,177;blue,177}] (17,2) rectangle ++(0.92,0.92);%
\fill[fill={rgb,255:red,240;green,234;blue,229}] (18,2) rectangle ++(0.92,0.92);%
\fill[fill={rgb,255:red,232;green,205;blue,202}] (19,2) rectangle ++(0.92,0.92);%
\fill[fill={rgb,255:red,230;green,193;blue,191}] (20,2) rectangle ++(0.92,0.92);%
\fill[fill={rgb,255:red,226;green,226;blue,227}] (21,2) rectangle ++(0.92,0.92);%
\fill[fill={rgb,255:red,180;green,194;blue,215}] (22,2) rectangle ++(0.92,0.92);%
\fill[fill={rgb,255:red,222;green,224;blue,226}] (23,2) rectangle ++(0.92,0.92);%
\fill[fill={rgb,255:red,107;green,143;blue,198}] (0,1) rectangle ++(0.92,0.92);%
\fill[fill={rgb,255:red,236;green,218;blue,214}] (1,1) rectangle ++(0.92,0.92);%
\fill[fill={rgb,255:red,238;green,226;blue,221}] (2,1) rectangle ++(0.92,0.92);%
\fill[fill={rgb,255:red,238;green,226;blue,221}] (3,1) rectangle ++(0.92,0.92);%
\fill[fill={rgb,255:red,226;green,226;blue,227}] (4,1) rectangle ++(0.92,0.92);%
\fill[fill={rgb,255:red,231;green,230;blue,228}] (5,1) rectangle ++(0.92,0.92);%
\fill[fill={rgb,255:red,207;green,102;blue,110}] (6,1) rectangle ++(0.92,0.92);%
\fill[fill={rgb,255:red,197;green,60;blue,72}] (7,1) rectangle ++(0.92,0.92);%
\fill[fill={rgb,255:red,216;green,137;blue,141}] (8,1) rectangle ++(0.92,0.92);%
\fill[fill={rgb,255:red,210;green,113;blue,119}] (9,1) rectangle ++(0.92,0.92);%
\fill[fill={rgb,255:red,201;green,75;blue,85}] (10,1) rectangle ++(0.92,0.92);%
\fill[fill={rgb,255:red,216;green,138;blue,142}] (11,1) rectangle ++(0.92,0.92);%
\fill[fill={rgb,255:red,212;green,120;blue,126}] (12,1) rectangle ++(0.92,0.92);%
\fill[fill={rgb,255:red,219;green,151;blue,154}] (13,1) rectangle ++(0.92,0.92);%
\fill[fill={rgb,255:red,217;green,143;blue,146}] (14,1) rectangle ++(0.92,0.92);%
\fill[fill={rgb,255:red,223;green,165;blue,166}] (15,1) rectangle ++(0.92,0.92);%
\fill[fill={rgb,255:red,229;green,192;blue,190}] (16,1) rectangle ++(0.92,0.92);%
\fill[fill={rgb,255:red,227;green,183;blue,183}] (17,1) rectangle ++(0.92,0.92);%
\fill[fill={rgb,255:red,198;green,207;blue,220}] (18,1) rectangle ++(0.92,0.92);%
\fill[fill={rgb,255:red,234;green,213;blue,209}] (19,1) rectangle ++(0.92,0.92);%
\fill[fill={rgb,255:red,239;green,230;blue,225}] (20,1) rectangle ++(0.92,0.92);%
\fill[fill={rgb,255:red,184;green,197;blue,216}] (21,1) rectangle ++(0.92,0.92);%
\fill[fill={rgb,255:red,165;green,183;blue,212}] (22,1) rectangle ++(0.92,0.92);%
\fill[fill={rgb,255:red,184;green,197;blue,216}] (23,1) rectangle ++(0.92,0.92);%
\fill[fill={rgb,255:red,59;green,110;blue,186}] (0,0) rectangle ++(0.92,0.92);%
\fill[fill={rgb,255:red,180;green,194;blue,215}] (1,0) rectangle ++(0.92,0.92);%
\fill[fill={rgb,255:red,201;green,209;blue,220}] (2,0) rectangle ++(0.92,0.92);%
\fill[fill={rgb,255:red,211;green,216;blue,223}] (3,0) rectangle ++(0.92,0.92);%
\fill[fill={rgb,255:red,218;green,221;blue,225}] (4,0) rectangle ++(0.92,0.92);%
\fill[fill={rgb,255:red,184;green,197;blue,216}] (5,0) rectangle ++(0.92,0.92);%
\fill[fill={rgb,255:red,218;green,147;blue,150}] (6,0) rectangle ++(0.92,0.92);%
\fill[fill={rgb,255:red,206;green,96;blue,104}] (7,0) rectangle ++(0.92,0.92);%
\fill[fill={rgb,255:red,224;green,169;blue,170}] (8,0) rectangle ++(0.92,0.92);%
\fill[fill={rgb,255:red,217;green,143;blue,147}] (9,0) rectangle ++(0.92,0.92);%
\fill[fill={rgb,255:red,210;green,113;blue,119}] (10,0) rectangle ++(0.92,0.92);%
\fill[fill={rgb,255:red,226;green,179;blue,179}] (11,0) rectangle ++(0.92,0.92);%
\fill[fill={rgb,255:red,218;green,147;blue,150}] (12,0) rectangle ++(0.92,0.92);%
\fill[fill={rgb,255:red,225;green,174;blue,175}] (13,0) rectangle ++(0.92,0.92);%
\fill[fill={rgb,255:red,220;green,154;blue,156}] (14,0) rectangle ++(0.92,0.92);%
\fill[fill={rgb,255:red,227;green,183;blue,183}] (15,0) rectangle ++(0.92,0.92);%
\fill[fill={rgb,255:red,228;green,186;blue,185}] (16,0) rectangle ++(0.92,0.92);%
\fill[fill={rgb,255:red,228;green,187;blue,186}] (17,0) rectangle ++(0.92,0.92);%
\fill[fill={rgb,255:red,204;green,211;blue,221}] (18,0) rectangle ++(0.92,0.92);%
\fill[fill={rgb,255:red,230;green,195;blue,194}] (19,0) rectangle ++(0.92,0.92);%
\fill[fill={rgb,255:red,235;green,214;blue,210}] (20,0) rectangle ++(0.92,0.92);%
\fill[fill={rgb,255:red,204;green,211;blue,221}] (21,0) rectangle ++(0.92,0.92);%
\fill[fill={rgb,255:red,191;green,202;blue,218}] (22,0) rectangle ++(0.92,0.92);%
\fill[fill={rgb,255:red,226;green,226;blue,227}] (23,0) rectangle ++(0.92,0.92);%
}
\newcommand{\PlanCellsdtwo}{%
\fill[pD] (0,3) rectangle ++(0.92,0.92);%
\fill[pD] (1,3) rectangle ++(0.92,0.92);%
\fill[pD] (2,3) rectangle ++(0.92,0.92);%
\fill[pD] (3,3) rectangle ++(0.92,0.92);%
\fill[pD] (4,3) rectangle ++(0.92,0.92);%
\fill[pD] (5,3) rectangle ++(0.92,0.92);%
\fill[pD] (6,3) rectangle ++(0.92,0.92);%
\fill[pD] (7,3) rectangle ++(0.92,0.92);%
\fill[pD] (8,3) rectangle ++(0.92,0.92);%
\fill[pD] (9,3) rectangle ++(0.92,0.92);%
\fill[pD] (10,3) rectangle ++(0.92,0.92);%
\fill[pD] (11,3) rectangle ++(0.92,0.92);%
\fill[pD] (12,3) rectangle ++(0.92,0.92);%
\fill[pD] (13,3) rectangle ++(0.92,0.92);%
\fill[pD] (14,3) rectangle ++(0.92,0.92);%
\fill[pD] (15,3) rectangle ++(0.92,0.92);%
\fill[pD] (16,3) rectangle ++(0.92,0.92);%
\fill[pD] (17,3) rectangle ++(0.92,0.92);%
\fill[pd] (18,3) rectangle ++(0.92,0.92);%
\fill[pD] (19,3) rectangle ++(0.92,0.92);%
\fill[pD] (20,3) rectangle ++(0.92,0.92);%
\fill[pd] (21,3) rectangle ++(0.92,0.92);%
\fill[pd] (22,3) rectangle ++(0.92,0.92);%
\fill[pd] (23,3) rectangle ++(0.92,0.92);%
\fill[pd] (0,2) rectangle ++(0.92,0.92);%
\fill[pd] (1,2) rectangle ++(0.92,0.92);%
\fill[pd] (2,2) rectangle ++(0.92,0.92);%
\fill[pd] (3,2) rectangle ++(0.92,0.92);%
\fill[pd] (4,2) rectangle ++(0.92,0.92);%
\fill[pd] (5,2) rectangle ++(0.92,0.92);%
\fill[pD] (6,2) rectangle ++(0.92,0.92);%
\fill[pD] (7,2) rectangle ++(0.92,0.92);%
\fill[pD] (8,2) rectangle ++(0.92,0.92);%
\fill[pD] (9,2) rectangle ++(0.92,0.92);%
\fill[pD] (10,2) rectangle ++(0.92,0.92);%
\fill[pD] (11,2) rectangle ++(0.92,0.92);%
\fill[pD] (12,2) rectangle ++(0.92,0.92);%
\fill[pD] (13,2) rectangle ++(0.92,0.92);%
\fill[pD] (14,2) rectangle ++(0.92,0.92);%
\fill[pD] (15,2) rectangle ++(0.92,0.92);%
\fill[pD] (16,2) rectangle ++(0.92,0.92);%
\fill[pD] (17,2) rectangle ++(0.92,0.92);%
\fill[pd] (18,2) rectangle ++(0.92,0.92);%
\fill[pD] (19,2) rectangle ++(0.92,0.92);%
\fill[pd] (20,2) rectangle ++(0.92,0.92);%
\fill[pd] (21,2) rectangle ++(0.92,0.92);%
\fill[pd] (22,2) rectangle ++(0.92,0.92);%
\fill[pd] (23,2) rectangle ++(0.92,0.92);%
\fill[pd] (0,1) rectangle ++(0.92,0.92);%
\fill[pd] (1,1) rectangle ++(0.92,0.92);%
\fill[pd] (2,1) rectangle ++(0.92,0.92);%
\fill[pd] (3,1) rectangle ++(0.92,0.92);%
\fill[pd] (4,1) rectangle ++(0.92,0.92);%
\fill[pd] (5,1) rectangle ++(0.92,0.92);%
\fill[pD] (6,1) rectangle ++(0.92,0.92);%
\fill[pD] (7,1) rectangle ++(0.92,0.92);%
\fill[pD] (8,1) rectangle ++(0.92,0.92);%
\fill[pD] (9,1) rectangle ++(0.92,0.92);%
\fill[pD] (10,1) rectangle ++(0.92,0.92);%
\fill[pD] (11,1) rectangle ++(0.92,0.92);%
\fill[pD] (12,1) rectangle ++(0.92,0.92);%
\fill[pD] (13,1) rectangle ++(0.92,0.92);%
\fill[pD] (14,1) rectangle ++(0.92,0.92);%
\fill[pD] (15,1) rectangle ++(0.92,0.92);%
\fill[pD] (16,1) rectangle ++(0.92,0.92);%
\fill[pD] (17,1) rectangle ++(0.92,0.92);%
\fill[pd] (18,1) rectangle ++(0.92,0.92);%
\fill[pD] (19,1) rectangle ++(0.92,0.92);%
\fill[pD] (20,1) rectangle ++(0.92,0.92);%
\fill[pd] (21,1) rectangle ++(0.92,0.92);%
\fill[pd] (22,1) rectangle ++(0.92,0.92);%
\fill[pd] (23,1) rectangle ++(0.92,0.92);%
\fill[pA] (0,0) rectangle ++(0.92,0.92);%
\fill[pA] (1,0) rectangle ++(0.92,0.92);%
\fill[pA] (2,0) rectangle ++(0.92,0.92);%
\fill[pA] (3,0) rectangle ++(0.92,0.92);%
\fill[pA] (4,0) rectangle ++(0.92,0.92);%
\fill[pA] (5,0) rectangle ++(0.92,0.92);%
\fill[pA] (6,0) rectangle ++(0.92,0.92);%
\fill[pA] (7,0) rectangle ++(0.92,0.92);%
\fill[pA] (8,0) rectangle ++(0.92,0.92);%
\fill[pA] (9,0) rectangle ++(0.92,0.92);%
\fill[pA] (10,0) rectangle ++(0.92,0.92);%
\fill[pA] (11,0) rectangle ++(0.92,0.92);%
\fill[pA] (12,0) rectangle ++(0.92,0.92);%
\fill[pA] (13,0) rectangle ++(0.92,0.92);%
\fill[pA] (14,0) rectangle ++(0.92,0.92);%
\fill[pA] (15,0) rectangle ++(0.92,0.92);%
\fill[pA] (16,0) rectangle ++(0.92,0.92);%
\fill[pA] (17,0) rectangle ++(0.92,0.92);%
\fill[pA] (18,0) rectangle ++(0.92,0.92);%
\fill[pA] (19,0) rectangle ++(0.92,0.92);%
\fill[pA] (20,0) rectangle ++(0.92,0.92);%
\fill[pA] (21,0) rectangle ++(0.92,0.92);%
\fill[pA] (22,0) rectangle ++(0.92,0.92);%
\fill[pA] (23,0) rectangle ++(0.92,0.92);%
}
\newcommand{\PlanCellsctwo}{%
\fill[pD] (0,3) rectangle ++(0.92,0.92);%
\fill[pD] (1,3) rectangle ++(0.92,0.92);%
\fill[pD] (2,3) rectangle ++(0.92,0.92);%
\fill[pD] (3,3) rectangle ++(0.92,0.92);%
\fill[pD] (4,3) rectangle ++(0.92,0.92);%
\fill[pD] (5,3) rectangle ++(0.92,0.92);%
\fill[pD] (6,3) rectangle ++(0.92,0.92);%
\fill[pD] (7,3) rectangle ++(0.92,0.92);%
\fill[pD] (8,3) rectangle ++(0.92,0.92);%
\fill[pD] (9,3) rectangle ++(0.92,0.92);%
\fill[pD] (10,3) rectangle ++(0.92,0.92);%
\fill[pD] (11,3) rectangle ++(0.92,0.92);%
\fill[pD] (12,3) rectangle ++(0.92,0.92);%
\fill[pD] (13,3) rectangle ++(0.92,0.92);%
\fill[pD] (14,3) rectangle ++(0.92,0.92);%
\fill[pD] (15,3) rectangle ++(0.92,0.92);%
\fill[pd] (16,3) rectangle ++(0.92,0.92);%
\fill[pd] (17,3) rectangle ++(0.92,0.92);%
\fill[pP] (18,3) rectangle ++(0.92,0.92);%
\fill[pd] (19,3) rectangle ++(0.92,0.92);%
\fill[pd] (20,3) rectangle ++(0.92,0.92);%
\fill[pd] (21,3) rectangle ++(0.92,0.92);%
\fill[pd] (22,3) rectangle ++(0.92,0.92);%
\fill[pd] (23,3) rectangle ++(0.92,0.92);%
\fill[pd] (0,2) rectangle ++(0.92,0.92);%
\fill[pd] (1,2) rectangle ++(0.92,0.92);%
\fill[pd] (2,2) rectangle ++(0.92,0.92);%
\fill[pd] (3,2) rectangle ++(0.92,0.92);%
\fill[pd] (4,2) rectangle ++(0.92,0.92);%
\fill[pd] (5,2) rectangle ++(0.92,0.92);%
\fill[pD] (6,2) rectangle ++(0.92,0.92);%
\fill[pD] (7,2) rectangle ++(0.92,0.92);%
\fill[pD] (8,2) rectangle ++(0.92,0.92);%
\fill[pD] (9,2) rectangle ++(0.92,0.92);%
\fill[pD] (10,2) rectangle ++(0.92,0.92);%
\fill[pD] (11,2) rectangle ++(0.92,0.92);%
\fill[pD] (12,2) rectangle ++(0.92,0.92);%
\fill[pD] (13,2) rectangle ++(0.92,0.92);%
\fill[pD] (14,2) rectangle ++(0.92,0.92);%
\fill[pd] (15,2) rectangle ++(0.92,0.92);%
\fill[pP] (16,2) rectangle ++(0.92,0.92);%
\fill[pd] (17,2) rectangle ++(0.92,0.92);%
\fill[pP] (18,2) rectangle ++(0.92,0.92);%
\fill[pP] (19,2) rectangle ++(0.92,0.92);%
\fill[pP] (20,2) rectangle ++(0.92,0.92);%
\fill[pP] (21,2) rectangle ++(0.92,0.92);%
\fill[pd] (22,2) rectangle ++(0.92,0.92);%
\fill[pd] (23,2) rectangle ++(0.92,0.92);%
\fill[pd] (0,1) rectangle ++(0.92,0.92);%
\fill[pd] (1,1) rectangle ++(0.92,0.92);%
\fill[pd] (2,1) rectangle ++(0.92,0.92);%
\fill[pP] (3,1) rectangle ++(0.92,0.92);%
\fill[pP] (4,1) rectangle ++(0.92,0.92);%
\fill[pP] (5,1) rectangle ++(0.92,0.92);%
\fill[pP] (6,1) rectangle ++(0.92,0.92);%
\fill[pP] (7,1) rectangle ++(0.92,0.92);%
\fill[pP] (8,1) rectangle ++(0.92,0.92);%
\fill[pP] (9,1) rectangle ++(0.92,0.92);%
\fill[pD] (10,1) rectangle ++(0.92,0.92);%
\fill[pP] (11,1) rectangle ++(0.92,0.92);%
\fill[pP] (12,1) rectangle ++(0.92,0.92);%
\fill[pP] (13,1) rectangle ++(0.92,0.92);%
\fill[pP] (14,1) rectangle ++(0.92,0.92);%
\fill[pP] (15,1) rectangle ++(0.92,0.92);%
\fill[pP] (16,1) rectangle ++(0.92,0.92);%
\fill[pP] (17,1) rectangle ++(0.92,0.92);%
\fill[pP] (18,1) rectangle ++(0.92,0.92);%
\fill[pP] (19,1) rectangle ++(0.92,0.92);%
\fill[pP] (20,1) rectangle ++(0.92,0.92);%
\fill[pP] (21,1) rectangle ++(0.92,0.92);%
\fill[pP] (22,1) rectangle ++(0.92,0.92);%
\fill[pP] (23,1) rectangle ++(0.92,0.92);%
\fill[pA] (0,0) rectangle ++(0.92,0.92);%
\fill[pA] (1,0) rectangle ++(0.92,0.92);%
\fill[pA] (2,0) rectangle ++(0.92,0.92);%
\fill[pA] (3,0) rectangle ++(0.92,0.92);%
\fill[pA] (4,0) rectangle ++(0.92,0.92);%
\fill[pA] (5,0) rectangle ++(0.92,0.92);%
\fill[pA] (6,0) rectangle ++(0.92,0.92);%
\fill[pA] (7,0) rectangle ++(0.92,0.92);%
\fill[pA] (8,0) rectangle ++(0.92,0.92);%
\fill[pA] (9,0) rectangle ++(0.92,0.92);%
\fill[pA] (10,0) rectangle ++(0.92,0.92);%
\fill[pA] (11,0) rectangle ++(0.92,0.92);%
\fill[pA] (12,0) rectangle ++(0.92,0.92);%
\fill[pA] (13,0) rectangle ++(0.92,0.92);%
\fill[pA] (14,0) rectangle ++(0.92,0.92);%
\fill[pA] (15,0) rectangle ++(0.92,0.92);%
\fill[pA] (16,0) rectangle ++(0.92,0.92);%
\fill[pA] (17,0) rectangle ++(0.92,0.92);%
\fill[pA] (18,0) rectangle ++(0.92,0.92);%
\fill[pA] (19,0) rectangle ++(0.92,0.92);%
\fill[pA] (20,0) rectangle ++(0.92,0.92);%
\fill[pA] (21,0) rectangle ++(0.92,0.92);%
\fill[pA] (22,0) rectangle ++(0.92,0.92);%
\fill[pA] (23,0) rectangle ++(0.92,0.92);%
}
\newcommand{\PlanCellsuniformshared}{%
\fill[pD] (0,3) rectangle ++(0.92,0.92);%
\fill[pD] (1,3) rectangle ++(0.92,0.92);%
\fill[pD] (2,3) rectangle ++(0.92,0.92);%
\fill[pD] (3,3) rectangle ++(0.92,0.92);%
\fill[pD] (4,3) rectangle ++(0.92,0.92);%
\fill[pD] (5,3) rectangle ++(0.92,0.92);%
\fill[pD] (6,3) rectangle ++(0.92,0.92);%
\fill[pD] (7,3) rectangle ++(0.92,0.92);%
\fill[pD] (8,3) rectangle ++(0.92,0.92);%
\fill[pD] (9,3) rectangle ++(0.92,0.92);%
\fill[pD] (10,3) rectangle ++(0.92,0.92);%
\fill[pD] (11,3) rectangle ++(0.92,0.92);%
\fill[pD] (12,3) rectangle ++(0.92,0.92);%
\fill[pD] (13,3) rectangle ++(0.92,0.92);%
\fill[pD] (14,3) rectangle ++(0.92,0.92);%
\fill[pD] (15,3) rectangle ++(0.92,0.92);%
\fill[pD] (16,3) rectangle ++(0.92,0.92);%
\fill[pD] (17,3) rectangle ++(0.92,0.92);%
\fill[pD] (18,3) rectangle ++(0.92,0.92);%
\fill[pD] (19,3) rectangle ++(0.92,0.92);%
\fill[pD] (20,3) rectangle ++(0.92,0.92);%
\fill[pD] (21,3) rectangle ++(0.92,0.92);%
\fill[pD] (22,3) rectangle ++(0.92,0.92);%
\fill[pD] (23,3) rectangle ++(0.92,0.92);%
\fill[pD] (0,2) rectangle ++(0.92,0.92);%
\fill[pD] (1,2) rectangle ++(0.92,0.92);%
\fill[pD] (2,2) rectangle ++(0.92,0.92);%
\fill[pD] (3,2) rectangle ++(0.92,0.92);%
\fill[pD] (4,2) rectangle ++(0.92,0.92);%
\fill[pD] (5,2) rectangle ++(0.92,0.92);%
\fill[pD] (6,2) rectangle ++(0.92,0.92);%
\fill[pD] (7,2) rectangle ++(0.92,0.92);%
\fill[pD] (8,2) rectangle ++(0.92,0.92);%
\fill[pD] (9,2) rectangle ++(0.92,0.92);%
\fill[pD] (10,2) rectangle ++(0.92,0.92);%
\fill[pD] (11,2) rectangle ++(0.92,0.92);%
\fill[pD] (12,2) rectangle ++(0.92,0.92);%
\fill[pD] (13,2) rectangle ++(0.92,0.92);%
\fill[pD] (14,2) rectangle ++(0.92,0.92);%
\fill[pD] (15,2) rectangle ++(0.92,0.92);%
\fill[pD] (16,2) rectangle ++(0.92,0.92);%
\fill[pD] (17,2) rectangle ++(0.92,0.92);%
\fill[pD] (18,2) rectangle ++(0.92,0.92);%
\fill[pD] (19,2) rectangle ++(0.92,0.92);%
\fill[pD] (20,2) rectangle ++(0.92,0.92);%
\fill[pD] (21,2) rectangle ++(0.92,0.92);%
\fill[pD] (22,2) rectangle ++(0.92,0.92);%
\fill[pD] (23,2) rectangle ++(0.92,0.92);%
\fill[pP] (0,1) rectangle ++(0.92,0.92);%
\fill[pP] (1,1) rectangle ++(0.92,0.92);%
\fill[pP] (2,1) rectangle ++(0.92,0.92);%
\fill[pP] (3,1) rectangle ++(0.92,0.92);%
\fill[pP] (4,1) rectangle ++(0.92,0.92);%
\fill[pP] (5,1) rectangle ++(0.92,0.92);%
\fill[pP] (6,1) rectangle ++(0.92,0.92);%
\fill[pP] (7,1) rectangle ++(0.92,0.92);%
\fill[pP] (8,1) rectangle ++(0.92,0.92);%
\fill[pP] (9,1) rectangle ++(0.92,0.92);%
\fill[pP] (10,1) rectangle ++(0.92,0.92);%
\fill[pP] (11,1) rectangle ++(0.92,0.92);%
\fill[pP] (12,1) rectangle ++(0.92,0.92);%
\fill[pP] (13,1) rectangle ++(0.92,0.92);%
\fill[pP] (14,1) rectangle ++(0.92,0.92);%
\fill[pP] (15,1) rectangle ++(0.92,0.92);%
\fill[pP] (16,1) rectangle ++(0.92,0.92);%
\fill[pP] (17,1) rectangle ++(0.92,0.92);%
\fill[pP] (18,1) rectangle ++(0.92,0.92);%
\fill[pP] (19,1) rectangle ++(0.92,0.92);%
\fill[pP] (20,1) rectangle ++(0.92,0.92);%
\fill[pP] (21,1) rectangle ++(0.92,0.92);%
\fill[pP] (22,1) rectangle ++(0.92,0.92);%
\fill[pP] (23,1) rectangle ++(0.92,0.92);%
\fill[pA] (0,0) rectangle ++(0.92,0.92);%
\fill[pA] (1,0) rectangle ++(0.92,0.92);%
\fill[pA] (2,0) rectangle ++(0.92,0.92);%
\fill[pA] (3,0) rectangle ++(0.92,0.92);%
\fill[pA] (4,0) rectangle ++(0.92,0.92);%
\fill[pA] (5,0) rectangle ++(0.92,0.92);%
\fill[pA] (6,0) rectangle ++(0.92,0.92);%
\fill[pA] (7,0) rectangle ++(0.92,0.92);%
\fill[pA] (8,0) rectangle ++(0.92,0.92);%
\fill[pA] (9,0) rectangle ++(0.92,0.92);%
\fill[pA] (10,0) rectangle ++(0.92,0.92);%
\fill[pA] (11,0) rectangle ++(0.92,0.92);%
\fill[pA] (12,0) rectangle ++(0.92,0.92);%
\fill[pA] (13,0) rectangle ++(0.92,0.92);%
\fill[pA] (14,0) rectangle ++(0.92,0.92);%
\fill[pA] (15,0) rectangle ++(0.92,0.92);%
\fill[pA] (16,0) rectangle ++(0.92,0.92);%
\fill[pA] (17,0) rectangle ++(0.92,0.92);%
\fill[pA] (18,0) rectangle ++(0.92,0.92);%
\fill[pA] (19,0) rectangle ++(0.92,0.92);%
\fill[pA] (20,0) rectangle ++(0.92,0.92);%
\fill[pA] (21,0) rectangle ++(0.92,0.92);%
\fill[pA] (22,0) rectangle ++(0.92,0.92);%
\fill[pA] (23,0) rectangle ++(0.92,0.92);%
}
\newcommand{\PlanCellsuniform}{%
\fill[pD] (0,3) rectangle ++(0.92,0.92);%
\fill[pD] (1,3) rectangle ++(0.92,0.92);%
\fill[pD] (2,3) rectangle ++(0.92,0.92);%
\fill[pD] (3,3) rectangle ++(0.92,0.92);%
\fill[pD] (4,3) rectangle ++(0.92,0.92);%
\fill[pD] (5,3) rectangle ++(0.92,0.92);%
\fill[pD] (6,3) rectangle ++(0.92,0.92);%
\fill[pD] (7,3) rectangle ++(0.92,0.92);%
\fill[pD] (8,3) rectangle ++(0.92,0.92);%
\fill[pD] (9,3) rectangle ++(0.92,0.92);%
\fill[pD] (10,3) rectangle ++(0.92,0.92);%
\fill[pD] (11,3) rectangle ++(0.92,0.92);%
\fill[pD] (12,3) rectangle ++(0.92,0.92);%
\fill[pD] (13,3) rectangle ++(0.92,0.92);%
\fill[pD] (14,3) rectangle ++(0.92,0.92);%
\fill[pD] (15,3) rectangle ++(0.92,0.92);%
\fill[pD] (16,3) rectangle ++(0.92,0.92);%
\fill[pD] (17,3) rectangle ++(0.92,0.92);%
\fill[pD] (18,3) rectangle ++(0.92,0.92);%
\fill[pD] (19,3) rectangle ++(0.92,0.92);%
\fill[pD] (20,3) rectangle ++(0.92,0.92);%
\fill[pD] (21,3) rectangle ++(0.92,0.92);%
\fill[pD] (22,3) rectangle ++(0.92,0.92);%
\fill[pD] (23,3) rectangle ++(0.92,0.92);%
\fill[pD] (0,2) rectangle ++(0.92,0.92);%
\fill[pD] (1,2) rectangle ++(0.92,0.92);%
\fill[pD] (2,2) rectangle ++(0.92,0.92);%
\fill[pD] (3,2) rectangle ++(0.92,0.92);%
\fill[pD] (4,2) rectangle ++(0.92,0.92);%
\fill[pD] (5,2) rectangle ++(0.92,0.92);%
\fill[pD] (6,2) rectangle ++(0.92,0.92);%
\fill[pD] (7,2) rectangle ++(0.92,0.92);%
\fill[pD] (8,2) rectangle ++(0.92,0.92);%
\fill[pD] (9,2) rectangle ++(0.92,0.92);%
\fill[pD] (10,2) rectangle ++(0.92,0.92);%
\fill[pD] (11,2) rectangle ++(0.92,0.92);%
\fill[pD] (12,2) rectangle ++(0.92,0.92);%
\fill[pD] (13,2) rectangle ++(0.92,0.92);%
\fill[pD] (14,2) rectangle ++(0.92,0.92);%
\fill[pD] (15,2) rectangle ++(0.92,0.92);%
\fill[pD] (16,2) rectangle ++(0.92,0.92);%
\fill[pD] (17,2) rectangle ++(0.92,0.92);%
\fill[pD] (18,2) rectangle ++(0.92,0.92);%
\fill[pD] (19,2) rectangle ++(0.92,0.92);%
\fill[pD] (20,2) rectangle ++(0.92,0.92);%
\fill[pD] (21,2) rectangle ++(0.92,0.92);%
\fill[pD] (22,2) rectangle ++(0.92,0.92);%
\fill[pD] (23,2) rectangle ++(0.92,0.92);%
\fill[pD] (0,1) rectangle ++(0.92,0.92);%
\fill[pD] (1,1) rectangle ++(0.92,0.92);%
\fill[pD] (2,1) rectangle ++(0.92,0.92);%
\fill[pD] (3,1) rectangle ++(0.92,0.92);%
\fill[pD] (4,1) rectangle ++(0.92,0.92);%
\fill[pD] (5,1) rectangle ++(0.92,0.92);%
\fill[pD] (6,1) rectangle ++(0.92,0.92);%
\fill[pD] (7,1) rectangle ++(0.92,0.92);%
\fill[pD] (8,1) rectangle ++(0.92,0.92);%
\fill[pD] (9,1) rectangle ++(0.92,0.92);%
\fill[pD] (10,1) rectangle ++(0.92,0.92);%
\fill[pD] (11,1) rectangle ++(0.92,0.92);%
\fill[pD] (12,1) rectangle ++(0.92,0.92);%
\fill[pD] (13,1) rectangle ++(0.92,0.92);%
\fill[pD] (14,1) rectangle ++(0.92,0.92);%
\fill[pD] (15,1) rectangle ++(0.92,0.92);%
\fill[pD] (16,1) rectangle ++(0.92,0.92);%
\fill[pD] (17,1) rectangle ++(0.92,0.92);%
\fill[pD] (18,1) rectangle ++(0.92,0.92);%
\fill[pD] (19,1) rectangle ++(0.92,0.92);%
\fill[pD] (20,1) rectangle ++(0.92,0.92);%
\fill[pD] (21,1) rectangle ++(0.92,0.92);%
\fill[pD] (22,1) rectangle ++(0.92,0.92);%
\fill[pD] (23,1) rectangle ++(0.92,0.92);%
\fill[pA] (0,0) rectangle ++(0.92,0.92);%
\fill[pA] (1,0) rectangle ++(0.92,0.92);%
\fill[pA] (2,0) rectangle ++(0.92,0.92);%
\fill[pA] (3,0) rectangle ++(0.92,0.92);%
\fill[pA] (4,0) rectangle ++(0.92,0.92);%
\fill[pA] (5,0) rectangle ++(0.92,0.92);%
\fill[pA] (6,0) rectangle ++(0.92,0.92);%
\fill[pA] (7,0) rectangle ++(0.92,0.92);%
\fill[pA] (8,0) rectangle ++(0.92,0.92);%
\fill[pA] (9,0) rectangle ++(0.92,0.92);%
\fill[pA] (10,0) rectangle ++(0.92,0.92);%
\fill[pA] (11,0) rectangle ++(0.92,0.92);%
\fill[pA] (12,0) rectangle ++(0.92,0.92);%
\fill[pA] (13,0) rectangle ++(0.92,0.92);%
\fill[pA] (14,0) rectangle ++(0.92,0.92);%
\fill[pA] (15,0) rectangle ++(0.92,0.92);%
\fill[pA] (16,0) rectangle ++(0.92,0.92);%
\fill[pA] (17,0) rectangle ++(0.92,0.92);%
\fill[pA] (18,0) rectangle ++(0.92,0.92);%
\fill[pA] (19,0) rectangle ++(0.92,0.92);%
\fill[pA] (20,0) rectangle ++(0.92,0.92);%
\fill[pA] (21,0) rectangle ++(0.92,0.92);%
\fill[pA] (22,0) rectangle ++(0.92,0.92);%
\fill[pA] (23,0) rectangle ++(0.92,0.92);%
}

%% file: figures/v4/gen_pareto.tex
\newcommand{\FrontierEightNoV}{(328.8,0.186) (398.0,0.37) (494.4,0.46) (528.9,0.867) (535.1,1.802) (543.2,4.21) (604.2,8.48) (721.2,21.5)}
\newcommand{\FrontierEightV}{(328.8,0.186) (398.0,0.37) (494.4,0.46) (527.7,0.465) (528.9,0.867) (550.2,0.873) (575.9,1.799) (604.2,8.48) (721.2,21.5)}

%% file: figures/v4/fig_plans.tex
\begin{tikzpicture}[x=0.3cm,y=0.27cm,font=\sffamily\scriptsize]
\foreach \name/\mac/\y/\info in {
  {uniform INT4 + sharing}/\PlanCellsuniformshared/12/{180 / 240 KiB \textperiodcentered{} KL 1.51 \textperiodcentered{} 505 tok/s},
  {read-optimized (knapsack)}/\PlanCellsdtwo/6/{215 / 215 KiB \textperiodcentered{} KL 0.87 \textperiodcentered{} 529 tok/s},
  {memory-optimized (knapsack)}/\PlanCellsctwo/0/{155 / 220 KiB \textperiodcentered{} KL 1.80 \textperiodcentered{} 535 tok/s}} {
  \begin{scope}[shift={(0,\y)}]
    \mac
    \foreach \i in {1,...,23} {\draw[white,line width=0.5pt] (\i-0.04,-0.02) -- (\i-0.04,3.94);}
    \foreach \j in {1,2,3} {\draw[white,line width=0.5pt] (-0.02,\j-0.04) -- (23.94,\j-0.04);}
    \foreach \l/\r in {1/3,2/2,3/1,4/0} {\node[anchor=east,font=\sffamily\tiny] at (-0.1,\r+0.46) {loop \l};}
    \node[anchor=south west,font=\sffamily\scriptsize\bfseries,inner sep=1pt] at (-0.1,4.0) {\name};
    \node[anchor=south east,font=\sffamily\tiny,inner sep=1pt] at (24,4.0) {\info};
  \end{scope}}
\foreach \x/\t in {0/1,11/12,23/24} {\node[anchor=north,font=\sffamily\tiny] at (\x+0.46,-0.1) {\t};}
\node[anchor=north,font=\sffamily\tiny] at (6,-0.1) {layer};
\begin{scope}[shift={(0,-2.6)}]
\fill[pD] (0,0) rectangle ++(0.92,0.92); \node[anchor=west,font=\sffamily\tiny] at (1,0.46) {own INT4};
\fill[pd] (5,0) rectangle ++(0.92,0.92); \node[anchor=west,font=\sffamily\tiny] at (6,0.46) {own INT2};
\fill[pP] (10,0) rectangle ++(0.92,0.92); \node[anchor=west,font=\sffamily\tiny] at (11,0.46) {shared};
\fill[pA] (17.5,0) rectangle ++(0.92,0.92); \node[anchor=west,font=\sffamily\tiny] at (18.5,0.46) {INT4 anchor};
\end{scope}
\end{tikzpicture}

%% file: figures/v4/fig_tlat_col.tex
\begin{tikzpicture}[lbl/.style={font=\sffamily\tiny, inner sep=0.8pt}]
\begin{groupplot}[group style={group size=1 by 2, vertical sep=3mm, x descriptions at=edge bottom},
  width=\columnwidth, height=3.7cm, xmode=log, xmin=11, xmax=92, xtick={12,16,24,32,48,64,80}, xticklabels={12,16,24,32,48,64,80},
  minor xtick={}, grid=major, grid style={black!8},
  tick label style={font=\sffamily\scriptsize}, label style={font=\sffamily\scriptsize},
  ylabel={decode tok/s}, ylabel shift=-2pt,
  xlabel={time per output token (ms, log scale; batch grows to the right)},
  every axis plot/.append style={line width=0.8pt, mark size=1.5pt}]
\nextgroupplot[ymin=0, ymax=690, ytick={0,200,400,600},
  legend columns=3, legend cell align=left, legend style={font=\sffamily\scriptsize, draw=black!25, at={(0.5,1.03)}, anchor=south,
    /tikz/every even column/.append style={column sep=4pt}}]
\node[font=\sffamily\scriptsize\bfseries, anchor=north west] at (rel axis cs:0.01,0.98) {8K context};
\draw[black!45, densely dotted, line width=0.6pt] (axis cs:55.8,0) -- (axis cs:55.8,690);
\node[lbl, text=black!55, anchor=south east] at (axis cs:55.2,8) {TPOT bound};
\addplot[black!65, mark=square*] coordinates {(29.84,33.5) (30.28,66.1) (32.42,123.4) (46.84,170.8) (77.46,154.9)}; \addlegendentry{BF16}
\addplot[teal!45, mark=diamond] coordinates {(12.95,77.2) (25.59,312.6) (39.37,406.4) (70.92,479.4)}; \addlegendentry{KIVI-4}
\addplot[lkblue!45, mark=star, thick] coordinates {(12.64,79.1) (23.54,339.9) (35.28,453.5) (61.12,556.3) (65.14,552.7)}; \addlegendentry{\sys{}-R (ours)}
\addplot[cRes, mark=triangle*] coordinates {(13.75,72.8) (16.73,119.5) (21.17,189.0) (30.12,265.6) (48.66,328.8) (80.01,312.4)}; \addlegendentry{FP8}
\addplot[teal!70!black, dashed, mark=diamond*] coordinates {(12.46,80.3) (24.54,326.0) (36.96,432.9) (64.43,527.7)}; \addlegendentry{KIVI-4 + our V (ablation)}
\addplot[lkblue!90!black, mark=star, very thick] coordinates {(12.67,78.9) (23.46,341.0) (35.38,452.2) (61.51,552.8) (83.19,577.0)}; \addlegendentry{\sys{}-M (ours)}
\node[lbl, anchor=north, text=black!65] at (axis cs:77.5,148) {12};
\node[lbl, anchor=north, text=cRes] at (axis cs:80.0,305) {25};
\node[lbl, anchor=north, text=teal!60!black] at (axis cs:70.9,472) {34};
\node[lbl, anchor=north east, text=teal!70!black] at (axis cs:64.0,522) {34};
\node[lbl, anchor=south, text=lkblue!60] at (axis cs:65.1,568) {36};
\node[lbl, anchor=south, text=lkblue!90!black] at (axis cs:83.2,583) {\textbf{48}};
\nextgroupplot[ymin=0, ymax=340, ytick={0,100,200,300}]
\node[font=\sffamily\scriptsize\bfseries, anchor=north west] at (rel axis cs:0.01,0.98) {16K context};
\draw[black!45, densely dotted, line width=0.6pt] (axis cs:55.8,0) -- (axis cs:55.8,340);
\addplot[black!65, mark=square*] coordinates {(47.84,20.9) (48.48,41.3) (53.82,74.3) (67.44,89.0)};
\addplot[teal!45, mark=diamond] coordinates {(14.35,69.7) (37.54,213.1) (57.92,241.7)};
\addplot[lkblue!45, mark=star, thick] coordinates {(14.02,71.3) (33.12,241.6) (49.64,282.0) (55.43,288.6)};
\addplot[cRes, mark=triangle*] coordinates {(16.05,62.3) (21.18,94.4) (30.19,132.5) (47.94,166.9) (73.73,176.3)};
\addplot[teal!70!black, dashed, mark=diamond*] coordinates {(14.36,69.7) (34.74,230.3) (52.28,267.8)};
\addplot[lkblue!90!black, mark=star, very thick] coordinates {(14.17,70.5) (33.39,239.6) (50.13,279.3) (79.16,303.2)};
\node[lbl, anchor=west, text=black!65] at (axis cs:68.5,89) {6};
\node[lbl, anchor=north, text=cRes] at (axis cs:73.7,170) {13};
\node[lbl, anchor=west, text=teal!60!black] at (axis cs:59.0,238) {14};
\node[lbl, anchor=north east, text=teal!70!black] at (axis cs:51.8,264) {14};
\node[lbl, anchor=south east, text=lkblue!60] at (axis cs:54.8,292) {16};
\node[lbl, anchor=south, text=lkblue!90!black] at (axis cs:79.2,308) {\textbf{24}};
\end{groupplot}
\end{tikzpicture}

%% file: figures/v4/fig_tput.tex
\begin{tikzpicture}
\pgfplotsset{
  fi/.style={ybar, ymajorgrids, grid style={dashed, gray!45}, axis line style={black, line width=0.4pt},
    tick label style={font=\sffamily\scriptsize}, label style={font=\sffamily\scriptsize},
    title style={font=\sffamily\bfseries\scriptsize, yshift=-4pt},
    every axis plot/.append style={draw=black, line width=0.3pt},
    nodes near coords, every node near coord/.append style={font=\sffamily\tiny, rotate=90, anchor=west, black, inner sep=1pt},
    xtick=\empty, xmin=0.55, xmax=1.45, bar width=5.2pt,
    /pgf/number format/.cd, fixed, precision=0, /pgfplots/.cd},
  pbf/.style={fill=white}, pfp8/.style={fill=fbLight}, pint4/.style={fill=fbMid}, pint4v/.style={fill=fbDark},
  pro/.style={fill=fbMid, postaction={pattern=north east lines}},
  pmo/.style={fill=fbNavy, postaction={pattern=north east lines, pattern color=white}},
}
\begin{groupplot}[group style={group size=2 by 2, horizontal sep=1.05cm, vertical sep=0.85cm}, width=4.7cm, height=3.1cm]
\nextgroupplot[fi, ymin=0, ymax=840, ytick={0,200,400,600}, ylabel={tok/s}, title={(a) Ouro-1.4B, 8K}]
\addplot[pbf] coordinates {(1,170.8)}; \addplot[pfp8] coordinates {(1,358.4)}; \addplot[pint4] coordinates {(1,479.4)};
\addplot[pint4v] coordinates {(1,527.7)}; \addplot[pro] coordinates {(1,556.3)}; \addplot[pmo] coordinates {(1,577.0)};
\nextgroupplot[fi, ymin=0, ymax=445, ytick={0,100,200,300}, title={(b) Ouro-1.4B, 16K}]
\addplot[pbf] coordinates {(1,89.0)}; \addplot[pfp8] coordinates {(1,180.7)}; \addplot[pint4] coordinates {(1,241.7)};
\addplot[pint4v] coordinates {(1,267.8)}; \addplot[pro] coordinates {(1,288.6)}; \addplot[pmo] coordinates {(1,303.2)};
\nextgroupplot[fi, ymin=0, ymax=385, ytick={0,100,200,300}, ylabel={tok/s}, title={(c) Ouro-2.6B, 8K}]
\addplot[pbf] coordinates {(1,81.3)}; \addplot[pfp8] coordinates {(1,167.6)}; \addplot[pint4] coordinates {(1,208.3)};
\addplot[pint4v] coordinates {(1,221.7)}; \addplot[pro] coordinates {(1,240.4)}; \addplot[pmo] coordinates {(1,263.9)};
\nextgroupplot[fi, ymin=0, ymax=200, ytick={0,50,100,150}, title={(d) Ouro-2.6B, 16K}]
\addplot[pbf] coordinates {(1,30.8)}; \addplot[pfp8] coordinates {(1,84.5)}; \addplot[pint4] coordinates {(1,109.5)};
\addplot[pint4v] coordinates {(1,118.1)}; \addplot[pro] coordinates {(1,124.1)}; \addplot[pmo] coordinates {(1,136.8)};
\end{groupplot}
\begin{axis}[hide axis, at={(group c1r1.north west)}, anchor=south west, yshift=0.55cm, scale only axis, width=0.1cm, height=0.1cm,
  xmin=0, xmax=1, ymin=0, ymax=1,
  legend style={at={(0,0)}, anchor=south west, draw=gray!60, font=\sffamily\scriptsize, legend columns=3,
    /tikz/every even column/.append style={column sep=6pt}},
  legend cell align=left]
\addlegendimage{area legend, pbf, draw=black}\addlegendentry{BF16}
\addlegendimage{area legend, pfp8, draw=black}\addlegendentry{FP8 + sharing}
\addlegendimage{area legend, pint4, draw=black}\addlegendentry{uniform INT4}
\addlegendimage{area legend, pint4v, draw=black}\addlegendentry{KIVI-4 + V}
\addlegendimage{area legend, pro, draw=black}\addlegendentry{\sys{}-R}
\addlegendimage{area legend, pmo, draw=black}\addlegendentry{\sys{}-M}
\end{axis}
\end{tikzpicture}

%% file: figures/v4/fig_pareto_full.tex
\begin{tikzpicture}[lab/.style={font=\sffamily\scriptsize,inner sep=1pt}]
\begin{axis}[width=\textwidth,height=4.3cm,ymode=log,ymin=0.08,ymax=40,xmin=140,xmax=760,
  xlabel={peak decode tok/s at 8K context, vLLM (higher is better)},ylabel={fp32 KL / \Pz{} (lower is better)},
  grid=major,grid style={black!8},tick label style={font=\sffamily\scriptsize},label style={font=\sffamily\scriptsize},
  legend columns=6,legend style={font=\sffamily\scriptsize,draw=black!25,at={(0.5,1.03)},anchor=south,/tikz/every even column/.append style={column sep=4pt}}]
\addlegendimage{lkblue,thick}\addlegendentry{frontier with V precision}
\addlegendimage{black!45,dashed,thick}\addlegendentry{frontier without}
\addlegendimage{only marks,mark=diamond*,mark size=3pt,lkblue!70!black}\addlegendentry{+V, fused kernel and graph}
\addlegendimage{only marks,mark=*,mark size=2pt,cPlan}\addlegendentry{knapsack plans}
\addlegendimage{only marks,mark=pentagon*,mark size=2.6pt,cDm}\addlegendentry{hand-made + sharing}
\addlegendimage{only marks,mark=*,mark size=1.8pt,cRes}\addlegendentry{INT4 + residuals}
\addplot[black!45,dashed,thick,forget plot] coordinates {\FrontierEightNoV};
\addplot[lkblue,thick,forget plot] coordinates {\FrontierEightV};
\addplot[only marks,mark=square*,mark size=2pt,cBase,forget plot] coordinates {(170.8,0.09) (328.8,0.186)};
\addplot[only marks,mark=diamond*,mark size=2.6pt,cBase,forget plot] coordinates {(494.4,0.46) (721.2,21.5)};
\addplot[only marks,mark=triangle*,mark size=2.6pt,cPtr,forget plot] coordinates {(182.1,1.00) (358.4,1.204)};
\addplot[only marks,mark=*,mark size=1.8pt,cRes,forget plot] coordinates {(370.3,1.42) (385.6,1.83) (386.3,2.12) (416.8,4.38)};
\addplot[only marks,mark=pentagon*,mark size=2.6pt,cDm,forget plot] coordinates {(505.4,1.507) (543.2,4.21) (604.2,8.48)};
\addplot[only marks,mark=*,mark size=2pt,cPlan,forget plot] coordinates {(479.8,0.47) (398.0,0.37) (401.9,1.08) (519.4,1.11) (528.9,0.867) (535.1,1.802)};
\addplot[only marks,mark=diamond*,mark size=3pt,lkblue!70!black,forget plot] coordinates {(550.2,0.873) (575.9,1.799) (527.7,0.465)};
\node[lab,anchor=south west] at (axis cs:174,0.092) {BF16 (KL 0)};
\node[lab,anchor=north west] at (axis cs:332,0.18) {FP8};
\node[lab,anchor=north east] at (axis cs:490,0.44) {uniform INT4};
\node[lab,anchor=east] at (axis cs:713,21.5) {uniform INT2};
\node[lab,anchor=south west,text=cPtr] at (axis cs:178,1.08) {BF16 + sharing};
\node[lab,anchor=north,text=cPtr] at (axis cs:358.4,1.1) {FP8 + sharing};
\node[lab,anchor=east,text=cDm] at (axis cs:500,1.507) {U4 + sharing};
\node[lab,anchor=north west,text=cDm] at (axis cs:548,4.0) {INT4/INT2 + sharing};
\node[lab,anchor=south east,text=cDm] at (axis cs:600,9.0) {INT2 + sharing};
\node[lab,anchor=east,text=cPlan] at (axis cs:525,0.80) {R};
\node[lab,anchor=south east,text=cPlan] at (axis cs:533,1.9) {M};
\node[lab,anchor=west,text=lkblue!70!black] at (axis cs:556,0.873) {\textbf{R+V}};
\node[lab,anchor=west,text=lkblue!70!black] at (axis cs:582,1.799) {\textbf{M+V}};
\node[lab,anchor=north west,text=lkblue!70!black] at (axis cs:531,0.44) {U4+V};
\end{axis}
\end{tikzpicture}

%% file: figures/v4/fig_pareto.tex
\begin{tikzpicture}[lab/.style={font=\sffamily\scriptsize,inner sep=1.5pt}]
\begin{axis}[width=\textwidth,height=4.0cm,ymode=log,ymin=0.08,ymax=40,xmin=140,xmax=760,
  xlabel={peak decode tok/s at 8K context, vLLM (higher is better)},ylabel={fp32 KL / \Pz{} (lower is better)},
  grid=major,grid style={black!8},tick label style={font=\sffamily\scriptsize},label style={font=\sffamily\scriptsize},
  legend columns=4,legend style={font=\sffamily\scriptsize,draw=black!25,at={(0.5,1.03)},anchor=south,/tikz/every even column/.append style={column sep=6pt}}]
\addlegendimage{black!55,thick}\addlegendentry{measured frontier}
\addlegendimage{only marks,mark=square*,mark size=2.2pt,black!60}\addlegendentry{vLLM BF16 / FP8}
\addlegendimage{only marks,mark=diamond*,mark size=2.8pt,teal!70!black}\addlegendentry{KIVI-4 / KIVI-2}
\addlegendimage{only marks,mark=triangle*,mark size=2.8pt,cRes}\addlegendentry{loop sharing (BF16 / FP8)}
\addlegendimage{only marks,mark=pentagon*,mark size=2.6pt,cDm}\addlegendentry{KIVI-4 + loop sharing}
\addlegendimage{only marks,mark=*,mark size=2.2pt,cPlan}\addlegendentry{ResidualQuant-style}
\addlegendimage{only marks,mark=star,mark size=3.6pt,very thick,lkblue!80!black}\addlegendentry{\sys{} (ours)}
\addplot[black!55,thick,forget plot] coordinates {(170.8,0.09) (328.8,0.186) (494.4,0.46) (550.2,0.873) (575.9,1.799) (721.2,21.5)};
\addplot[only marks,mark=square*,mark size=2.2pt,black!60,forget plot] coordinates {(170.8,0.09) (328.8,0.186)};
\addplot[only marks,mark=diamond*,mark size=2.8pt,teal!70!black,forget plot] coordinates {(494.4,0.46) (721.2,21.5)};
\addplot[only marks,mark=triangle*,mark size=2.8pt,cRes,forget plot] coordinates {(182.1,1.00) (358.4,1.204)};
\addplot[only marks,mark=pentagon*,mark size=2.6pt,cDm,forget plot] coordinates {(505.4,1.507)};
\addplot[only marks,mark=*,mark size=2.2pt,cPlan,forget plot] coordinates {(416.8,4.38)};
\addplot[only marks,mark=star,mark size=3.6pt,very thick,lkblue!80!black,forget plot] coordinates {(550.2,0.873) (575.9,1.799)};
\node[lab,anchor=south west] at (axis cs:174,0.092) {BF16 (KL 0)};
\node[lab,anchor=north west] at (axis cs:333,0.18) {FP8 (vLLM)};
\node[lab,anchor=north west] at (axis cs:498,0.44) {KIVI-4};
\node[lab,anchor=east] at (axis cs:712,21.5) {KIVI-2};
\node[lab,anchor=south west] at (axis cs:186,1.03) {loop sharing (BF16; = 1)};
\node[lab,anchor=north] at (axis cs:358.4,1.12) {loop sharing + FP8};
\node[lab,anchor=south east] at (axis cs:501,1.6) {KIVI-4 + loop sharing};
\node[lab,anchor=east] at (axis cs:410,4.38) {ResidualQuant-style};
\node[lab,anchor=east,text=lkblue!80!black] at (axis cs:543,0.95) {\textbf{\sys{}-R (ours)}};
\node[lab,anchor=west,text=lkblue!80!black] at (axis cs:584,1.799) {\textbf{\sys{}-M (ours)}};
\end{axis}
\end{tikzpicture}

%% file: figures/v4/fig_slocap.tex
\begin{tikzpicture}
\begin{axis}[xbar, width=0.74\columnwidth, height=4.2cm, bar width=6.5pt, xmin=0, xmax=9.7, ymin=0.4, ymax=7.6,
  ytick={1,2,3,4,5,6,7}, y dir=reverse,
  yticklabels={\textbf{\sys{}-M (ours)},\textbf{\sys{}-R (ours)},KIVI-4 + our V,KIVI-4,FP8 (vLLM),FP8 + loop-3 sharing,BF16 (vLLM)},
  xtick={0,2,4,6,8}, xticklabels={0,2$\times$,4$\times$,6$\times$,8$\times$},
  xmajorgrids, grid style={dashed, gray!40}, axis line style={black, line width=0.4pt},
  tick label style={font=\sffamily\scriptsize}, label style={font=\sffamily\scriptsize},
  xlabel={16K SLO goodput, multiple of BF16 (seed 101)},
  every axis plot/.append style={draw=black, line width=0.3pt, bar shift=0pt},
  nodes near coords={\pgfmathprintnumber[fixed, fixed zerofill, precision=1]{\pgfplotspointmeta}$\times$},
  every node near coord/.append style={font=\sffamily\scriptsize, anchor=west, xshift=1pt}]
\addplot[fill=fbNavy, postaction={pattern=north east lines, pattern color=white}, every node near coord/.append style={font=\sffamily\scriptsize\bfseries}] coordinates {(8.00,1)};
\addplot[fill=fbMid, postaction={pattern=north east lines}] coordinates {(6.00,2)};
\addplot[fill=fbDark] coordinates {(5.00,3)};
\addplot[fill=fbMid] coordinates {(4.00,4)};
\addplot[fill=fbLight] coordinates {(3.20,5)};
\addplot[fill=fbLight, postaction={pattern=north east lines}] coordinates {(2.56,6)};
\addplot[fill=white] coordinates {(1.00,7)};
\end{axis}
\end{tikzpicture}

%% file: figures/v4/fig_codecs.tex
\begin{tikzpicture}[lab/.style={font=\sffamily\footnotesize,inner sep=1pt}]
\begin{axis}[width=0.72\textwidth,height=7cm,ymode=log,ymin=0.12,ymax=45,xmin=100,xmax=560,
  xlabel={history KV read per token, KiB (4 loops)},ylabel={fp32 KL / \Pz{}},
  grid=major,grid style={black!8},tick label style={font=\sffamily\small},label style={font=\sffamily\small}]
\addplot[only marks,mark=*,mark size=1.8pt,cBase] coordinates {(174,3.482) (162.5,4.950) (156,2.752) (156,10.081) (204,1.050) (240,0.460) (144,21.485) (120,33.427) (126,24.748) (384,0.186)};
\addplot[only marks,mark=triangle*,mark size=2.4pt,cRes] coordinates {(540,0.303) (310.5,2.005)};
\addplot[only marks,mark=triangle*,mark size=2.4pt,cPtr] coordinates {(384,1.204)};
\addplot[only marks,mark=*,mark size=2pt,cPlan] coordinates {(215,0.867) (220,1.802)};
\addplot[only marks,mark=star,mark size=3pt,thick,lkblue] coordinates {(186.3,0.873) (188.9,1.799)};
\node[lab,anchor=west] at (axis cs:132,27) {2-bit codecs (KIVI-2, TurboQuant-2.5)};
\node[lab,anchor=west,align=left] at (axis cs:178,5.2) {3-bit codecs (TurboQuant-3.5,\\ KVQuant-3b, rotated 3-bit)};
\node[lab,anchor=south] at (axis cs:204,1.12) {rot-4b};
\node[lab,anchor=north west] at (axis cs:240,0.45) {KIVI-4 g32};
\node[lab,anchor=west] at (axis cs:388,0.186) {FP8};
\node[lab,anchor=west,text=cPtr] at (axis cs:388,1.204) {FP8 + sharing};
\node[lab,anchor=east,text=cRes] at (axis cs:535,0.303) {FlashLoop chain};
\node[lab,anchor=west,text=cRes] at (axis cs:314,2.0) {ResidualQuant (reimpl.)};
\node[lab,anchor=west,text=cPlan] at (axis cs:219,0.86) {R};
\node[lab,anchor=west,text=cPlan] at (axis cs:223,1.802) {M};
\node[lab,anchor=east,text=lkblue] at (axis cs:183,0.873) {R+V};
\node[lab,anchor=east,text=lkblue] at (axis cs:185,1.8) {M+V};
\end{axis}
\end{tikzpicture}